\documentclass{article}

\PassOptionsToPackage{numbers,sort&compress}{natbib}

\usepackage[eandd,final]{neurips_2026} 

\usepackage[utf8]{inputenc} 
\usepackage[T1]{fontenc}    

\usepackage{hyperref}       
\usepackage{url}            
\usepackage{booktabs}       
\usepackage{amsfonts}       
\usepackage{nicefrac}       
\usepackage{microtype}      
\usepackage{xcolor}         

\usepackage{graphicx}
\usepackage{amssymb}

\usepackage{algorithm}

\usepackage{algpseudocode}
\usepackage{amsmath}
\usepackage{mathrsfs}
\usepackage{booktabs}
\usepackage{lscape}
\usepackage{multirow}
\usepackage{tabularx} 
\usepackage{arydshln}
\usepackage{dashrule}

\usepackage[breakable]{tcolorbox}
\usepackage{multicol}
\usepackage{xcolor}

\usepackage{enumitem}

\usepackage{xspace}
\usepackage{wrapfig}

\definecolor{paperlink}{RGB}{0,70,140}

\hypersetup{
  colorlinks=true,
  urlcolor=magenta,
  linkcolor=black,
  citecolor=black
}

\newif\ifcrchanges
\crchangestrue

\newlength{\crtablecaptionskip}
\newcommand{\crtablespacing}{%
  \setlength{\abovecaptionskip}{\crtablecaptionskip}%
  \setlength{\belowcaptionskip}{0pt}%
}

\def\ourdataset{\textsc{ManiTaskOT-200}\xspace}
\def\ourgentask{ManiTaskGen-RAS-40K}

\title{ManiTaskGen: A Comprehensive Task Generator for Benchmarking and Improving Vision-Language Agents on Embodied Decision-Making}

\author{%
  \textbf{Liu Dai\textsuperscript{*}}\quad
  \textbf{Haina Wang\textsuperscript{*}}\quad
  \textbf{Weikang Wan}\quad
  \textbf{Hao Su}\\[0.6em]
  University of California San Diego, USA\\
  \textsuperscript{*}Equal contribution \\
  {\normalfont
    \texttt{\{l2dai,haw120,w2wan,haosu\}@ucsd.edu}
  }\\
}

\begin{document}
\setcounter{tocdepth}{-3}

\maketitle

\begin{abstract}
Building embodied agents capable of accomplishing arbitrary tasks is a core objective towards achieving embodied artificial general intelligence (E-AGI). While recent work has advanced such general robot policies, their training and evaluation are often limited to tasks within specific scenes, involving restricted instructions and scenarios. Existing benchmarks also typically rely on manual annotation of limited tasks in a few scenes. We argue that exploring the full spectrum of feasible tasks within any given scene is crucial, as they provide both extensive benchmarks for evaluation and valuable resources for agent improvement.
Towards this end, we introduce ManiTaskGen, 
a novel system that automatically generates comprehensive, diverse, feasible mobile manipulation tasks for any given scene.
The generated tasks encompass both process-based, specific instructions (e.g., "move object from X to Y") and outcome-based, abstract instructions (e.g., "clear the table"). We apply ManiTaskGen to both simulated and real-world scenes, demonstrating the validity and diversity of the generated tasks. We then leverage these tasks to automatically construct benchmarks, evaluating high-level embodied decision-making through a grounded skill interface for agents built upon existing vision-language models (VLMs). Furthermore, we propose a simple yet effective method that utilizes ManiTaskGen tasks to enhance embodied decision-making.
Overall, this work presents 
a cross-scene task generation framework facilitating both benchmarking and improvement of embodied decision-making agents. Our code is available at \url{https://manitaskgen.github.io}.
\end{abstract}

\section{Introduction}
Consider an embodied agent endowed with robust primitive skills for mobile manipulation: the ability to navigate to any accessible location, grasp any movable object, and place it wherever it fits. A fundamental question then arises: \textit{what is the full extent of the task space this agent can successfully address in a given environment?} This space appears infinite. 
Despite this immense potential task space, recent efforts have focused on developing general-purpose embodied agents~\cite{szot2024multimodal, huang2023embodied, batra2020rearrangement, ahn2022can, brohan2023rt, driess2023palm, szot2023large} capable of completing any feasible task within it. 
A promising direction leverages VLMs~\cite{Meta2024Llama3.2, team2024gemma, bai2023qwen} for high-level decision-making, followed by either the explicit composition of skill primitives~\cite{gao2024physically, huang2022language, gu2022multi} or the implicit integration with underlying action modules to predict joint-level actions, the latter thread is commonly referred to as Vision-Language-Action (VLA) models~\cite{kim2024openvla, black2410pi0, brohan2023rt}. To evaluate such agents, numerous benchmarks~\cite{zhang2024m3bench, li2025embodied, liu2024visualagentbench, choi2024lota, yang2025embodiedbench} have also been proposed to assess their decision-making capabilities.
\begin{figure}[t]
    \centering
    \includegraphics[width=\columnwidth]{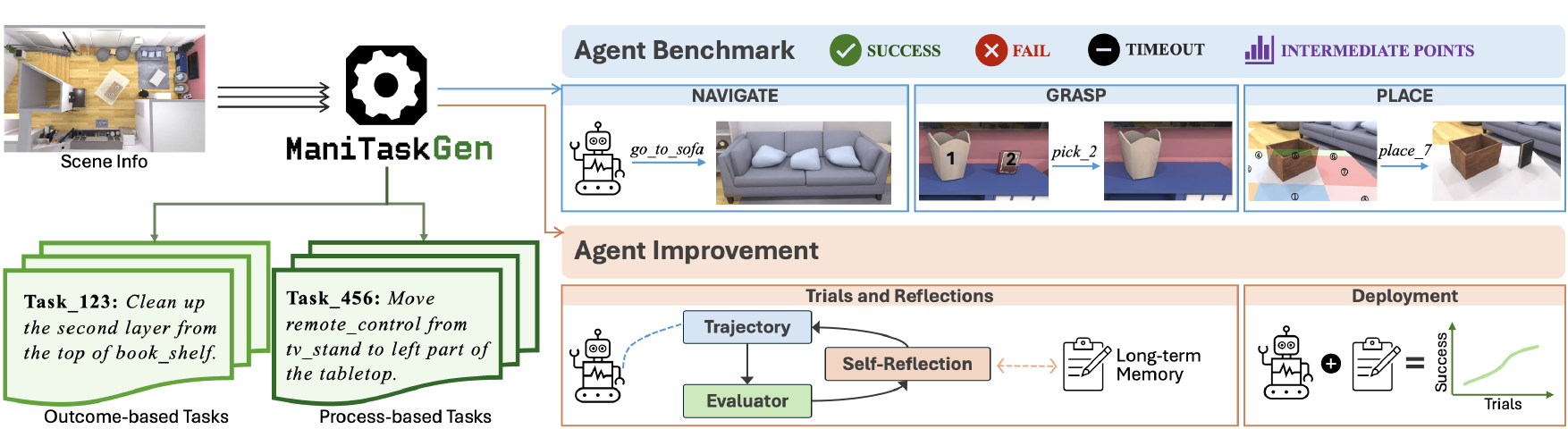}
    \vspace{-7mm}
    \caption{\textbf{Framework Overview.} ManiTaskGen systematically generates process-based and outcome-based navigate–pick–place tasks from scene information. The generated tasks support benchmark construction and feedback-based improvement of embodied decision-making agents.} 
    \vspace{-7mm}
    \label{fig:teaser}
\end{figure}

Despite these efforts, existing work often faces significant limitations regarding the tasks used in both policy training and benchmark design. Common practice~\cite{li2023behavior, gu2022multi, cheng2025embodiedeval, yang2025embodiedbench} involves manually selecting a few scenes and authoring a limited number of scene-specific tasks related to objects within those scenes, which serve as training or testing objectives for embodied agents. Such manual approaches are labor-intensive and require considerable human effort to design suitable tasks, resulting in prohibitively high scaling costs and making it nearly impossible to generalize task creation to the diverse range of scenes encountered in both simulated and real-world environments. This fundamental limitation creates a substantial gap between current agents and the ultimate goal of E-AGI: while the aim is an agent capable of universal generalization across diverse scenes and tasks, existing works are confined to training and evaluating the agents within limited scenarios and task variations.

\begin{wrapfigure}{r}{0.48\textwidth}
  \centering
  \includegraphics[width=1.0\linewidth]{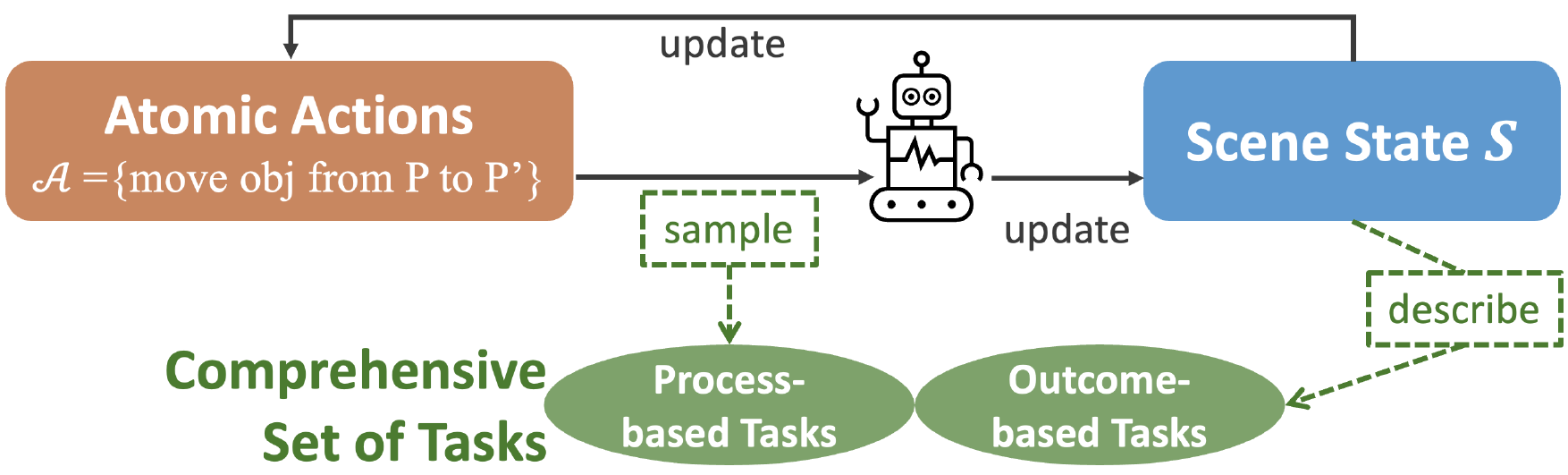}
  \vspace{-5mm}
  \caption{\textbf{Agent-Env Interaction Cycle.} 
  This cycle provides a fundamental depiction of how an embodied agent executes a task in a scene.
  }
  \label{fig:exec_loop}
  \vspace{-1mm}
\end{wrapfigure}

To address these limitations, we present ManiTaskGen, a cross-scene task generator for navigate--pick--place rearrangement tasks. Given object poses, bounding boxes, and optional meshes, it systematically generates feasible tasks under an explicit skill and placement representation.
We ensure logical comprehensiveness by grounding task generation in a systematic analysis of the fundamental agent-environment interaction cycle, as conceptually illustrated in Fig.~\ref{fig:exec_loop}.

This cycle, where executing atomic actions (e.g., single object relocation) updates the scene state and determines subsequent available actions, inherently defines the space of all possible tasks. Based on this inherent structure, we rigorously categorize all possible tasks into two principal types:
\textbf{process-based tasks}, capturing action sequences or trajectories (e.g., "move object A from X to Y"), and \textbf{outcome-based tasks}, representing reachable target states (e.g., "make the table clean"). ManiTaskGen employs distinct strategies for generating each type. Process-based tasks are generated by explicitly sampling and composing atomic action sequences, which are derived from a novel Receptacle-Aware 3D Scene Graph encoding all objects and fine-grained potential placements within the scene. Outcome-based tasks are generated using a hybrid template-based approach combined with VLM voting mechanism to produce diverse descriptions of plausible target states.

We assess the validity and diversity of the generated tasks by applying ManiTaskGen to both simulated (e.g., ReplicaCAD~\cite{szot2021habitat}, AI2THOR~\cite{ai2thor}) and real-world scenes (e.g., SUN-RGBD~\cite{song2015sun}). Furthermore, we propose a framework for automatically constructing benchmarks in simulators using the generated tasks to assess embodied agent decision-making capabilities. Finally, we demonstrate how feedback from generated tasks can improve VLM agents at inference time through self reflection ~\cite{shinn2023reflexion, yao2023react} .

In summary, our contributions include:
(1) We propose ManiTaskGen, a universal system for cross-scene task generation and coverage within the defined navigate--pick--place rearrangement scope.
(2) Leveraging the generated tasks, we evaluate the high-level embodied decision-making capabilities of current VLMs under a grounded skill interface.
(3) We demonstrate the utility of ManiTaskGen tasks for improving embodied decision-making through an inference-time self-reflection method.

\section{Related Work}

\noindent\textbf{Task Generation for Embodied Agents.}
Recent efforts have explored task generation for digital agents~\cite{hu2024agentgen, jin2023mini, cobbe2020leveraging, raileanu2020ride} and augmenting RL objectives~\cite{fang2020adaptive, wang2023gensim}. Among these, ALFRED~\cite{shridhar2020alfred} is the most relevant to our work. It combines a procedural task planner with human-annotated task directives. However, ALFRED focuses on expanding task trajectories rather than task definitions and lacks a comprehensive coverage of both process- and outcome-based mobile manipulation tasks. Moreover, it is limited to specific scenes. 
In contrast, ManiTaskGen enables scene-grounded mobile manipulation task generation across scenes, emphasizing scalability and variability in task formulation.

\noindent\textbf{Datasets and Benchmarks for Embodied Agents.}
Numerous datasets and benchmarks have been proposed for training~\cite{o2024open, ai2thor, habitatrearrangechallenge2022, batra2020rearrangement, mu2021maniskill} and evaluating~\cite{li2023behavior, shridhar2020alfred, mu2021maniskill} embodied agents, including those tailored to LLM/VLM-based decision-making agents~\cite{li2025embodied, liu2024visualagentbench, yang2025embodiedbench, cheng2025embodiedeval}. A common limitation of these benchmarks is their reliance on manually annotated tasks confined to a finite set of predefined scenes. In contrast, ManiTaskGen automates task generation across compatible scenes under the adopted skill and scene representation. Beyond serving as a static dataset, it also provides  dynamic evaluation and optimization platform for embodied decision-making agents, making it a more versatile resource.

\noindent\textbf{Improvement of VLM-based Agents.}
Numerous studies have explored various strategies to improve LLM/VLM-based digital agents~\cite{hong2024cogagent, zhai2024fine, pan2024autonomous, yao2023react, luong2024reft} and embodied agents~\cite{ramrakhya2025grounding, wang2024rl, black2410pi0, szot2023large, szot2024grounding}.
One important line of work employs collected demonstrations for supervised fine-tuning (SFT)~\cite{black2410pi0, hong2024cogagent, song2024trial, zeng2023agenttuning}. Given sufficient data, SFT ensures the resulting model achieves comparable performance with the source demonstrations. However, demonstration collection is labor-intensive and expensive, making it challenging to scale across diverse tasks and scenarios.
Another prominent thread of methods leverages Reinforcement Learning (RL)~\cite{zhai2024fine, ziegler2019fine, shinn2023reflexion, grigsby2025vlm} to tune agent policies, through learning from reward models based on human feedback~\cite{ouyang2022training, ziegler2019fine}, task-specific handcrafted reward functions~\cite{carta2023grounding, zhai2024fine}, or reward generators~\cite{ramrakhya2025grounding, shinn2023reflexion}.
To demonstrate the application of ManiTaskGen tasks for agent improvement, we designed and implemented a simple inference-time RL method inspired by Reflexion~\cite{shinn2023reflexion} and ReAct~\cite{yao2023react}. The method leverages the generated tasks and evaluations for self-reflection, and experimental results demonstrate its effectiveness.
Future work involves extending ManiTaskGen's compatibility with more advanced RL-based improvement methods~\cite{ramrakhya2025grounding, wang2024rl}. Overall, ManiTaskGen produces abundant resources for agent improvement, including task instructions as well as final and sub-step evaluations, thereby empowering various approaches to tune VLM-based agents.

\section{ManiTaskGen: Comprehensive Task Generation}
\label{sec:task_gen}

\subsection{Premise and Formalization}
\label{subsec:premise}
We first formalize the problem space for generating comprehensive mobile manipulation tasks solvable by an agent equipped with fundamental navigate, pick, and place capabilities.

We define the scene state $\mathcal{S}$ by the states of all objects $\mathcal{O}=\{o_1, \dots, o_N\}$ within it. The state of an object $o_i$ is $s_i = (p_i, c_i)$, including its pose $p_i$ and its containment state $c_i$. The containment state specifies the surface it is currently located on (e.g., "on surface A", "on internal surface B of a multi-layer object") or if it is held by the agent's gripper ('held'). The scene state is $\mathcal{S} = \{s_1, \dots, s_N\}$.

The set of atomic actions available to the agent is $\mathcal{A} = \{a_0, a_1, a_2, \dots\}$. Based on the agent's core capabilities, we define an atomic action $a \in \mathcal{A}$ as a parameterized object relocation operation. This conceptually represents a single, high-level step in the interaction, such as moving object $o$ from its current location $p$ to a new valid placement position $p'$. Formally, executing a feasible action $a$ in state $\mathcal{S}$ transitions the scene to a new state $\mathcal{S}' = a(\mathcal{S})$. These actions primarily modify the position and containment state of the manipulated object, and update the gripper's state.

The fundamental agent-environment interaction (as conceptually illustrated in Fig.~\ref{fig:exec_loop}) follows a discrete-time cycle $\mathcal{S}_t \xrightarrow{a_t} \mathcal{S}_{t+1}$, where at each step $t$, the agent selects an action $a_t$ from the set of atomic actions feasible in state $\mathcal{S}_t$. The feasibility of actions depends on the current state (e.g., an object must be movable; a target location must be a valid surface that can accommodate the object).

Based on this formal framework, we categorize tasks within the navigate–pick–place interaction model according to whether their objectives specify an action sequence or a desired final state. \textbf{Process-based tasks} explicitly specify a desired sequence of atomic actions $\langle a_1, \dots, a_n \rangle$ that constitute a feasible trajectory through the state space. For example, the instruction ``Move the book from the shelf to the left part of the tabletop'' corresponds to a sequence of parameterized object relocation actions. \textbf{Outcome-based tasks} specify a desired target scene state $\mathcal{S}_{target}$ without dictating the intermediate actions required to reach it. For example, the instruction ``Clear the table'' corresponds to a target state where the containment states of relevant objects satisfy specific criteria relative to the table (e.g., no objects are on the table's surface).

This formalization provides a rigorous basis for systematically defining and generating task instances. Sec.~\ref{subsec:scene_graph} will detail how to utilize the given scene information to construct a Receptacle-Aware 3D Scene Graph which enables comprehensive retrieval of all objects and receptacles. In Sec.~\ref{subsec:task_gen}, we illustrate how to further generate both process-based and outcome-based tasks.

\subsection{Receptacle-Aware 3D Scene Graph Construction}
\label{subsec:scene_graph}
We construct a scene graph $\mathbb{S}$ which serves as a structured representation of the scene, encoding both objects and available receptacles. The input scene information includes: object poses, object bounding boxes, and object mesh models (optional for extracting interior receptacles within objects). 

The construction process involves two main steps. First, we initialize the scene graph as a structural object spatial relationship tree. This tree is based on spatial containment on surfaces: the root node represents the ground, while all other nodes correspond to scene objects. Parent-child relationships are determined by spatial containment---an object is assigned as a child of another if it rests on its surface. In addition, we will extract the interior surfaces (if any) of an object if its mesh model is available, and provide a more precise record of multi-level object placement during tree initialization. We also compute and record relative positions and distances between objects on the same surface.

\begin{wrapfigure}{r}{0.48\textwidth}
    \centering
    \includegraphics[width=\linewidth]{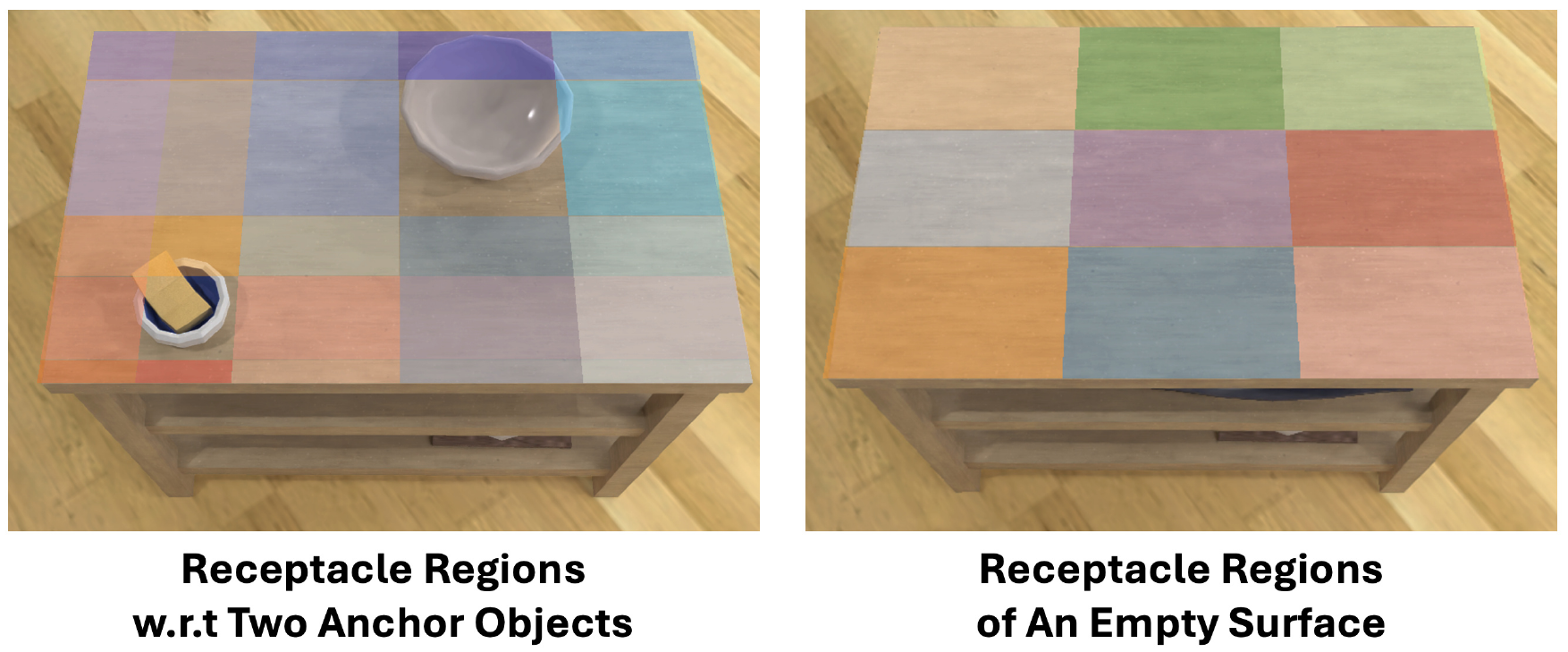}
    \vspace{-5mm} 
    \caption{\textbf{Visualization of Segmented Receptacle Regions on a Surface.} 
    }
    \label{fig:free_space}
\end{wrapfigure}
After that, we identify the available receptacles within the scene. Building upon the established object relationship tree, we characterize receptacles by segmenting the free space of surfaces. Specifically, we treat each object as an anchor, and segment the free space around it into eight directional receptacle regions, with boundaries extending until they encounter another object or the edge of the supporting surface. For unoccupied surfaces, we employ a 3×3 grid segmentation by default. Visualization of our receptacle segmentation is provided in Fig.~\ref{fig:free_space}. Receptacle properties such as location, size, and direction relative to its anchor object or the surface are recorded. We also capture complex cases, such as a single receptacle indexed by multiple objects or the potential for merging adjacent receptacles into larger ones. Finally, all these information are stored as attributes of each node of the scene graph. This representation enables efficient retrieval of any receptacle by querying objects or surfaces. We include more details on the scene graph construction process in Appendix~\ref{appendix_sec:scene_graph}.

The resulting 3D Scene Graph $\mathbb{S}$ integrates object nodes (encoding properties like pose, bounding box, and relationships) and receptacle information (encoding properties like location, size, and relationships to objects/surfaces). This unified representation of the scene is crucial for deriving the set of feasible atomic object relocation actions, as detailed in the following section.

\subsection{Generating Tasks}
\label{subsec:task_gen}
Building upon the Receptacle-Aware 3D Scene Graph $\mathbb{S}$ introduced in Sec.~\ref{subsec:scene_graph}, this subsection details our methodologies for generating comprehensive sets of process- and outcome-based tasks.

\noindent\textbf{Atomic Action Derivation}. We first derive the complete set of feasible atomic actions ($\mathcal{A}$) based on $\mathbb{S}$, which explicitly encodes the properties of every object (location, size, relationships) as well as all available receptacles. As formalized in Sec.~\ref{subsec:premise}, an atomic action corresponds to a parameterized object relocation operation, such as moving object $o$ from its current location $P$ to a new valid placement position $P'$. Accordingly, by identifying every object-receptacle pair where the object can be feasibly placed on the receptacle's surface, we enumerate the full set of possible target locations $P'$ for each object. The set of atomic actions $\mathcal{A}$ is thus defined by all currently feasible object-to-receptacle relocation operations within the scene.

\noindent\textbf{Process-based Tasks}. As formalized in Sec.~\ref{subsec:premise}, process-based tasks explicitly specify a desired sequence of atomic actions $\langle a_1, \dots, a_n \rangle$. Leveraging the agent-environment interaction cycle (Fig.~\ref{fig:exec_loop}) and the derived atomic action set $\mathcal{A}$, we generate diverse process-based tasks by systematically sampling and composing feasible action sequences. Single-step tasks are simply individual actions sampled from $\mathcal{A}$. Multi-step tasks are formed by chaining sequences of actions. The interaction cycle enables this chaining: after executing an action $a_t$ in state $\mathcal{S}_t$, it transitions to $\mathcal{S}_{t+1}$, then we can sample a subsequent action $a_{t+1}$ that is feasible in the new state $\mathcal{S}_{t+1}$.

We compose these sequential actions using logical connectors to form complex process-based task instructions. The most common one for defining a task sequence is THEN (e.g., "Execute $a_t$ THEN execute $a_{t+1}$"). Other connectors like AND or OR can also be used to generate more diverse task structures, reflecting different types of process specifications. By iteratively sampling and chaining such feasible actions across multiple steps, we construct process-based tasks of varying lengths and complexity. 

To further enhance linguistic and spatial diversity, we define and describe the target location for each object relocation action within a sampled sequence using distinct spatial strategies derived from the scene graph. These strategies include: \textit{move to a named surface}, \textit{move to a location around a specific object}, \textit{move to a location with specific direction relative to a specific object}, or \textit{move to a location between two objects}. Furthermore, LLMs can be optionally used to rephrase the entire generated task instruction, increasing linguistic variation while preserving semantics.

\noindent\textbf{Outcome-based Tasks}. Generating outcome-based tasks, which define desired target scene states, is more challenging as it requires abstract state descriptions. A naive approach of using VLMs or LLMs directly on the scene information often yields impractical tasks and limited diversity due to model limitations in understanding complex 3D scenes (further discussed in Sec.~\ref{subsec:task_diversity}).

To address this, we employ a hybrid approach combining template-based generation and VLM-based filtering. We introduce \ourdataset, a manually curated outcome-based task template dataset acquired from human-written instructions on diverse scenes, comprising 200 structured templates (details in Appendix~\ref{appendix_subsec:dataset}). Examples of these templates include:
\begin{quote}
    "Create a tidy arrangement on [PLATFORM0]." \\
    "Design a "jigsaw puzzle" on [PLATFORM0] by arranging the objects in a way that creates a cohesive whole."\\
    "Reorganize the items on [PLATFORM0] and [PLATFORM1] to create a thematic display that tells a story or represents a concept."
\end{quote}
Given a scene, we generate outcome-based tasks by instantiating \ourdataset templates with scene-specific objects. To screen task feasibility, we then employ an ensemble of VLMs to vote on the executability of each generated instruction, filtering out impractical tasks and refining the final set. We provide more details on this process in Appendix~\ref{appendix_subsec:vlm_voting}, and a separate analysis for filtering and execution-judging reliability in Appendix~\ref{appendix_subsec:evaluation_discrepancy} .

\section{Evaluation of ManiTaskGen Tasks }

\begin{figure*}[t]
    \centering
    \includegraphics[width=0.9\linewidth
    ]{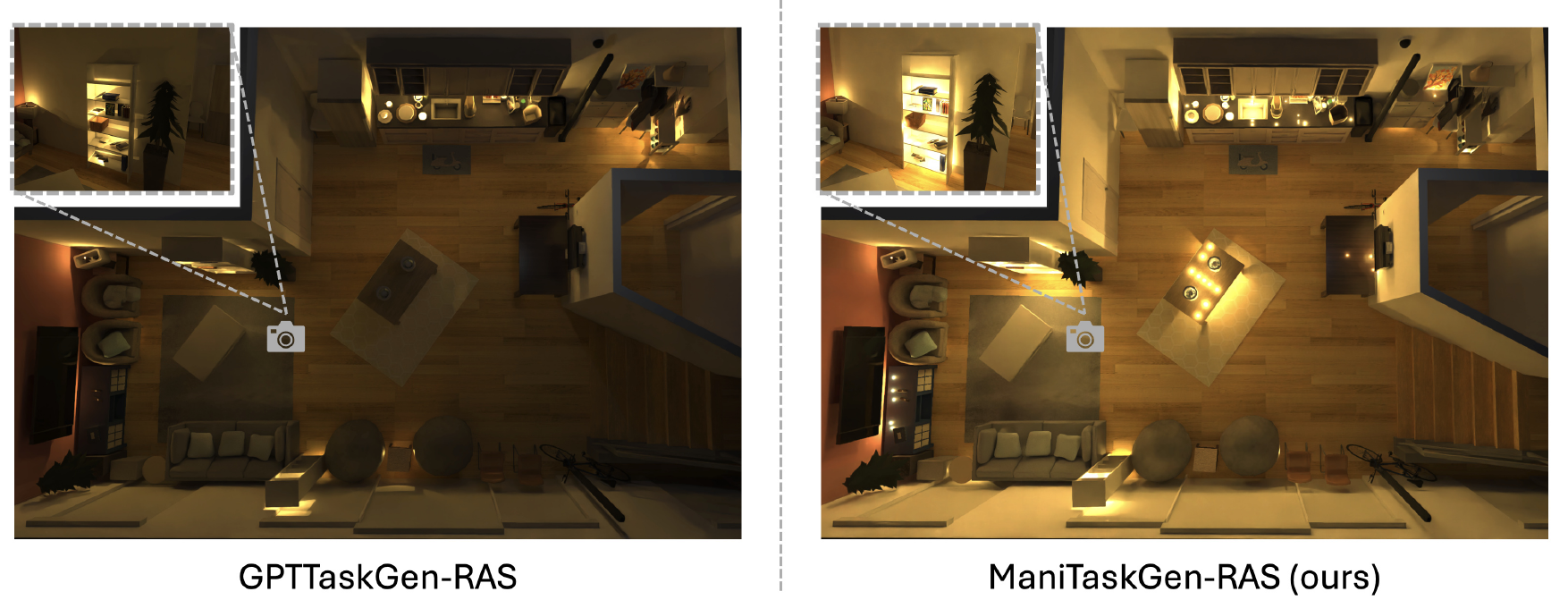}
    \caption{\textbf{The "lightmap"s Which Show the Diversity of Generated Tasks. } Each time an object or location is mentioned in a task, we add a highlight at the corresponding position. The brightness intensity and distribution reveal that our method generates tasks covering more objects and locations.} 
    \vspace{-5mm}
    \label{fig:task_diversity_heatmap}
\end{figure*}

\label{sec:eval}
To evaluate ManiTaskGen in generating comprehensive tasks for varying scenes, we apply it to both simulated (ReplicaCAD~\cite{szot2021habitat}, AI2THOR~\cite{ai2thor}) and real-world scene datasets (SUN-RGBD~\cite{song2015sun}).

\begin{wraptable}{r}{0.48\textwidth} 
    \centering
    \renewcommand{\arraystretch}{1.2}
    \setlength{\tabcolsep}{3pt} 
    \resizebox{\linewidth}{!}{
    \begin{tabular}{lcc}
        \toprule
        \textbf{Datasets} & \textbf{Num. of Scenes} & \textbf{Num. of Tasks} \\
        \midrule

        GenSim~\cite{wang2023gensim} & - & 100 \\
        $\lambda$~\cite{jaafar2025lambdabenchmarkdataefficiencylonghorizon} & 20 & 521 \\
        M3Bench~\cite{zhang2024m3bench} & 119 & 31,050 \\
        ALFRED~\cite{shridhar2020alfred} & 120 & 25,743 \\
        Language Rearrangement ~\cite{szot2023large} & 1 & 1,000 \\
        Embodied Agent Interface ~\cite{li2025embodied} & 2 & 438 \\
        EmbodiedBench~\cite{yang2025embodiedbench} & 4 & 1,128 \\
        \midrule
        \ourgentask{} (Ours) & 3 & $ 39,871$ \\
         ManiTaskGen-RAS (Ours) & 100 & $ \approx 3.96 \times 10^9$ \\
        \bottomrule
    \end{tabular}%
    }
    \vspace{-3mm}
    \caption{\textbf{Comparison among Scene-Task Dataset Statistics.}}
    \label{tab:task_num}
\end{wraptable}
These datasets provide scene information including object poses and bounding boxes, with ReplicaCAD and AI2THOR covering additional object mesh models which enable generating tasks w.r.t object interior surfaces. For this evaluation, we selected 100 scenes from the three datasets and used ManiTaskGen to generate a dataset of tasks, named as ManiTaskGen-RAS. 
Note that ManiTaskGen's generative process supports arbitrarily complex and length of process-based tasks by sampling from multiple interaction cycles, and instantiating outcome-based task templates with diverse objects also generates a vast number of tasks. Thus, the potential size of the task space is theoretically infinite. Yet, for the purpose of statistical analysis, we selected one representative scene from each dataset and curated a finite subset of these tasks, named as \ourgentask{}, comprising a total of 39,871 tasks.
Specifically, for the process-based tasks, we include 39,221 instances comprising single-step pick-and-place tasks (sampled from a single interaction cycle) and two-step tasks (sampled from two consecutive cycles and connected with the logical connector THEN). For the outcome-based tasks, we generated them by instantiating templates part from \ourdataset and employed an ensemble of VLMs (GPT-4o~\cite{openai2024gpt4technicalreport}, Gemini-2.5-pro~\cite{geminiteam2024geminifamilyhighlycapable}, Claude-3.7-sonnet~\cite{anthropic2024aboutclaude}) to vote on their feasibility. This filtering process resulted in a final set of 650 outcome-based task instructions. More details of ManiTaskGen-RAS and \ourgentask{} are provided in Appendix~\ref{appendix_subsec:manitaskgen_ras}.

We compare our generated datasets with existing embodied decision-making datasets in Tab.~\ref{tab:task_num}. 
Notably, using even only 3 scenes, \ourgentask{} contains significantly more tasks than other datasets. Furthermore, our method is scene-agnostic, meaning it can be applied to any given scene, allowing for the incorporation of additional scene data sources to further expand the task set.

Next, we present further results of evaluating the quality of generated tasks from two key aspects : \textbf{Validity} and \textbf{Diversity}.
For a fair comparison, we implement a GPT-based task generation approach as a baseline, referred to as GPTTaskGen. Specifically, we feed each scene’s object information along with their images to GPT-4o~\cite{openai2024gpt4technicalreport}, instructing it to generate the tasks. We apply this baseline method to the same 3 scenes as in \ourgentask{} to generate 10,000 process-based tasks and 1,000 outcome-based tasks, referred to as GPTTaskGen-RAS. This task set serves as a direct comparison to evaluate the validity and diversity of the tasks produced by our method.

\begin{wraptable}{r}{0.48\textwidth}
    \centering
    \renewcommand{\arraystretch}{1.2}
    \setlength{\tabcolsep}{7pt} 
    \resizebox{\linewidth}{!}{
    \begin{tabular}{lcccc} 
        \toprule
        {} & \textbf{Process-based Tasks} & \textbf{Outcome-based Tasks}  \\
        \midrule
        GPTTaskGen-RAS & 29.4\% & 21.1 \%   \\
        \ourgentask{} (Ours)  & 94.0\%  & 86.5 \%   \\
        \bottomrule
    \end{tabular}%
    }
    \vspace{-3mm}
    \caption{\textbf{Human-Verified Task Validity Rate.}}
    \label{tab:task_val}
\end{wraptable}

\label{subsec:task_diversity}
\noindent\textbf{Validity Assessment.} 
We evaluate the validity of the generated tasks by conducting human verification on \ourgentask{} and GPTTaskGen-RAS, with the results reported in Tab.~\ref{tab:task_val}.
For process-based tasks, although our algorithm ensures the target location has sufficient space to accommodate the moved object, some tasks may still be infeasible due to occlusions or obstacles, making the target position difficult to reach or observe. Nevertheless, our results show that most tasks are valid, with significantly higher validity rates compared to the baseline.
For outcome-based tasks, the validity rate is expectedly lower, as it relies on a VLM-based filtering mechanism. Yet our method produces mostly valid tasks.

\noindent\textbf{Diversity Assessment.} 
We proceed to assess the diversity of the generated task set, which reflects how well the tasks cover various scenarios.
To compare task diversity, we randomly sample 100 tasks for the same apartment scene of ReplicaCAD from \ourgentask{} and GPTTaskGen-RAS, and count their object and location coverings. Fig.~\ref{fig:task_diversity_heatmap} presents two "lightmap"s that visualize the distribution of involved objects and locations. Specifically, we light up the centroid of an object or location whenever it appears in a task.
The figure clearly demonstrates that our method generates a wider range of tasks, covering more diverse objects and locations compared to the baseline. In addition, we provide quantitative evaluation results in Appendix~\ref{appendix_subsubsec:semantic_coverage}.

\begin{figure*}[t]
    \centering
    \includegraphics[width=\linewidth]{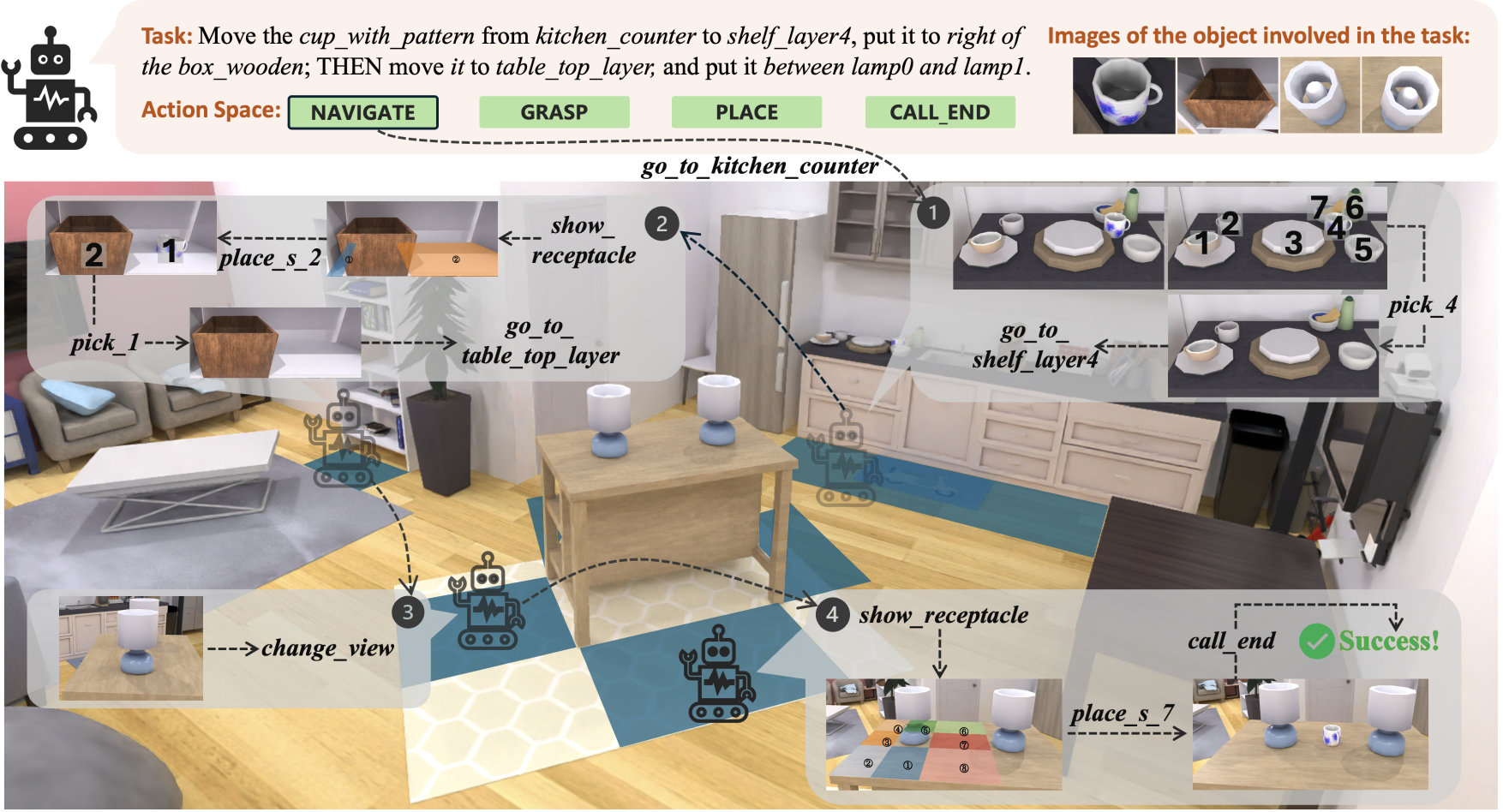}
    \vspace{-5mm}
    \caption{\textbf{Visualization of An Example Testing Episode.} The agent is equipped with abstracted navigation (\texttt{go\_to, change\_view}), grasping (\texttt{pick}) and placing (\texttt{show\_receptacle, place}) skills. Blue marks indicate the walkable locations around ground objects involved in this episode.} 
     \vspace{-5mm}
    \label{fig:testflow}
\end{figure*}

\section{Benchmarking and Improving Embodied Decision-Making}
\label{sec:benchmark_and_improve}
In this section, we showcase two important applications of ManiTaskGen: benchmarking and improving embodied decision-making agents. In Sec.~\ref{subsec:benchmark_vlm}, we design an automatic framework for constructing benchmarks using tasks generated by ManiTaskGen and evaluate the decision-making capabilities of existing VLM agents in both simulated and real-robot experiments. In Sec.~\ref{subsec:improve_vlm}, we further propose an inference-time improvement method that utilizes ManiTaskGen tasks and validate its effectiveness through experimental results.

\subsection{Benchmarking VLM Agents}
\label{subsec:benchmark_vlm}
We evaluate the high-level embodied decision-making capabilities of VLM agents using tasks generated by ManiTaskGen. We first construct a benchmark in simulation~\cite{xiang2020sapien}, where agents receive observations and interact with the environment through an abstracted action space. We introduce the action space and test flow in Sec.~\ref{subsubsec:benchmark_actionspace_and_testflow}, define the task difficulty levels and evaluation criteria in Sec.~\ref{subsubsec:benchmark_task_and_eval}, and present simulation benchmark results in Sec.~\ref{subsubsec:benchmark_results}. We further evaluate high-level decision-making in a physical scene with teleoperated low-level execution in Sec.~\ref{subsubsec:real_robot}.


\subsubsection{Action Space and Test Flow}
\label{subsubsec:benchmark_actionspace_and_testflow}

\noindent\textbf{Action Space.}
To evaluate high-level decision-making, we define a discrete, abstracted action space that the VLM-based agent selects from at each timestep. The scene graph $\mathbb{S}$ introduced in Sec.~\ref{subsec:scene_graph} is used to calculate feasible walkable areas around each ground object, which define possible navigation targets, and also to identify platforms within objects and segment them into receptacle regions. The detailed action space includes:
\begin{itemize}
    \item \texttt{go\_to(platform\_id)}: Navigate to a walkable area around a specific platform of an object.
    \item \texttt{change\_view}: Re-navigate to a different walkable area (if any) of the current platform.
    \item \texttt{pick(object\_id)}: Grasp the object with the specified ID. The \texttt{object\_id}s are associated with the tags provided as the visual observation when the agent is located near a platform.
    \item \texttt{show\_receptacle(object\_id)}: Visualize feasible receptacle regions (tagged by \texttt{receptacle\_ID}s) associated with a specified object on the current platform.
    \item \texttt{place}: Place the held object at a specified location. We offer two placement modes:
        \begin{itemize}
            \item \texttt{place\_r}: Place the object at a random feasible receptacle on the current platform.
            \item \texttt{place\_s([object\_id:receptacle\_id])}: Place the object within the space defined by the list of receptacle regions associated with specified objects on current platform. If multiple receptacles are provided, the system merges them into a larger placement region if they are connected or overlapped.
        \end{itemize}
    \item \texttt{call\_end}: Signals the agent's intention to complete the task and terminates the episode.
\end{itemize}

\noindent\textbf{Test Flow.}
An example episode of the benchmark test flow is visualized in Fig.~\ref{fig:testflow}. At each timestep, the agent receives multimodal observations from the system, including rendered images (with tags for objects or, after \texttt{show\_receptacle}, for receptacles), and text information. The system executes the given actions, updates the environment, and provides the next observation. An episode terminates when the agent executes \texttt{call\_end} or a preset timestep limit is reached. After termination, the episode is evaluated following the criteria described in the next section.

\subsubsection{Task Difficulty and Evaluation Criteria}
\label{subsubsec:benchmark_task_and_eval}

\noindent\textbf{Task Difficulty Levels.}
We classify ManiTaskGen tasks into four difficulty levels. Levels 1 to 3 are process-based tasks, with increasing structural and perceptual complexity. Level 1 comprises single-step pick-and-place tasks involving unique target objects. Level 2 introduces perceptual ambiguity with non-unique target objects requiring disambiguation and additional description, such as moving a red cup when multiple cups are present on the same platform. Level 3 consists of two sub-tasks formed by chaining atomic actions (two one-step pick-and-place tasks from Level 1 or 2) using the logical connector THEN. Level 4 includes all outcome-based tasks.

\noindent\textbf{Evaluation Criteria.}
For process-based tasks (Levels 1 to 3), the expected final scene state is precisely defined, enabling automatic success verification by comparing the initial and final scene graphs. For each testing episode, we conclude following evaluation metrics:
(1) \textbf{Success Rate (SR)}; (2) \textbf{Intermediate Points (IP)}.
For Level 1 \& 2 tasks, a successful episode should include the following four sub-steps, each contributing 25 points of IP:
Navigate to the correct starting location;
Grasp the correct object;
Navigate to the correct destination with the right object;
Place the right object in the correct place.
IP for Level 3 tasks is computed by averaging points from the two sequential sub-tasks. 
Regarding Level 4 tasks, which involve abstract descriptions of scene state changes (e.g., "make the desk cleaner"), defining an unbiased and precise success-state scene graph for automatic evaluation is challenging. To address this, we leverage both human annotators and VLM-based agents to give subjective evaluations by comparing pre- and post-execution scene images. The results are provided in Tab.~\ref{tab:vlm_results} and Appendix~\ref{appendix_subsec:evaluation_discrepancy}.

\begin{table*}[t]
\centering
\crtablespacing
\setlength{\tabcolsep}{10pt} 
\scriptsize
\resizebox{\textwidth}{!}{
\begin{tabular}{@{}lccccccccc@{}}
\toprule
\multicolumn{1}{c}{\multirow{2}{*}{}} & \multicolumn{2}{c}{\textbf{Level 1}} & \multicolumn{2}{c}{\textbf{Level 2}} & \multicolumn{2}{c}{\textbf{Level 3}} & \multicolumn{1}{c}{\textbf{Level 4}} & \multicolumn{2}{c}{\textbf{Average}} \\ \cmidrule(l){2-10} 
\multicolumn{1}{c}{}                                                           & IP           & SR (\%)          & IP           & SR (\%)          & IP           & SR (\%)  & SR(\%)        & IP            & SR (\%)       \\ \midrule
Human                                                                          & 96.0           & 82.5          & 95.0           & 80.5          & 95.0           & 80.0    & 90.0       & 95.3             & 83.3             \\
Random                                                                         & 0.7          & 0.0           & 0.7          & 0.0           & 0.7          & 0.0      & 0.0     & 0.7              & 0.0             \\ \midrule
GPT-5.2\cite{openai2024gpt4technicalreport}                               & 88.1          & 63.9        & \underline{\textbf{67.9}}          &  34.7         & 72.8         &  20.0    & 30.0     & \underline{\textbf{76.3}}          &  37.2          \\
GPT-4o\cite{openai2024gpt4technicalreport}                                    & 71.6           & 40.1          & 42.9          & 16.3          &  57.0         &  8.2    & 26.0  & 57.2            &  22.7          \\
GPT-4.1\cite{openai2024gpt4technicalreport}                               & 67.5           & 36.4          & 46.4           & 20.4          & 54.4           &  4.2     & 28.0    &56.1           &  22.3           \\

GPT-4.1-mini \cite{openai2024gpt4technicalreport}               & 67.0           & 34.0          & 45.9           & 22.4         & 45.3          &  0.0     & 17.0    &  52.7             &  18.4          \\
Gemini-3.1-pro \cite{geminiteam2024geminifamilyhighlycapable}                 & 87.3           & 65.1          & 54.6        & 24.5         & \underline{\textbf{75.3}}          & \underline{\textbf{22.2}}    &33.0     & 72.4              &  36.2            \\
Gemini-2.5-pro \cite{geminiteam2024geminifamilyhighlycapable}                 & 82.3         & 51.5          & 54.1          & 22.4          & 68.9          &  13.3   & 27.0      & 68.4              & 28.6            \\
Gemini-2.5-flash\cite{geminiteam2024geminifamilyhighlycapable}                & 75.0           & 40.3         & 40.8           & 14.3         &  56.0          &  4.1   & 25.0      & 57.3              &  20.9           \\

Claude-4.5-sonnet \cite{anthropic2024aboutclaude}                             & \underline{\textbf{88.5}}         & \underline{\textbf{69.3}}      & 64.9           & \underline{\textbf{36.7}}          & 71.4        &  20.0   & \underline{\textbf{35.0}}     &      74.9        &   \underline{\textbf{40.3}}         \\

Claude-3.7-sonnet \cite{anthropic2024aboutclaude}                             & 73.9           & 45.7         & 55.6           & 28.6          & 54.7         &   8.9   & 28.0    &      61.4       &    27.8       \\

Claude-4.5-haiku \cite{anthropic2024aboutclaude}                              & 82.6          & 55.4       & 54.1        & 16.3        & 53.6         &  6.7 & 27.0       &  63.4           &   26.4          \\ \hdashline
Qwen-3-VL-235B-A22B-instruct \cite{wang2024qwen2}                                                                     & 85.0         & 55.0         &53.1          & 14.3         & 65.3          &   13.3 & 28.0     & 67.8            &   27.7         \\
Qwen-2.5-VL-72B-Ins \cite{wang2024qwen2}                                                                     & 55.1           & 28.2          & 37.2           & 8.2          & 43.5          &    4.4  & 24.0     &  45.3            &   16.2          \\

\end{tabular}
}
\caption{\textbf{Evaluation Results on Existing VLMs.} We cover both proprietary (upper part) and open-source models (lower part). Here, IP refers to Intermediate Points, and SR refers to Success Rate. }
\label{tab:vlm_results}
\end{table*}

\subsubsection{Benchmarking Results on VLM Agents}
\label{subsubsec:benchmark_results}
We evaluate existing VLM-based embodied agents in the simulated scenes~\cite{szot2021habitat, ai2thor} from \ourgentask. Specifically, we randomly sample 1000 tasks for each of the level 1-3 tasks and 100 level 4 tasks. A timestep limit of 20 is set for Level 1 \& 2 tasks, and 40 for Level 3 \& 4 tasks. Results are shown in Tab.~\ref{tab:vlm_results}.
We observe that all models achieve a low average SR (roughly under 40\%), significantly below human-level performance. Performance generally decreases with increasing task difficulty. Longer horizon tasks (Level 3 \& 4) lead to significantly lower performance compared to single-step tasks, highlighting the substantial challenge posed by the generated tasks. 
\begin{wrapfigure}{r}{0.48\textwidth} 
    \vspace{-3mm}
    \centering
    \includegraphics[width=\linewidth, keepaspectratio]{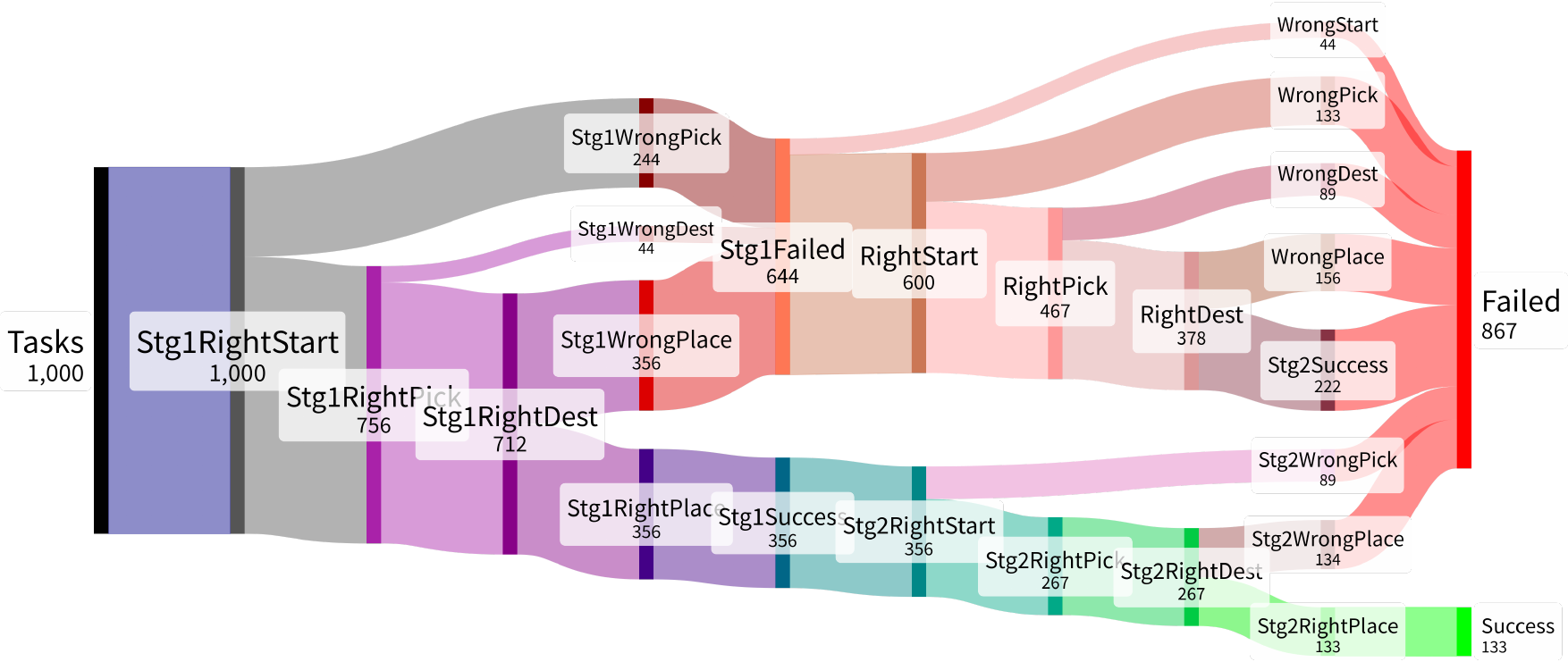}
    \vspace{-3mm}
    \caption{\textbf{Success and Failure Modes} from Gemini-2.5-pro in each stage of Level 3 tasks.}
    \label{fig:sankey_diagram}
\end{wrapfigure}
To further analyze failure cases of the embodied agents, in Fig.~\ref{fig:sankey_diagram}, we visualize the sub-step mistake distribution from Gemini-2.5-pro~\cite{geminiteam2024geminifamilyhighlycapable} for Level 3 tasks. We observe that wrong placement accounts for the largest proportion of mistakes, suggesting that VLMs' spatial understanding capabilities may serve as a primary bottleneck for mobile manipulation tasks. Furthermore, analysis of execution traces reveals that while the agent performs well in navigation in early stages of the episode (e.g., navigating to the first target object), it makes more navigation mistakes in later steps, indicating that longer-horizon tasks pose a greater challenge to the VLM agent's decision-making capabilities.

\subsubsection{Teleoperated Real-Robot Evaluation}
\label{subsubsec:real_robot}
\begin{figure}[t]
  \centering
  \includegraphics[width=0.8\linewidth]{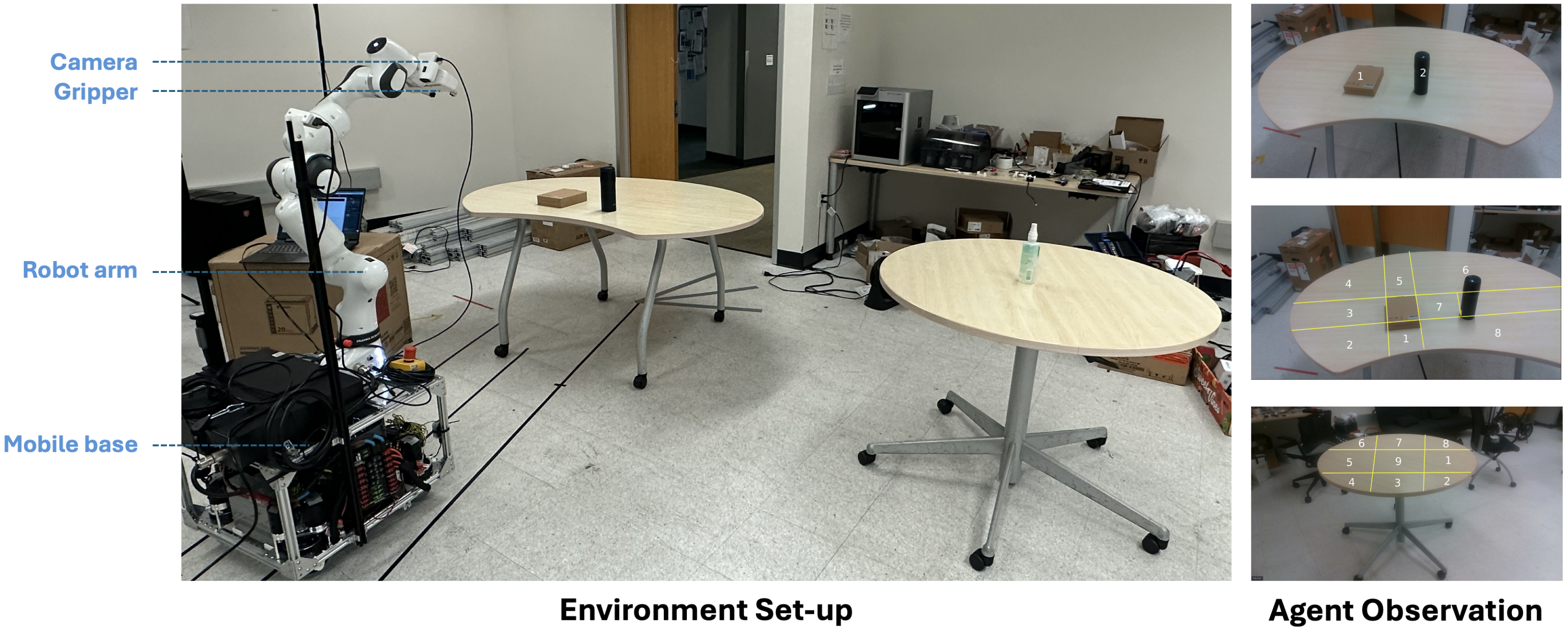}
  \caption{\textbf{Set-up of Real Robot Experiments.} 
  The camera view was manually annotated with numerical tags included in the agent's visual input. All irrelevant objects are masked out.
  }
  \label{fig:real_robot}
\end{figure}
\begin{wraptable}{r}{0.48\textwidth} 
    \centering
    \crtablespacing
    \renewcommand{\arraystretch}{1.2}
    \setlength{\tabcolsep}{3pt} 
    \resizebox{\linewidth}{!}{
    \begin{tabular}{lcccc}
        \toprule
        \textbf{Model Name} & \textbf{IP (Level1\&2)} & \textbf{SR (Level1\&2)} & \textbf{IP (Level3)} & \textbf{SR (Level3)} \\
        \midrule
       Gemini-2.5-pro\cite{geminiteam2024geminifamilyhighlycapable} & 97.5 & 90\% & 100 & 100\% \\
        \bottomrule
    \end{tabular}%
    }
    \caption{\textbf{Real Robot Experiment Results.} 
    }
    \label{tab:real_results}
\end{wraptable}

In this section, we demonstrate using a physical robot for real world task generation and evaluation. The setup is shown in Fig.~\ref{fig:real_robot}. First, we constructed a scene comprising two tables and three surface objects. We then used reconstruction techniques~\cite{PolyCam} to obtain the coordinates, pose, and bounding boxes for each object, and fed them into ManiTaskGen. All irrelevant objects are ignored during the reconstruction process. We generated 10 tasks for Level 1\&2, and 10 tasks for Level 3. Next, we deployed Gemini-2.5-pro~\cite{geminiteam2024geminifamilyhighlycapable} on a physical mobile-base robot for evaluation. As established in Sec.~\ref{subsubsec:benchmark_actionspace_and_testflow}, the VLM's output remains high-level decisions (abstracted actions). Given the current lack of a sufficiently generalizable grasping and placing policy for real robots, the execution of the actions was performed via human teleoperation. The results are shown in Tab.~\ref{tab:real_results}. The agent achieved a good performance due to the relatively simple setting, yet ManiTaskGen is readily applicable to more complex settings. For future work, we plan to integrate automated scene reconstruction, object detection and annotation techniques to further reduce the labor cost associated with real-world evaluation using physical robots.

\subsection{Improving VLM Agents}
\label{subsec:improve_vlm}
Beyond benchmarking, ManiTaskGen provides task-level and intermediate-step feedback, which can be useful for improving VLM-based agents. 
We demonstrate this application through an inference-time self-reflection method inspired by Reflexion~\cite{shinn2023reflexion}. 
Specifically, we design a self-reflection model that processes the evaluation results of each episode to generate a verbal summary. This summary is then stored in a long-term memory and used as part of the input for the agent when attempting the task in future episodes. We present details of our optimization process in Sec.~\ref{subsubsec:improvement_details}, and experimental results in Sec.~\ref{subsubsec:improvement_results}.


\subsubsection{Inference-time Improvement through Self-Reflection}
\label{subsubsec:improvement_details}
Given a trial episode and its evaluation results (including final success judgment and intermediate step evaluations), a rule-based self-reflection module automatically generates the following summary:
\begin{itemize}
    \item The task goal, the history of the trial episode, and whether the task was ultimately successful.
    \item If the task was not successful, a summary indicating which intermediate step goals were achieved and the point in the action history up to which they were successfully completed. As described in Sec.~\ref{subsubsec:benchmark_task_and_eval}, these correct intermediate steps include: navigate to the starting location, grasp the object, navigate to the destination location, and place the object .
    \item For the intermediate goals underachieved, what are the next suggested actions.
    \item Corresponding observation images in each part of the summary.
\end{itemize}
We maintain a long-term memory to store these verbal summaries from multiple trial episodes. The optimization process is iterative. After each trial, the corresponding self-reflection updates the long-term memory, which is used as part of the input for the agent when attempting subsequent trials.

\subsubsection{Improvement Results}
\label{subsubsec:improvement_results}
\begin{table}[htbp]
    \centering
    \crtablespacing
    \small
    \renewcommand{\arraystretch}{1.0}
    \setlength{\tabcolsep}{3pt}
    \begin{tabular}{lcccc}
        \toprule
        \textbf{Model Name} & \textbf{IP (Before)} & \textbf{IP (After)} & \textbf{SR (Before)} & \textbf{SR (After)} \\
        \midrule

       GPT-4.1\cite{openai2024gpt4technicalreport} & 75.5 & \textbf{79.4} & 38\% & \textbf{49\%} \\
       Gemini-2.5-flash\cite{geminiteam2024geminifamilyhighlycapable} & 79.4 & \textbf{81.7} & 36\% & \textbf{58\%} \\
       Claude-3.5-haiku \cite{anthropic2024aboutclaude} & 68.9 & \textbf{79.4} & 36\% & \textbf{51\%} \\
        \bottomrule
    \end{tabular}
    \caption{\textbf{Improvement of Agent Performance.} 
    }
    \label{tab:improve_results}
\end{table}
\begin{wrapfigure}{r}{0.48\textwidth}
  \centering
  \includegraphics[width=1.0\linewidth]{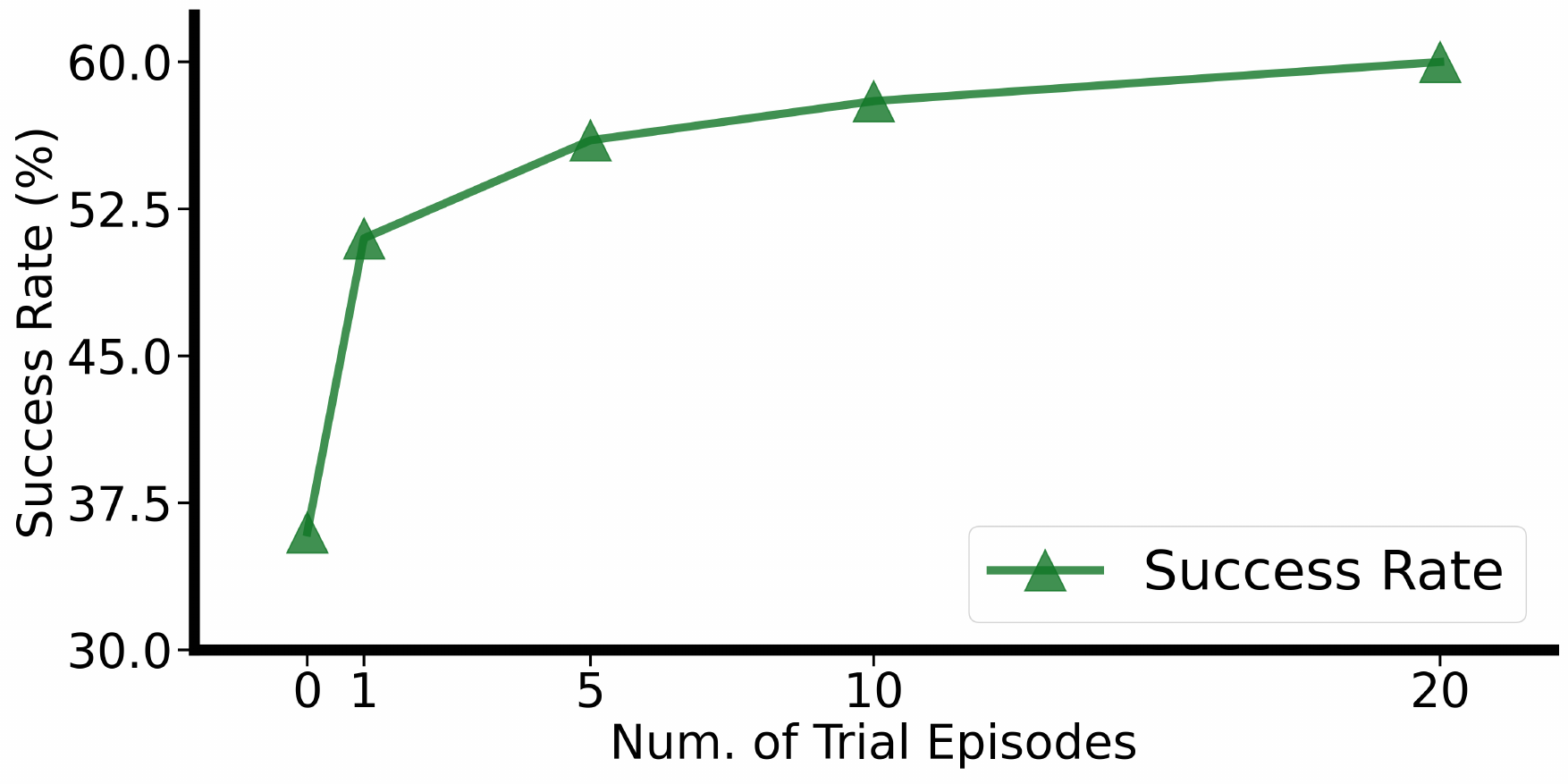}
  \vspace{-8mm}
  \caption{\textbf{Improvement of Gemini-2.5-flash Agent.} We find that more trial episodes continuously contribute to the agent performance.}
  \label{fig:improve_scaling}
\end{wrapfigure}
We construct reflection memory using trial tasks from the ReplicaCAD~\cite{szot2021habitat} scene of ManiTaskGen-RAS and evaluate the resulting agent on tasks from the AI2THOR~\cite{ai2thor} scene. Tab.~\ref{tab:improve_results} compares agent performance with and without the reflection memory. We use 10 trial episodes to construct the reflection memory and evaluate the agent on 100 randomly sampled Level-1 test tasks. As shown in the table, our optimization significantly improved both the task execution success rate and the Intermediate Points score. We further investigated the impact of the number of trial episodes on final agent performance. Results are presented in Fig. ~\ref{fig:improve_scaling}. Preliminary results show that agent performance continuously improves with an increasing number of self-reflection trial episodes. Similar findings have been observed in previous work~\cite{shinn2023reflexion}. Therefore, these findings further underscore the significance of ManiTaskGen. Because it can generate and evaluate tasks in arbitrary scenes, thus it provides abundant resources for agent improvement.

\section{Conclusion and Future Work}
\label{sec:conclusion}
We introduce ManiTaskGen for systematic, scene-grounded task generation under a navigate--pick--place rearrangement abstraction. ManiTaskGen can generate a comprehensive set of long-horizon mobile manipulation tasks, covering both process-based and outcome-based tasks, thereby providing a diverse set of testing scenarios and improving resources for embodied decision-making agents.
Our experiments demonstrate the validity and diversity of the generated tasks, while also showcasing their practical usability by benchmarking and improving the decision-making capabilities of existing VLM-based agents.

\noindent\textbf{Limitations and future work.}
Currently, ManiTaskGen has focused on embodied decision-making agents with navigate, grasp, and place capabilities.  Future work will extend to a broader skill set, including fine-grained object manipulation and tasks involving non-pose state changes.

\clearpage

\begin{ack}

\textbf{Funding.} This research received no external funding.

\textbf{Competing Interests.} The authors declare no competing interests.

\end{ack}

\bibliographystyle{unsrtnat}
\bibliography{reference}

@inproceedings{li2023behavior,
  title={Behavior-1k: A benchmark for embodied ai with 1,000 everyday activities and realistic simulation},
  author={Li, Chengshu and Zhang, Ruohan and Wong, Josiah and Gokmen, Cem and Srivastava, Sanjana and Mart{\'\i}n-Mart{\'\i}n, Roberto and Wang, Chen and Levine, Gabrael and Lingelbach, Michael and Sun, Jiankai and others},
  booktitle={Conference on Robot Learning},
  pages={80--93},
  year={2023},
  organization={PMLR}
}

@Misc{amazonturk,
  author       = {Amazon},
  howpublished = {\url{https://www.mturk.com/}},
  title        = {Amazon Mechanical Turk},
  year = {2024},
}

@inproceedings{szot2021habitat,
    title     =     {Habitat 2.0: Training Home Assistants to Rearrange their Habitat},
    author    =     {Andrew Szot and Alex Clegg and Eric Undersander and Erik Wijmans and Yili Zhao and John Turner and Noah Maestre and Mustafa Mukadam and Devendra Chaplot and Oleksandr Maksymets and Aaron Gokaslan and Vladimir Vondrus and Sameer Dharur and Franziska Meier and Wojciech Galuba and Angel Chang and Zsolt Kira and Vladlen Koltun and Jitendra Malik and Manolis Savva and Dhruv Batra},
    booktitle   =     {Advances in Neural Information Processing Systems (NeurIPS)},
    year      =     {2021}
}

@article{fang2020adaptive,
  title={Adaptive procedural task generation for hard-exploration problems},
  author={Fang, Kuan and Zhu, Yuke and Savarese, Silvio and Fei-Fei, Li},
  journal={arXiv preprint arXiv:2007.00350},
  year={2020}
}

@article{wang2023gensim,
  title={Gensim: Generating robotic simulation tasks via large language models},
  author={Wang, Lirui and Ling, Yiyang and Yuan, Zhecheng and Shridhar, Mohit and Bao, Chen and Qin, Yuzhe and Wang, Bailin and Xu, Huazhe and Wang, Xiaolong},
  journal={arXiv preprint arXiv:2310.01361},
  year={2023}
}

@misc{jaafar2025lambdabenchmarkdataefficiencylonghorizon,
      title={{$\lambda$}: A Benchmark for Data-Efficiency in Long-Horizon Indoor Mobile Manipulation Robotics}, 
      author={Ahmed Jaafar and Shreyas Sundara Raman and Yichen Wei and Sudarshan Harithas and Sofia Juliani and Anneke Wernerfelt and Benedict Quartey and Ifrah Idrees and Jason Xinyu Liu and Stefanie Tellex},
      year={2025},
      eprint={2412.05313},
      archivePrefix={arXiv},
      primaryClass={cs.RO},
      url={https://arxiv.org/abs/2412.05313}, 
}

@article{zhang2024m3bench,
  title={M3Bench: Benchmarking Whole-body Motion Generation for Mobile Manipulation in 3D Scenes},
  author={Zhang, Zeyu and Yan, Sixu and Han, Muzhi and Wang, Zaijin and Wang, Xinggang and Zhu, Song-Chun and Liu, Hangxin},
  journal={arXiv preprint arXiv:2410.06678},
  year={2024}
}

@misc{habitatrearrangechallenge2022,
  title={Habitat Rearrangement Challenge 2022},
  author={Szot, Andrew and Yadav, Karmesh and Clegg, Alex and Berges, Vincent-Pierre and Gokaslan, Aaron and Chang, Angel and Savva, Manolis and Kira, Zsolt and Batra, Dhruv},
  howpublished={\url{https://aihabitat.org/challenge/2022_rearrange}},
  year={2022}
}

@article{gu2022multi,
  title={Multi-skill mobile manipulation for object rearrangement},
  author={Gu, Jiayuan and Chaplot, Devendra Singh and Su, Hao and Malik, Jitendra},
  journal={arXiv preprint arXiv:2209.02778},
  year={2022}
}

@article{batra2020rearrangement,
author = {Batra, Dhruv and Chang, Angel and Chernova, Sonia and Davison, Andrew and Deng, Jia and Koltun, Vladlen and Levine, Sergey and Malik, Jitendra and Mordatch, Igor and Mottaghi, Roozbeh and Savva, Manolis and Su, Hao},
title = {{R}earrangement: {A} {C}hallenge for {E}mbodied {A}{I}},
journal = {arXiv preprint},
year = {2020}}

@article{liu2024visualagentbench,
  title={Visualagentbench: Towards large multimodal models as visual foundation agents},
  author={Liu, Xiao and Zhang, Tianjie and Gu, Yu and Iong, Iat Long and Xu, Yifan and Song, Xixuan and Zhang, Shudan and Lai, Hanyu and Liu, Xinyi and Zhao, Hanlin and others},
  journal={arXiv preprint arXiv:2408.06327},
  year={2024}
}

@inproceedings{shridhar2020alfred,
  title={Alfred: A benchmark for interpreting grounded instructions for everyday tasks},
  author={Shridhar, Mohit and Thomason, Jesse and Gordon, Daniel and Bisk, Yonatan and Han, Winson and Mottaghi, Roozbeh and Zettlemoyer, Luke and Fox, Dieter},
  booktitle={Proceedings of the IEEE/CVF conference on computer vision and pattern recognition},
  pages={10740--10749},
  year={2020}
}

@inproceedings{cobbe2020leveraging,
  title={Leveraging procedural generation to benchmark reinforcement learning},
  author={Cobbe, Karl and Hesse, Chris and Hilton, Jacob and Schulman, John},
  booktitle={International conference on machine learning},
  pages={2048--2056},
  year={2020},
  organization={PMLR}
}

@article{raileanu2020ride,
  title={Ride: Rewarding impact-driven exploration for procedurally-generated environments},
  author={Raileanu, Roberta and Rockt{\"a}schel, Tim},
  journal={arXiv preprint arXiv:2002.12292},
  year={2020}
}

@article{hu2024agentgen,
  title={Agentgen: Enhancing planning abilities for large language model based agent via environment and task generation},
  author={Hu, Mengkang and Zhao, Pu and Xu, Can and Sun, Qingfeng and Lou, Jianguang and Lin, Qingwei and Luo, Ping and Rajmohan, Saravan and Zhang, Dongmei},
  journal={arXiv preprint arXiv:2408.00764},
  year={2024}
}

@inproceedings{xiang2020sapien,
  title={Sapien: A simulated part-based interactive environment},
  author={Xiang, Fanbo and Qin, Yuzhe and Mo, Kaichun and Xia, Yikuan and Zhu, Hao and Liu, Fangchen and Liu, Minghua and Jiang, Hanxiao and Yuan, Yifu and Wang, He and others},
  booktitle={Proceedings of the IEEE/CVF conference on computer vision and pattern recognition},
  pages={11097--11107},
  year={2020}
}

@misc{geminiteam2024geminifamilyhighlycapable,
      title={Gemini: A Family of Highly Capable Multimodal Models}, 
      author={Gemini Team and Rohan Anil and Sebastian Borgeaud and Jean-Baptiste Alayrac and Jiahui Yu and Radu Soricut and Johan Schalkwyk and Andrew M. Dai and Anja Hauth, et al.},
      year={2024},
      eprint={2312.11805},
      archivePrefix={arXiv},
      primaryClass={cs.CL},
      url={https://arxiv.org/abs/2312.11805}, 
}

@misc{openai2024gpt4technicalreport,
      title={GPT-4 Technical Report}, 
      author={Josh Achiam and Steven Adler and Sandhini Agarwal and Lama Ahmad and Ilge Akkaya and Florencia Leoni Aleman and Diogo Almeida and Janko Altenschmidt and Sam Altman and Shyamal Anadkat et al.},
      year={2024},
      eprint={2303.08774},
      archivePrefix={arXiv},
      primaryClass={cs.CL},
      url={https://arxiv.org/abs/2303.08774}, 
}

@misc{anthropic2024aboutclaude,
  author = {Anthropic},
  title = {About Claude Models},
  howpublished = {\url{https://docs.anthropic.com/en/docs/about-claude/models}},
  year = {2024},
  note = {Accessed: 2024-09-03}
}

@article{li2025embodied,
  title={Embodied agent interface: Benchmarking llms for embodied decision making},
  author={Li, Manling and Zhao, Shiyu and Wang, Qineng and Wang, Kangrui and Zhou, Yu and Srivastava, Sanjana and Gokmen, Cem and Lee, Tony and Li, Erran Li and Zhang, Ruohan and others},
  journal={Advances in Neural Information Processing Systems},
  volume={37},
  pages={100428--100534},
  year={2025}
}

@inproceedings{szot2023large,
  title={Large language models as generalizable policies for embodied tasks},
  author={Szot, Andrew and Schwarzer, Max and Agrawal, Harsh and Mazoure, Bogdan and Metcalf, Rin and Talbott, Walter and Mackraz, Natalie and Hjelm, R Devon and Toshev, Alexander T},
  booktitle={The Twelfth International Conference on Learning Representations},
  year={2023}
}

@article{wang2024qwen2,
  title={Qwen2-vl: Enhancing vision-language model's perception of the world at any resolution},
  author={Wang, Peng and Bai, Shuai and Tan, Sinan and Wang, Shijie and Fan, Zhihao and Bai, Jinze and Chen, Keqin and Liu, Xuejing and Wang, Jialin and Ge, Wenbin and others},
  journal={arXiv preprint arXiv:2409.12191},
  year={2024}
}

@misc{Meta2024Llama3.2,
  author = {Meta},
  title = {Llama 3.2: Revolutionizing Edge AI and Vision with Open, Customizable Models},
  year = {2024},
  url = {https://ai.meta.com/blog/llama-3-2-connect-2024-vision-edge-mobile-devices/},
  note = {Accessed: 2025-02-15}
}

@article{ai2thor,
  author={Eric Kolve and Roozbeh Mottaghi and Winson Han and
          Eli VanderBilt and Luca Weihs and Alvaro Herrasti and
          Daniel Gordon and Yuke Zhu and Abhinav Gupta and
          Ali Farhadi},
  title={{AI2-THOR: An Interactive 3D Environment for Visual AI}},
  journal={arXiv},
  year={2017}
}

@inproceedings{song2015sun,
  title={Sun rgb-d: A rgb-d scene understanding benchmark suite},
  author={Song, Shuran and Lichtenberg, Samuel P and Xiao, Jianxiong},
  booktitle={Proceedings of the IEEE conference on computer vision and pattern recognition},
  pages={567--576},
  year={2015}
}

@inproceedings{dai2017scannet,
  title={Scannet: Richly-annotated 3d reconstructions of indoor scenes},
  author={Dai, Angela and Chang, Angel X and Savva, Manolis and Halber, Maciej and Funkhouser, Thomas and Nie{\ss}ner, Matthias},
  booktitle={Proceedings of the IEEE conference on computer vision and pattern recognition},
  pages={5828--5839},
  year={2017}
}

@article{shinn2023reflexion,
  title={Reflexion: Language agents with verbal reinforcement learning},
  author={Shinn, Noah and Cassano, Federico and Gopinath, Ashwin and Narasimhan, Karthik and Yao, Shunyu},
  journal={Advances in Neural Information Processing Systems},
  volume={36},
  pages={8634--8652},
  year={2023}
}

@inproceedings{yao2023react,
  title={React: Synergizing reasoning and acting in language models},
  author={Yao, Shunyu and Zhao, Jeffrey and Yu, Dian and Du, Nan and Shafran, Izhak and Narasimhan, Karthik and Cao, Yuan},
  booktitle={International Conference on Learning Representations (ICLR)},
  year={2023}
}

@article{black2410pi0,
  title={$\pi$0: A vision-language-action flow model for general robot control, 2024},
  author={Black, Kevin and Brown, Noah and Driess, Danny and Esmail, Adnan and Equi, Michael and Finn, Chelsea and Fusai, Niccolo and Groom, Lachy and Hausman, Karol and Ichter, Brian and others},
  year={2024},
  journal={URL https://arxiv. org/abs/2410.24164}
}

@article{kim2024openvla,
  title={Openvla: An open-source vision-language-action model},
  author={Kim, Moo Jin and Pertsch, Karl and Karamcheti, Siddharth and Xiao, Ted and Balakrishna, Ashwin and Nair, Suraj and Rafailov, Rafael and Foster, Ethan and Lam, Grace and Sanketi, Pannag and others},
  journal={arXiv preprint arXiv:2406.09246},
  year={2024}
}

@article{huang2023embodied,
  title={An embodied generalist agent in 3d world},
  author={Huang, Jiangyong and Yong, Silong and Ma, Xiaojian and Linghu, Xiongkun and Li, Puhao and Wang, Yan and Li, Qing and Zhu, Song-Chun and Jia, Baoxiong and Huang, Siyuan},
  journal={arXiv preprint arXiv:2311.12871},
  year={2023}
}

@article{szot2024multimodal,
  title={From Multimodal LLMs to Generalist Embodied Agents: Methods and Lessons},
  author={Szot, Andrew and Mazoure, Bogdan and Attia, Omar and Timofeev, Aleksei and Agrawal, Harsh and Hjelm, Devon and Gan, Zhe and Kira, Zsolt and Toshev, Alexander},
  journal={arXiv preprint arXiv:2412.08442},
  year={2024}
}

@article{ahn2022can,
  title={Do as i can, not as i say: Grounding language in robotic affordances},
  author={Ahn, Michael and Brohan, Anthony and Brown, Noah and Chebotar, Yevgen and Cortes, Omar and David, Byron and Finn, Chelsea and Fu, Chuyuan and Gopalakrishnan, Keerthana and Hausman, Karol and others},
  journal={arXiv preprint arXiv:2204.01691},
  year={2022}
}

@article{brohan2023rt,
  title={Rt-2: Vision-language-action models transfer web knowledge to robotic control},
  author={Brohan, Anthony and Brown, Noah and Carbajal, Justice and Chebotar, Yevgen and Chen, Xi and Choromanski, Krzysztof and Ding, Tianli and Driess, Danny and Dubey, Avinava and Finn, Chelsea and others},
  journal={arXiv preprint arXiv:2307.15818},
  year={2023}
}

@article{driess2023palm,
  title={Palm-e: An embodied multimodal language model},
  author={Driess, Danny and Xia, Fei and Sajjadi, Mehdi SM and Lynch, Corey and Chowdhery, Aakanksha and Wahid, Ayzaan and Tompson, Jonathan and Vuong, Quan and Yu, Tianhe and Huang, Wenlong and others},
journal={arXiv preprint},
  year={2023}
}

@article{team2024gemma,
  title={Gemma: Open models based on gemini research and technology},
  author={Team, Gemma and Mesnard, Thomas and Hardin, Cassidy and Dadashi, Robert and Bhupatiraju, Surya and Pathak, Shreya and Sifre, Laurent and Rivi{\`e}re, Morgane and Kale, Mihir Sanjay and Love, Juliette and others},
  journal={arXiv preprint arXiv:2403.08295},
  year={2024}
}

@article{bai2023qwen,
  title={Qwen technical report},
  author={Bai, Jinze and Bai, Shuai and Chu, Yunfei and Cui, Zeyu and Dang, Kai and Deng, Xiaodong and Fan, Yang and Ge, Wenbin and Han, Yu and Huang, Fei and others},
  journal={arXiv preprint arXiv:2309.16609},
  year={2023}
}

@inproceedings{gao2024physically,
  title={Physically grounded vision-language models for robotic manipulation},
  author={Gao, Jensen and Sarkar, Bidipta and Xia, Fei and Xiao, Ted and Wu, Jiajun and Ichter, Brian and Majumdar, Anirudha and Sadigh, Dorsa},
  booktitle={2024 IEEE International Conference on Robotics and Automation (ICRA)},
  pages={12462--12469},
  year={2024},
  organization={IEEE}
}

@inproceedings{huang2022language,
  title={Language models as zero-shot planners: Extracting actionable knowledge for embodied agents},
  author={Huang, Wenlong and Abbeel, Pieter and Pathak, Deepak and Mordatch, Igor},
  booktitle={International conference on machine learning},
  pages={9118--9147},
  year={2022},
  organization={PMLR}
}

@inproceedings{o2024open,
  title={Open x-embodiment: Robotic learning datasets and rt-x models: Open x-embodiment collaboration 0},
  author={O’Neill, Abby and Rehman, Abdul and Maddukuri, Abhiram and Gupta, Abhishek and Padalkar, Abhishek and Lee, Abraham and Pooley, Acorn and Gupta, Agrim and Mandlekar, Ajay and Jain, Ajinkya and others},
  booktitle={2024 IEEE International Conference on Robotics and Automation (ICRA)},
  pages={6892--6903},
  year={2024},
  organization={IEEE}
}

@article{yang2025embodiedbench,
  title={EmbodiedBench: Comprehensive Benchmarking Multi-modal Large Language Models for Vision-Driven Embodied Agents},
  author={Yang, Rui and Chen, Hanyang and Zhang, Junyu and Zhao, Mark and Qian, Cheng and Wang, Kangrui and Wang, Qineng and Koripella, Teja Venkat and Movahedi, Marziyeh and Li, Manling and others},
  journal={arXiv preprint arXiv:2502.09560},
  year={2025}
}

@article{choi2024lota,
  title={Lota-bench: Benchmarking language-oriented task planners for embodied agents},
  author={Choi, Jae-Woo and Yoon, Youngwoo and Ong, Hyobin and Kim, Jaehong and Jang, Minsu},
  journal={arXiv preprint arXiv:2402.08178},
  year={2024}
}

@article{cheng2025embodiedeval,
  title={EmbodiedEval: Evaluate Multimodal LLMs as Embodied Agents},
  author={Cheng, Zhili and Tu, Yuge and Li, Ran and Dai, Shiqi and Hu, Jinyi and Hu, Shengding and Li, Jiahao and Shi, Yang and Yu, Tianyu and Chen, Weize and others},
  journal={arXiv preprint arXiv:2501.11858},
  year={2025}
}

@article{luong2024reft,
  title={Reft: Reasoning with reinforced fine-tuning},
  author={Luong, Trung Quoc and Zhang, Xinbo and Jie, Zhanming and Sun, Peng and Jin, Xiaoran and Li, Hang},
  journal={arXiv preprint arXiv:2401.08967},
  volume={3},
  year={2024}
}

@article{zhai2024fine,
  title={Fine-tuning large vision-language models as decision-making agents via reinforcement learning},
  author={Zhai, Simon and Bai, Hao and Lin, Zipeng and Pan, Jiayi and Tong, Peter and Zhou, Yifei and Suhr, Alane and Xie, Saining and LeCun, Yann and Ma, Yi and others},
  journal={Advances in neural information processing systems},
  volume={37},
  pages={110935--110971},
  year={2024}
}

@article{mu2021maniskill,
  title={Maniskill: Generalizable manipulation skill benchmark with large-scale demonstrations},
  author={Mu, Tongzhou and Ling, Zhan and Xiang, Fanbo and Yang, Derek and Li, Xuanlin and Tao, Stone and Huang, Zhiao and Jia, Zhiwei and Su, Hao},
  journal={arXiv preprint arXiv:2107.14483},
  year={2021}
}

@article{jin2023mini,
  title={Mini-behavior: A procedurally generated benchmark for long-horizon decision-making in embodied ai},
  author={Jin, Emily and Hu, Jiaheng and Huang, Zhuoyi and Zhang, Ruohan and Wu, Jiajun and Fei-Fei, Li and Mart{\'\i}n-Mart{\'\i}n, Roberto},
  journal={arXiv preprint arXiv:2310.01824},
  year={2023}
}

@inproceedings{hong2024cogagent,
  title={Cogagent: A visual language model for gui agents},
  author={Hong, Wenyi and Wang, Weihan and Lv, Qingsong and Xu, Jiazheng and Yu, Wenmeng and Ji, Junhui and Wang, Yan and Wang, Zihan and Dong, Yuxiao and Ding, Ming and others},
  booktitle={Proceedings of the IEEE/CVF Conference on Computer Vision and Pattern Recognition},
  pages={14281--14290},
  year={2024}
}

@article{ramrakhya2025grounding,
  title={Grounding Multimodal LLMs to Embodied Agents that Ask for Help with Reinforcement Learning},
  author={Ramrakhya, Ram and Chang, Matthew and Puig, Xavier and Desai, Ruta and Kira, Zsolt and Mottaghi, Roozbeh},
  journal={arXiv preprint arXiv:2504.00907},
  year={2025}
}

@article{pan2024autonomous,
  title={Autonomous evaluation and refinement of digital agents},
  author={Pan, Jiayi and Zhang, Yichi and Tomlin, Nicholas and Zhou, Yifei and Levine, Sergey and Suhr, Alane},
  journal={arXiv preprint arXiv:2404.06474},
  year={2024}
}

@article{wang2024rl,
  title={Rl-vlm-f: Reinforcement learning from vision language foundation model feedback},
  author={Wang, Yufei and Sun, Zhanyi and Zhang, Jesse and Xian, Zhou and Biyik, Erdem and Held, David and Erickson, Zackory},
  journal={arXiv preprint arXiv:2402.03681},
  year={2024}
}

@article{szot2024grounding,
  title={Grounding multimodal large language models in actions},
  author={Szot, Andrew and Mazoure, Bogdan and Agrawal, Harsh and Hjelm, R Devon and Kira, Zsolt and Toshev, Alexander},
  journal={Advances in Neural Information Processing Systems},
  volume={37},
  pages={20198--20224},
  year={2024}
}

@article{song2024trial,
  title={Trial and error: Exploration-based trajectory optimization for llm agents},
  author={Song, Yifan and Yin, Da and Yue, Xiang and Huang, Jie and Li, Sujian and Lin, Bill Yuchen},
  journal={arXiv preprint arXiv:2403.02502},
  year={2024}
}

@article{zeng2023agenttuning,
  title={Agenttuning: Enabling generalized agent abilities for llms},
  author={Zeng, Aohan and Liu, Mingdao and Lu, Rui and Wang, Bowen and Liu, Xiao and Dong, Yuxiao and Tang, Jie},
  journal={arXiv preprint arXiv:2310.12823},
  year={2023}
}

@article{grigsby2025vlm,
  title={VLM Q-Learning: Aligning Vision-Language Models for Interactive Decision-Making},
  author={Grigsby, Jake and Zhu, Yuke and Ryoo, Michael and Niebles, Juan Carlos},
  journal={arXiv preprint arXiv:2505.03181},
  year={2025}
}

@article{ziegler2019fine,
  title={Fine-tuning language models from human preferences},
  author={Ziegler, Daniel M and Stiennon, Nisan and Wu, Jeffrey and Brown, Tom B and Radford, Alec and Amodei, Dario and Christiano, Paul and Irving, Geoffrey},
  journal={arXiv preprint arXiv:1909.08593},
  year={2019}
}

@article{ouyang2022training,
  title={Training language models to follow instructions with human feedback},
  author={Ouyang, Long and Wu, Jeffrey and Jiang, Xu and Almeida, Diogo and Wainwright, Carroll and Mishkin, Pamela and Zhang, Chong and Agarwal, Sandhini and Slama, Katarina and Ray, Alex and others},
  journal={Advances in neural information processing systems},
  volume={35},
  pages={27730--27744},
  year={2022}
}

@inproceedings{carta2023grounding,
  title={Grounding large language models in interactive environments with online reinforcement learning},
  author={Carta, Thomas and Romac, Cl{\'e}ment and Wolf, Thomas and Lamprier, Sylvain and Sigaud, Olivier and Oudeyer, Pierre-Yves},
  booktitle={International Conference on Machine Learning},
  pages={3676--3713},
  year={2023},
  organization={PMLR}
}

@misc{PolyCam,
 author = {https://poly.cam/},
 title = {PolyCam},
 year = 2025
}



\newpage

\newpage

\appendix

\part*{Appendix}

\section*{Table of Contents} 

\begingroup

    \setlist[itemize]{leftmargin=*,topsep=1em}
    \setlist[itemize,1]{label={}} 
    \setlist[itemize,2]{label={}} 
    \setlist[itemize,3]{label={}} 

{\color{blue!80!black}\noindent\rule{\textwidth}{1pt}}
\begin{itemize}[leftmargin=*,itemsep=1em] 
    \item  \textcolor{blue!80!black}{\textbf{A: Sensitivity to Position/Localization Noise}} \dotfill \pageref{appendix_sec:localization_noise}
    
    \item \textcolor{blue!80!black}{\textbf{B:  Constructing Receptacle-Aware 3D Scene Graph}} \dotfill \pageref{appendix_sec:scene_graph}
    \begin{itemize}[leftmargin=2em,itemsep=1em]
        \item 
        \textcolor{blue!80!black}
        {B.1: Scene Graph Initialization} \dotfill 
        \pageref{appendix_subsec:initializing_scene_graph_tree}
        \item  
        \textcolor{blue!80!black}
        {B.2: Calculating Receptacle Regions} \dotfill \pageref{appendix_subsec:receptacles_regions}
        \begin{itemize}[leftmargin=2em,itemsep=1em]
            \item \textcolor{blue!80!black}{B.2.1: Initial Receptacles} \dotfill \pageref{appendix_subsubsec:initial_receptacles}
            \item \textcolor{blue!80!black}{B.2.2: Refined Receptacles} \dotfill \pageref{appendix_subsubsec:refined_receptacles}
            \item \textcolor{blue!80!black}{B.2.3: Merging Receptacles} \dotfill \pageref{appendix_subsubsec:merge_free_space}
        \end{itemize}
    \end{itemize}
    
    \item \textcolor{blue!80!black}{\textbf{C:  Generating Outcome-based Tasks}} \dotfill \pageref{appendix_sec:generating_outcome_based_tasks}
    \begin{itemize}[leftmargin=2em,itemsep=1em]
        \item \textcolor{blue!80!black}{C.1: Details of \ourdataset Dataset } \dotfill \pageref{appendix_subsec:dataset}
        \begin{itemize}[leftmargin=2em,itemsep=1em]
            \item \textcolor{blue!80!black}{C.1.1: Dataset Collection Process} \dotfill \pageref{appendix_subsubsec:dataset_collection_process}
            \item \textcolor{blue!80!black}{C.1.2: More Examples of \ourdataset} \dotfill \pageref{appendix_subsubsec:more_examples_of_ourdataset}
            \item \textcolor{blue!80!black}{C.1.3: Primary Intent and Semantic/Reasoning Coverage \dotfill \pageref{appendix_subsubsec:semantic_coverage}}
        \end{itemize}
        \item \textcolor{blue!80!black}{C.2: VLM-based Voting Mechanism} \dotfill \pageref{appendix_subsec:vlm_voting}
    \end{itemize}
    
    \item \textcolor{blue!80!black}{\textbf{D:  Further Experiment Results and Details}} \dotfill \pageref{appendix_sec:more_details_on_experiments}
    \begin{itemize}[leftmargin=2em,itemsep=1em]
        \item \textcolor{blue!80!black}{D.1: Details of ManiTaskGen-RAS and \ourgentask } \dotfill \pageref{appendix_subsec:manitaskgen_ras}
        \begin{itemize}[leftmargin=2em,itemsep=1em]
            \item \textcolor{blue!80!black}{D.1.1: Scene Data Preprocessing} \dotfill \pageref{appendix_subsubsec:preprocessing}
            \item \textcolor{blue!80!black}{D.1.2: Generating Tasks of Controlled Complexity and Length} \dotfill \pageref{appendix_subsubsec:atomic_task_generation}
            \item  \textcolor{blue!80!black}{D.1.3: Controlled GPTTaskGen Comparison and Original Validity Results \dotfill \pageref{appendix_subsubsec:controlled_baseline}}
        \end{itemize}
        \item  \textcolor{blue!80!black}{D.2: Reliability of Feasibility Filtering and Execution Judging \dotfill
        \pageref{appendix_subsec:evaluation_discrepancy}}
        \item \textcolor{blue!80!black}{D.3: Examples of Complete Benchmark Episodes} \dotfill \pageref{appendix_subsec:benchmark_example}
        \begin{itemize}[leftmargin=2em,itemsep=1em]
            \item \textcolor{blue!80!black}{D.3.1: Successful Episode} \dotfill \pageref{appendix_subsubsec:success_episode}
            \item  
            \textcolor{blue!80!black}
            {D.3.2: Failed Episode} \dotfill  
            \pageref{appendix_subsubsec:fail_episode}
        \end{itemize}
    \end{itemize}
\end{itemize}
\endgroup

\newpage


\section{{Sensitivity to Position/Localization Noise}}
\label{appendix_sec:localization_noise}
\label{appendix_subsec:localization_noise}

As demonstrated in Sec.~\ref{subsubsec:real_robot}, ManiTaskGen can be used not only to benchmark VLM agents in simulation but also to evaluate their high-level decision-making capabilities in real-world scenes. Since acquiring real-world scene data and reconstructing these scenes can introduce localization errors, we conduct this supplementary study to examine how input localization noise affects the validity of tasks generated by ManiTaskGen. Specifically, we perturb the input object positions before scene-graph construction, rebuild support relations and receptacle regions, and rerun task generation.

For each object $i$, the horizontal coordinates become $x'_i=x_i+\epsilon_{i,x}$ and $y'_i=y_i+\epsilon_{i,y}$, where each coordinate perturbation is uniform on $[-a,a]$ meters. The reported 5\,cm and 25\,cm settings mean $a=0.05$ and $a=0.25$, respectively: they are per-axis bounds, not standard deviations or fixed Euclidean displacement magnitudes. Only the $x,y$ position components are perturbed, not orientation, height, or bounding-box dimensions.

For each noise setting and each of Levels 1--4, 100 generated tasks are manually assessed for validity. Level 3 requires both relocation steps to be valid; Level 4 comprises outcome-based instructions. Table~\ref{tab:localization_noise} reports the full L1--L4 results from the supplementary study.

\begin{table}[htbp]
\centering

\begin{tabular}{lrrrr}
\toprule
Per-axis noise bound & L1 (\%) & L2 (\%) & L3, two-step (\%) & L4 (\%) \\
\midrule
No noise & 94 & 94 & 90 & 86 \\
5\,cm & 89 & 87 & 79 & 85 \\
25\,cm & 71 & 72 & 53 & 85 \\
\bottomrule
\end{tabular}
\caption{\textbf{Task validity under input localization noise.} Each cell summarizes 100 manually verified tasks generated after rebuilding the scene graph for that noise setting. This separate study does not replace the original validity comparison or measure robot success rates.}
\label{tab:localization_noise}
\end{table}

At the 5\,cm bound, Levels 1--3 retain 79--89\% validity, below their 90--94\% noise-free values. Increasing the bound to 25\,cm reduces them to 53--72\%. Two-step tasks are most sensitive, consistent with either step being able to invalidate the whole task. Larger position errors can corrupt support relations, free-space segmentation, and object-fit estimates. Level-4 validity is 86\%, 85\%, and 85\%; a possible explanation is that these goals depend less directly on precise receptacle geometry. This stability should not be generalized to other perception failures or autonomous execution. Missed, merged, and split object instances are not tested.

\section{Constructing Receptacle-Aware 3D Scene Graph}
\label{appendix_sec:scene_graph}
As described in Sec.~\ref{subsec:scene_graph}, the process of constructing the 3D scene graph mainly consists of 2 stages: (1) Initializing the scene graph based on mutual spatial relationships between objects; (2) Calculating receptacle regions anchored by objects. We provide further details on each part below.

\subsection{Scene Graph Initialization}\label{appendix_subsec:initializing_scene_graph_tree}
\noindent First, based on each object's pose and the size of its bounding box, we infer the spatial relationships between objects along the vertical axis. These spatial relationships are used to initialize the scene graph. In the scene graph, each node represents an object, and a child node indicates that the corresponding object is placed on the upper surface of its parent. For scenes that provide object mesh models (e.g., ReplicaCAD and AI2THOR), we additionally extract internal surfaces suitable for object placement and record them in nodes' attributes. Consequently, the scene graph is extended such that a child node can be placed on either an internal or external surface of its parent node. To facilitate subsequent identification and indexing of receptacles within the scene, we also define the following attributes for each node (i.e., each object) during the scene graph initialization stage.

\noindent\textbf{Ground Objects and Surface Objects.} We divide all objects in a scene into ground objects and surface objects, where ground objects are those resting directly on the ground, while surface objects are those supported by other objects. We assume the robot can only manipulate surface objects. The implementation consequently uses the navigate--pick--place scope and support-surface/receptacle representation of Sec.~\ref{subsec:premise}; reconstructed physical-scene inputs and their limitations are discussed in Sec.~\ref{subsubsec:real_robot}.

\noindent\textbf{Walkable Spaces.} For each ground object, we define the space unobstructed in the vertical direction of its bounding box as its walkable space. These walkable spaces serve as navigation targets for the robot when traversing the scene. For surface objects placed on ground objects, the corresponding ground object's walkable space is also used as the robot's navigation target when approaching those surface objects. In our implementation, we adopt a threshold of 50 cm to determine whether the unobstructed space in a given direction is sufficient to be considered walkable. A single ground object may have multiple walkable spaces, corresponding to different edges of its bounding box.

\begin{figure}[t]
    \centering
    \includegraphics[width=\columnwidth]{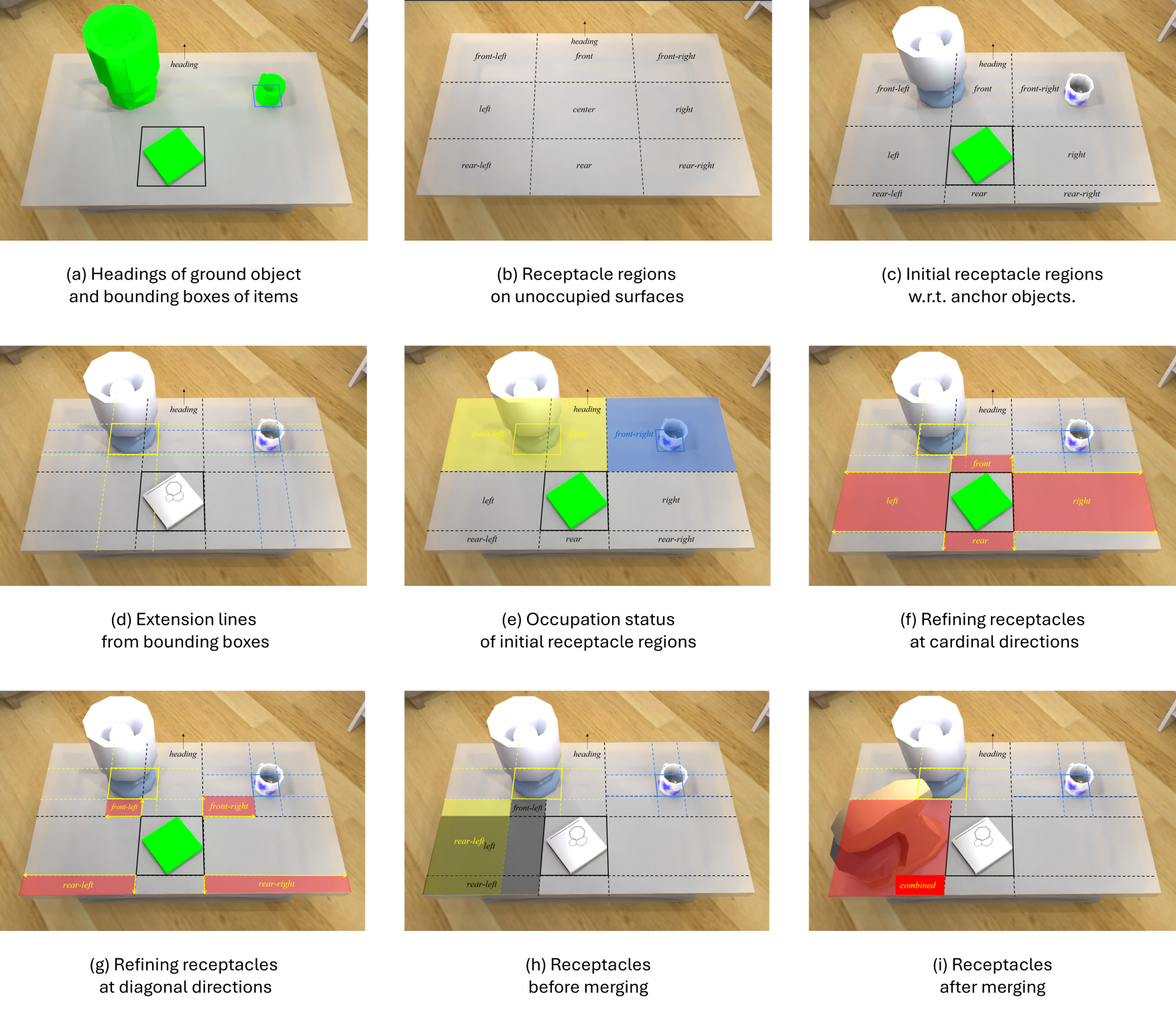}
    \vspace{-3mm}
    \caption{\textbf{Visualization of Rules Involved in Scene Graph Construction.}} 
    \vspace{-3mm}
    \label{fig:receptacle_demonstration}
\end{figure}

\noindent\textbf{Headings of Objects.} To support the generation of direction-related tasks (e.g., placing object A to the left of object B), we also define the heading of each object. For ground objects, we assume that their heading is determined by the robot's standing orientation when navigating to the area around the object (decided by the walkable spaces). Specifically, the direction the robot is facing is taken as the \textit{front} (heading) of the ground object, as shown in Fig.\ref{fig:receptacle_demonstration} (a). If a ground object has multiple walkable spaces, its heading is not fixed and depends on the robot's actual standing orientation. For surface objects, we assume their heading always aligns with that of the supporting ground object. Therefore, for surface objects whose bounding boxes are not parallel to those of their supporting ground objects, we still use the ground object's heading as the sole reference when reasoning about their directional relationships.

\subsection{Calculating Receptacle Regions}\label{appendix_subsec:receptacles_regions}
We construct accurate receptacle region retrieval information in three steps. First, we compute an initial set of rough receptacles based on empty surfaces or the spatial relationships between existing objects and the supporting surfaces (Sec.~\ref{appendix_subsubsec:initial_receptacles}). Next, we refine these receptacles by considering the relative positions of objects on the same surface (Sec.~\ref{appendix_subsubsec:refined_receptacles}). Finally, we introduce rules for merging adjacent or overlapping receptacles (Sec.~\ref{appendix_subsubsec:merge_free_space}).

\subsubsection{Initial Receptacles}\label{appendix_subsubsec:initial_receptacles}

\noindent Initial receptacles is computed mainly to better determine the relative positioning of objects.

\noindent For unoccupied surfaces, we divide the area into a 3×3 grid consisting of nine identical rectangular regions, as visualized in Fig.~\ref{fig:receptacle_demonstration} (b). The interior rectangle is defined as the "center" region. For the surrounding eight regions, we designate their directions based on the surface's heading: the direction aligned with the heading is labeled "front," with the remaining regions proceeding counterclockwise as "front-left," "left," "rear-left," "rear," "rear-right," "right," and "front-right." These receptacle regions are directly indexed by surfaces in the scene graph.
For occupied surfaces, we use a different approach while maintaining the same directional terminology, as shown in Fig.~\ref{fig:receptacle_demonstration} (c) and (d). We treat each surface object as an anchor object to index the receptacle regions around it. Specifically, we extend the vertices and edges of each object's bounding box outward until they reach the platform's boundaries. The regions swept by the extended edges define the object's initial receptacles in four cardinal directions (front, rear, left, right), while the paths traced by the extended vertices determine the initial receptacles in the four diagonal directions (front-left, front-right, rear-left, rear-right).

\noindent With these initial receptacles for each anchor object, we then determine the relative positioning of other objects on the same surface by identifying which of our calculated initial receptacles they overlap with. Fig.~\ref{fig:receptacle_demonstration} (e) illustrates this process. The relative positioning and occupation information will also be recorded in the scene graph.

\subsubsection{Refined Receptacles}\label{appendix_subsubsec:refined_receptacles}
We proceed to refine the initial receptacles based on accurate object positions and bounding boxes, and record their properties into the scene graph.

\noindent For cardinal directions, we project the edge of an object's bounding box along its perpendicular direction until it intersects another object's bounding box or the platform boundary. The region swept during this projection forms the refined receptacle in this cardinal direction, as shown in Fig.~\ref{fig:receptacle_demonstration} (f).

\noindent For diagonal directions, we project the vertices of the object's bounding box along two corresponding cardinal directions until they encounter the extension line of another object's bounding box or the platform boundary. The rectangle formed by these two projection paths represents the refined receptacle in this diagonal direction, as illustrated in Fig. ~\ref{fig:receptacle_demonstration} (g).

\noindent Thus, these refined receptacles are treated as final ones indexed by anchor objects. Together with the receptacle regions indexed directly by empty surfaces, they establish a systematic representation for efficient and retrieval of the free spaces on surfaces.

\subsubsection{Merging Receptacles}
\label{appendix_subsubsec:merge_free_space}
\noindent Merging receptacles is applied in task generation and during interaction for object placement, to support retrieving multiple refined receptacles.

\noindent When generating process-based tasks, for any surface object in the scene, we traverse all receptacle regions in the scene to determine whether there is sufficient space to place the object. If not, for multiple receptacle regions belonging to the same surface, we attempt to merge up to four adjacent or overlapping receptacle regions. If the merged receptacle region provides enough space to accommodate the object, we then generate the corresponding task. Fig.~\ref{fig:receptacle_demonstration} (h) shows an example of merging four receptacles when generating the task of placing a large kitchen utensil in the scene.

\noindent During interaction between the agent and the environment, the agent may select any number of receptacles on a surface, regardless of whether they are indexed by the same anchor object. The system will then attempt to merge them into a larger receptacle region if they are connected or overlapped, and to assess if the resulted one has enough space to fit the object, as shown in Fig.~\ref{fig:receptacle_demonstration} (i). 

\section{Generating Outcome-based Tasks}\label{appendix_sec:generating_outcome_based_tasks}
\subsection {Details of \ourdataset Dataset}\label{appendix_subsec:details_of_ourdataset}
\label{appendix_subsec:dataset}

\subsubsection{Dataset Collection Process}\label{appendix_subsubsec:dataset_collection_process}
The \ourdataset dataset contains 200 templates for outcome-based tasks, processed based on the human instructions collected from Amazon Mechanical Turk (AMT)~\cite{amazonturk}.
We first rendered scene images from real-world and simulated datasets, including Habitat~\cite{szot2021habitat}, ScanNet~\cite{dai2017scannet}, and SUN-RGBD~\cite{song2015sun}. Next, we launched an instruction collection job in AMT~\cite{amazonturk} to gather natural language instructions from human annotators specific for the provided images. The detailed job description is shown below:

\begin{quote} 
\textbf{General Description}: Imagine a powerful mobile robot that can move any object except immovable ground furniture. Describe in natural language the tasks it can perform, especially high-level complex tasks.

\textbf{Detailed Instruction}:
Imagine we have a powerful mobile robot that can move and place any non-ground object in a room. Your task is to describe high-level, complex tasks the robot can perform in this room using natural language. Pictures of the room and its layout are provided here (link). First, please review layout.png, [0][ROOM].png, and then all the other images to understand the room layout and object reference information. Make sure your tasks are reasonable and closely follow the provided images.

\textbf{Key Points}: (1) For immovable ground objects in the room, when referring to these objects, strictly use the numbered labels listed below: [1][DESK], [2][SHELVING], etc. (2) When referring to other small objects, you can describe them freely, but make sure to enclose the object names in square brackets, like this: [book], [plastic bottle], [small robotic arm]. (3) Please use your imagination as much as possible and give complex, high-level, and imaginative instructions.

\textbf{Examples of tasks we want}:

(1) Help me clear the top of [1][DESK].

(2) Organize all [soda cans] on the top layer of [2][SHELVING] to line them up in a row.

(3) Tidy up the [2][SHELVING] to make it aesthetically pleasing.

(4) The [7][DESK] is too messy. Organize the objects on top of it to make it neat.

(5) Take out the [cans] from under the [paper box] on the left side of [3][DESK] and arrange them into an equilateral triangle.

We view these examples as postive ones because they contain complex, high-level instructions.

\textbf{Examples of tasks we do NOT want}:

(1) Move the [paperbox] under [1][DESK] to [8][DESK]

(2) Pick up the [tiny robotic arm] from [3][DESK] and place it on the [5][CHAIR].

We view these examples as negative ones because these tasks are too simple or the instructions are too detailed.

\end{quote}

Based on the collected human instructions, we manually filtered out low-quality and repeated ones and removed specific object names in each instruction to create the final templates. In the end, we organized a set of 200 task templates, named as \ourdataset.This is template-level quality control and duplicate removal, not semantic or behavioral deduplication of the full RAS-40K task set.

\subsubsection{More Examples of \ourdataset}\label{appendix_subsubsec:more_examples_of_ourdataset}
We provide more templates from \ourdataset as follows:
\begin{quote} 
"Align all [SUB-PLATFORM-OBJECTS00] on [PLATFORM0] symmetrically."

"Create a "staircase" on [PLATFORM0] by arranging the objects to resemble steps."

"Create a pattern with the [SUB-PLATFORM-OBJECTS00] on [PLATFORM0] by placing them in a spiral, starting from the center of the [PLATFORM0] and expanding outward."

"Arrange [SUB-PLATFORM-CATEGORY-OBJECTS00] on [PLATFORM0] by height and type."

"Design a thematic organization system on [PLATFORM0] for all [SUB-PLATFORM-OBJECTS00], separating items by color and function."

"Organize the entire contents of [PLATFORM0] by categorizing items into labeled sections, with frequently used items at a central location for easy access."

"Declutter the right side of [PLATFORM0] by moving all loose objects to the left side and aligning them by size."

"Design a "mirrored room" on top of [PLATFORM0] by arranging the objects to create a sense of symmetry and reflection."

"Create an artistic arrangement on [PLATFORM0]."

"Carefully arrange all the [SUB-PLATFORM-CATEGORY-OBJECTS00] on [PLATFORM0] in ascending order of size, leaving a clear space in the center for the [SUB-PLATFORM-OBJECTS00] from [PLATFORM0] to be placed as a decorative piece."

\end{quote}

\subsubsection{{Primary Intent and Semantic/Reasoning Coverage}}
\label{appendix_subsubsec:semantic_coverage}

To complement the spatial coverage metrics reported in the diversity assessment in Sec.~\ref{sec:eval}, we further characterize the diversity of outcome-based instructions in terms of their intended goals and semantic/reasoning requirements. We analyze all 200 templates in \ourdataset and all 650 retained outcome-based instances in \ourgentask{} from the three representative scenes. Two annotations are complementary: a \emph{primary intent} identifies the dominant final purpose and is mutually exclusive, whereas \emph{semantic/reasoning tags} identify requirements expressed by the instruction and are multi-label. Table~\ref{tab:primary_intent_coverage} combines the template and instance intent distributions in the same category order; Table~\ref{tab:semantic_tag_coverage} does the same for tags.

\noindent\textbf{Category overlap.} A task receives one primary label according to its dominant final purpose, not a separate intent for every operation used to achieve it. Attribute ordering and semantic grouping can therefore coexist as tags even when ordering is the primary intent. For example, ``Arrange [SUB-PLATFORM-CATEGORY-OBJECTS00] on [PLATFORM0] by height and type'' is labeled ordering/alignment, with attribute-comparison and semantic-categorization tags. Generic tidying applies when no more specific dominant purpose is stated; ``Others'' includes deliberate disorganization. Tags preserve secondary requirements rather than forcing mutually exclusive capability labels. The tags characterize instruction semantics, while executable operations remain limited to navigate, pick, and place.

\begin{table}[p]
\centering

\small
\setlength{\tabcolsep}{4pt}
\renewcommand{\arraystretch}{1.15}
\begin{tabularx}{\textwidth}{@{}p{3.25cm}Xrr@{}}
\toprule
Primary intent & Definition / decision rule & Templates & Instances \\
 & & $n$ (\%) & $n$ (\%) \\
\midrule
Ordering / alignment & Explicit order, gradient, row, grid, line, symmetry, or regular alignment. & 36 (18.0) & 57 (8.8) \\
Display / presentation / curation & Informative, interactive, aesthetic, product, gallery, or focal-point display. & 31 (15.5) & 104 (16.0) \\
Functional workspace / workflow setup & Workspace, studio, lab, meeting/collaboration area, or multi-step workflow. & 24 (12.0) & 51 (7.8) \\
Artistic / geometric arrangement & Non-representational artistic composition, geometric pattern, sculpture, or installation. & 24 (12.0) & 57 (8.8) \\
Thematic scene / structural construction & Recognizable scene, model, navigable layout, or physical structure, e.g., city, maze, garden, bridge. & 24 (12.0) & 40 (6.2) \\
Grouping / categorization & Categories, sections, clusters, or designated areas based on semantic similarity. & 21 (10.5) & 32 (4.9) \\
Storage / relocation / redistribution & Moving, storing, containing, or redistributing items across platforms, shelves, layers, or room locations. & 20 (10.0) & 149 (22.9) \\
General organization / tidying & Generic tidying/reorganization without a more specific dominant purpose. & 8 (4.0) & 51 (7.8) \\
Accessibility / ergonomic optimization & Reachability, retrieval speed, eye-level placement, ergonomics, or movement efficiency. & 7 (3.5) & 70 (10.8) \\
Clearing / decluttering & Removing objects, reducing the retained set, clearing a surface, or freeing space. & 4 (2.0) & 25 (3.8) \\
Others & Intents not covered above, including deliberate disorganization. & 1 (0.5) & 14 (2.2) \\
\midrule
Total & One intent per template/instance. & 200 (100) & 650 (100) \\
\bottomrule
\end{tabularx}
\caption{\textbf{Primary-intent distributions of outcome-based templates and task instances.} Percentages use 200 templates and 650 instances, respectively; instance percentages are rounded independently. Counts are taken from the archived template classification and instance-intent statistics.}
\label{tab:primary_intent_coverage}
\end{table}

\begin{table}[p]
\centering

\small
\setlength{\tabcolsep}{4pt}
\renewcommand{\arraystretch}{1.15}
\begin{tabularx}{\textwidth}{@{}p{3.25cm}Xrr@{}}
\toprule
Semantic / reasoning tag & Scope / examples & Templates & Instances \\
 & & $n$ (\%) & (\%) \\
\midrule
Thematic / aesthetic composition & Visual harmony, display, story/theme, color scheme, atmosphere, centerpiece. & 99 (49.5) & 49.5 \\
Semantic categorization & Category, type, material, function, power source, project, similarity, thematic grouping. & 61 (30.5) & 16.0 \\
Symmetry / geometric reasoning & Symmetry, mirror, circle, spiral, grid, triangle, geometric shapes, maze/path geometry. & 60 (30.0) & 19.5 \\
Attribute comparison \& ordering & Size, height, length, color/brightness, frequency, relevance, chronology, alphabetical order. & 54 (27.0) & 28.0 \\
Accessibility / ergonomics / workflow & Access, reachability, retrieval, interaction, efficiency, collaboration, handover, navigation. & 53 (26.5) & 25.8 \\
Directional / spatial relation & Left/right, front/back, center/edge/corner, around, near/far, layers, eye level. & 50 (25.0) & 31.2 \\
Alignment / spacing / proportion & Row/column/line, even spacing, equidistance, parallel orientation, balance, proportion. & 49 (24.5) & 6.3 \\
Stacking / containment / physical structure & Stack, pile, bundle, fold, contain/store, connect, support, fit, stable structure. & 32 (16.0) & 5.8 \\
Cross-platform / layer transfer & Gathering or moving across platforms, shelves, layers, or room locations. & 31 (15.5) & 20.6 \\
Selection / decluttering / cleaning & Clear, remove unnecessary items, retain a subset, clean, or free a work area. & 23 (11.5) & 9.4 \\
Others & No explicit operation beyond a generic instruction, or deliberate disorganization. & 5 (2.5) & 6.2 \\
\midrule
Total tag assignments & Tags are non-exclusive. & 517 (258.5) & --- \\
Mean tags per item & Reported instance mean is approximate. & 2.59 & $\approx 2.18$ \\
\bottomrule
\end{tabularx}
\caption{\textbf{Multi-label semantic and reasoning coverage of outcome-based templates and task instances.}  Percentages use 200 templates and 650 instances; they need not sum to 100\%. Instance-level tag frequencies are reported as rounded percentages, with approximately 2.18 tags per instance.}
\label{tab:semantic_tag_coverage}
\end{table}

All 11 primary intents and all tag families occur in both sets. The largest intent share changes from 18.0\% (ordering/alignment in templates) to 22.9\% (storage/redistribution in instances). Templates have $517/200=2.585$ tags on average, reported as 2.59. The approximately 2.18 tags per instance and the coverage of spatial, attribute, workflow, transfer, and geometric requirements characterize semantic demands beyond object-name substitution. Together, these annotations characterize the semantic requirements of the generated instructions.

\clearpage

\subsection {VLM-based Voting Mechanism}
\label{appendix_subsec:vlm_voting}
As described in Sec.~\ref{subsec:task_gen}, the process of generating outcome-based tasks involves 2 steps. Firstly, we fill the templates from \ourdataset with valid combinations of platforms and objects, and then we employ a VLM-based voting mechanism to roll out reasonable tasks.

For the original dataset generation, we used an ensemble of 3 VLMs (GPT-4o~\cite{openai2024gpt4technicalreport}, Gemini-2.5-pro~\cite{geminiteam2024geminifamilyhighlycapable}, Claude-3.7-sonnet~\cite{anthropic2024aboutclaude}) to assess a given task. For each VLM, we introduce the task goals and the capabilities of the agent (navigate, pick, place), along with pictures of all platforms involved in the task. Each VLM is required to output 1 of 3 assessment levels: feasible, partially feasible or not feasible. A task will be included in the final dataset only if at least 2 out of the 3 VLMs vote for it as a feasible one. Appendix~\ref{appendix_subsec:evaluation_discrepancy} reports a separate reliability study with supplemental models, including rejected candidates and all four vote-count bins (Table~\ref{tab:filtering_vote_bins}); it does not redefine the original generation ensemble. The detailed prompt we use for this process is shown below:
\begin{tcolorbox}[title = {Prompts for Outcome-based Task Assessment},width=\textwidth, breakable]

Task: [TASK\_DESCRIPTION, IMAGES\_OF\_INVOLVED\_\allowbreak OBJECTS\_AND\_PLATFORMS]
            
Evaluate if the given task is feasible for such a robot whose abilities involve: navigate, pick up any movable object and place objects anywhere they fit on a given platform.

Assessment criteria:

1. Are all required objects present in the scene?

2. Is there sufficient free space on the target platform given the required object placement?

3. Are the required state achievable given the robot's capabilities and the objects we have in the involved platforms?

4. Would completing the task create any unstable or physically impossible object arrangements?

Only output a single line: "Feasible", "Partially feasible", or "Not feasible". Do not output anything else.
\end{tcolorbox}

\section{Further Experiment Results and Details}\label{appendix_sec:more_details_on_experiments}
\subsection{Details of ManiTaskGen-RAS and \ourgentask}
\label{appendix_subsec:manitaskgen_ras}
\begin{figure}[t]
    \centering
    \includegraphics[width=\columnwidth]{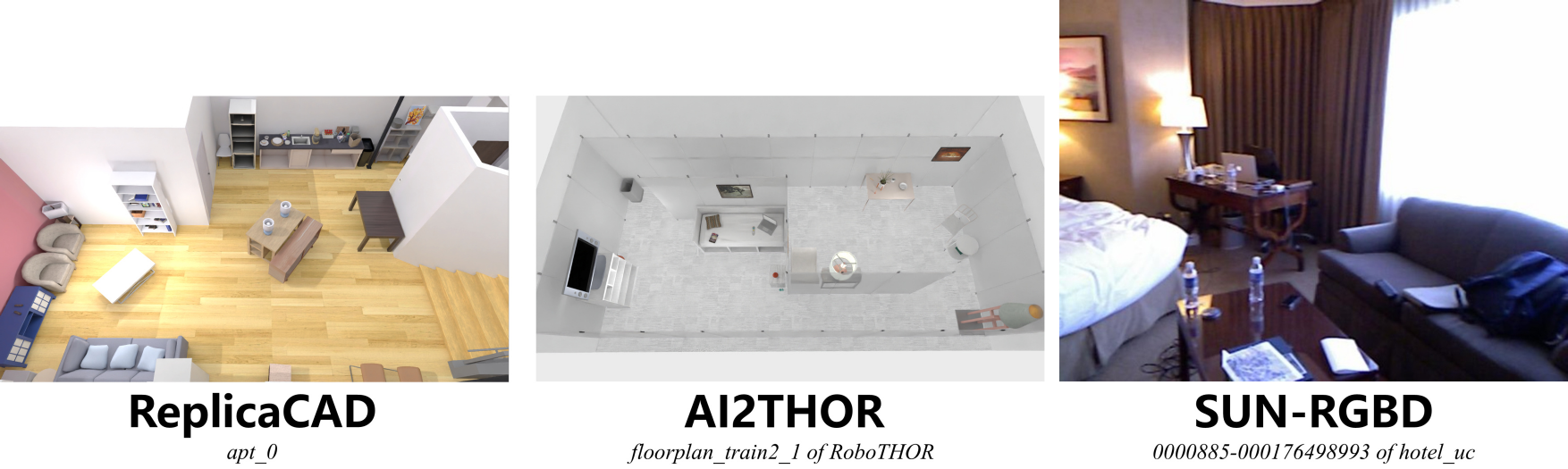}
    \vspace{-3mm}
    \caption{\textbf{Visualization of the 3 Scenes Used to Generate \ourgentask.} } 
    \vspace{-3mm}
    \label{fig:scene_overview}
\end{figure}

As described in Sec.~\ref{sec:eval}, we applied ManiTaskGen to 100 scenes from ReplicaCAD, AI2THOR, and SUN-RGBD. Table~\ref{tab:manitaskgenras_task_counts} reports finite represented task-space counts: single-step candidates at Levels 1--2, two-step combinations at Level 3, and estimated outcome-task counts at Level 4. The approximately $3.96\times10^9$ Level-3 combinations are not billions of stored or independently executed episodes. Level-4 counts are estimates because exhaustive template instantiation and VLM filtering are computationally costly. By contrast, Table~\ref{tab:task level-distribution} reports the actual 39,871 instantiated RAS-40K instructions: 39,221 process-based tasks and 650 outcome-based tasks. The outcome-based tasks instantiate \ourdataset templates and use the original voting ensemble described in Appendix~\ref{appendix_subsec:vlm_voting}. Process-based tasks share the four relocation patterns in Table~\ref{tab:AtomicTaskGenerationLogic}, composed into one- or two-step sequences in RAS-40K. Different objects, source/target platforms, spatial constraints, perceptual ambiguity, and action orders provide distinct evaluation conditions; task counts alone do not establish independent behaviors, and full-dataset semantic or behavioral redundancy has not been quantified.

To further curate ManiTaskGen-RAS-40K for detailed evaluation, we select one representative scene from each of the three dataset sources (ReplicaCAD, AI2THOR, and SUN-RGBD) out of the 100 scenes. For ReplicaCAD, we select the scene \textit{apt\_0};  For AI2THOR, we select the scene \textit{floorplan\_train2\_1  of RoboTHOR}; For SUN-RGBD, we choose the scene \textit{0000885-000176498993 of hotel\_uc}. Fig.~\ref{fig:scene_overview} shows the visualization of these scenes. Tab.~\ref{tab:task level-distribution} shows the statistics of ManiTaskGen-RAS-40K. 

\begin{table}[htbp]
  \centering
  \crtablespacing
  \small
  \renewcommand{\arraystretch}{1.1}
  \begin{tabular}{lcccc}
    \toprule
    {\textbf{Dataset}} & {\textbf{Scenes}} & \multicolumn{3}{c}{\textbf{Represented task-space count}} \\
    \cmidrule(lr){3-5}
     & & \textbf{Level 1+2} & \textbf{Level 3} & \textbf{Level 4}\\

    \midrule
     SUNRGBD  & 5 & $ 486
     $ & $ 52,956 $ & $200$ \\
    \midrule
    ReplicaCAD &  5 & $119,682$ & $ 2.86 \times 10^9$ & $75,000$ \\

    \midrule
     AI2THOR & 90 & $213,446$ & $1.10 \times 10^9$ & $250,000$ \\
    \midrule

    \textbf{Total} &  \textbf{100} & $\mathbf{333,614}$ & $\mathbf{3.96 \times 10^9}$ & $\mathbf{325,200}$\\
    \bottomrule
  \end{tabular}
     \caption{\textbf{Task-space statistics across 100 source scenes.} Levels 1--2 count represented single-step candidates; Level 3 counts two-step combinations under the generation rules; Level 4 is an estimate rather than an exhaustive VLM-filtered enumeration. None of these columns measures semantically deduplicated behaviors. Actual RAS-40K instances are listed in Table~\ref{tab:task level-distribution}.}
    
    \label{tab:manitaskgenras_task_counts}
\end{table}

\begin{table}[htbp]
\centering
\crtablespacing
\begin{tabular}{lccccc}

\toprule
\textbf{Dataset} & \textbf{Level 1} & \textbf{Level 2} & \textbf{Level 3} & \textbf{Level 4} & \textbf{Total}\\
\midrule
ReplicaCAD & 2195  & 24153 & 5000 & 500 & 31848\\
AI2THOR    & 2218  & 0     & 5000 & 100 & 7318\\
SUN-RGBD   & 35    & 120   & 500  & 50  & 705\\
Total & 4448 & 24273 & 10500 & 650 & 39871 \\
\bottomrule
\end{tabular}
\caption{\textbf{Statistics of ManiTaskGen-RAS-40K across 3 Source Scenes.}}
\label{tab:task level-distribution}
\end{table}

Next, we demonstrate the detail of generating tasks using the selected scenes. In Sec.~\ref{appendix_subsubsec:preprocessing}, we illustrate the data preprocessing pipeline; In Sec.~\ref{appendix_subsubsec:atomic_task_generation}, we describe how we use the processed data to generate \ourgentask{} of controlled task complexity and length. Sec.~\ref{appendix_subsubsec:controlled_baseline} reports the controlled comparison with GPTTaskGen baselines and the three-scene task-validity evaluation.

\subsubsection{Scene Data Preprocessing}
\label{appendix_subsubsec:preprocessing}
For each given scene, our preprocessing pipeline consists of two parts: \textbf{Item Renaming} and \textbf{Removal of Overlapping Objects}.

\begin{figure}
    \centering
    \includegraphics[width=\textwidth]{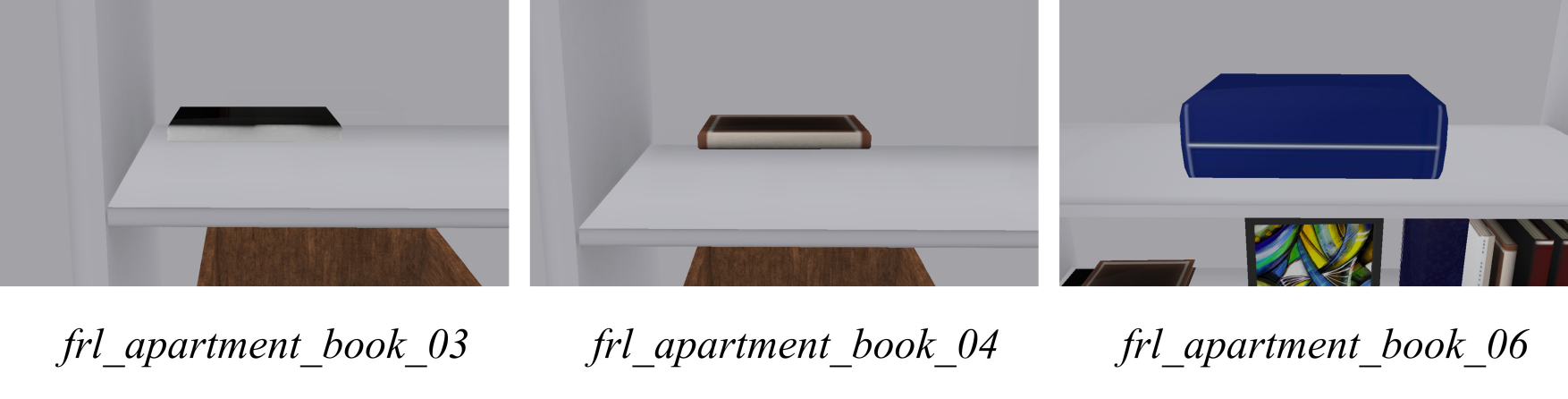}
    \includegraphics[width=\textwidth]{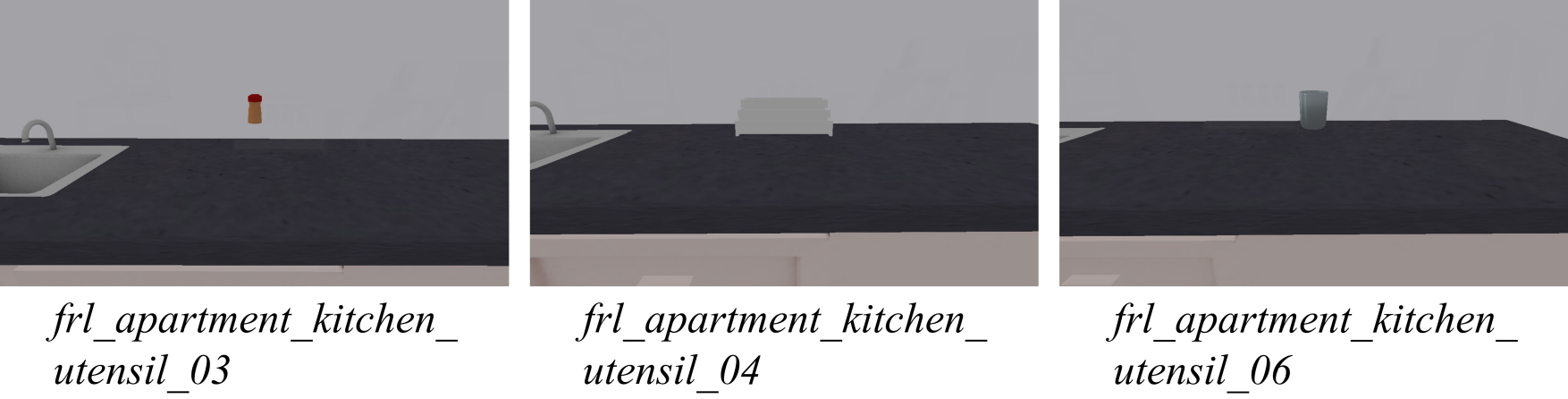}
     \vspace{-1.5em}
    \caption{\textbf{Ambiguous Item Naming in ReplicaCAD.}}

    \label{fig:ReplicaAmbiguousItems}
\end{figure}

\noindent\textbf{Item Renaming.} We observe that the original object naming in certain input scenes (e.g., ReplicaCAD and AI2THOR) is somewhat casual. As shown in Fig.~\ref{fig:ReplicaAmbiguousItems}, many books and kitchen utensils share identical base names and are only distinguished by numerical suffixes. This naming convention makes it difficult for the generated process-based tasks to differentiate between Level-1 and Level-2 tasks (as defined in Sec.~\ref{subsubsec:benchmark_task_and_eval}).To address this, we employ a VLM to rename objects in the given scene. The new naming scheme follows the format "\{category\_name\}\_\{specific\_name\}". When a process-based task involves an object that shares the same \{category\_name\} with other objects on the same surface, the task is classified as Level-2; otherwise, it is treated as Level-1. The prompt we provide to the VLM is detailed below. The renaming results for ReplicaCAD are presented in Tab.~\ref{tab:ReplicaCADItemNames}, 

\begin{tcolorbox}[title = {Prompts for Item Renaming },width=\textwidth, breakable]

You are a professional item classification and naming assistant. You need to provide a unique and meaningful name for each item based on indoor object images.

I need your help to classify and rename a series of indoor object images. Each image contains an item, probably on a supporting platform, to be named. Please follow these requirements:

1. Naming Rules:

- Each item name must be unique

- Names should reflect the item's category and characteristics

- Use English naming format '\{category\_name\}\_\{specific\_name\}', no non-alphabetic characters allowed

2. Naming Examples:

- A red mug -> kitchenware\_red\_mug

- A textbook -> book\_textbook

- A desk lamp -> electronics\_desk\_lamp

Please review the images and provide standardized names according to the above rules.  The following images are the items to be classified, one item per image.

[Item Image]

The provided image contains an item to be renamed, probably being placed on some platform. Your response should be a single line of the new name, with '\_' connecting different words, e.g.'utensil\_white\_soap\_dispenser' (without quotes).Don't include redundant characters, don't include characters other than alphabets and underlines, don't coincide with any previous name you've given in the [ITEM NAME LIST].
\end{tcolorbox}


\begin{table}[htbp]
\centering
\crtablespacing
\begin{tabular}{|c|c|}
\hline
\textbf{Name (before)} & \textbf{Name (after)} \\
\hline
frl\_apartment\_kitchen\_utensil\_01 & kitchenware\_ceramic\_lidded\_container \\
frl\_apartment\_kitchen\_utensil\_02 & bathroom\_light\_green\_pump\_bottle \\
frl\_apartment\_kitchen\_utensil\_03 & kitchenware\_spice\_container \\
frl\_apartment\_kitchen\_utensil\_04 & kitchenware\_white\_paper\_towel\_holder \\
frl\_apartment\_kitchen\_utensil\_05 & kitchenware\_navy\_blue\_lidded\_dish \\
frl\_apartment\_kitchen\_utensil\_06 & kitchenware\_short\_glass \\
frl\_apartment\_kitchen\_utensil\_08 & kitchenware\_grey\_mug \\
frl\_apartment\_kitchen\_utensil\_09 & bathroom\_dark\_blue\_bottle \\
frl\_apartment\_knifeblock & kitchenware\_wooden\_knife\_block \\
frl\_apartment\_lamp\_01 & lighting\_modern\_table\_lamp \\
frl\_apartment\_lamp\_02 & lighting\_blue\_base\_table\_lamp \\
frl\_apartment\_pan\_01 & kitchenware\_dark\_blue\_saucepan \\
frl\_apartment\_picture\_02 & decor\_autumn\_tree\_canvas \\
frl\_apartment\_picture\_03 & decor\_silver\_photo\_frame \\
frl\_apartment\_picture\_04 & decor\_abstract\_framed\_print \\
frl\_apartment\_plate\_01 & kitchenware\_white\_octagonal\_plate \\
frl\_apartment\_plate\_02 & kitchenware\_wooden\_cake\_stand \\
frl\_apartment\_remote-control\_01 & electronics\_black\_remote\_control \\
frl\_apartment\_shoebox\_01 & storage\_dark\_grey\_round\_box \\
frl\_apartment\_shoe\_01 & shoes\_black\_oxford\_shoes \\
frl\_apartment\_shoe\_02 & shoes\_brown\_leather\_oxford \\
frl\_apartment\_shoe\_03 & shoes\_red\_slingback\_sandal \\
frl\_apartment\_shoe\_04 & shoes\_blue\_sneaker \\
frl\_apartment\_small\_appliance\_01 & kitchenware\_beige\_coffee\_grinder \\
frl\_apartment\_small\_appliance\_02 & bathroom\_grey\_soap\_dispenser \\
frl\_apartment\_sponge\_dish & kitchenware\_small\_white\_bowl\_with\_butter \\
frl\_apartment\_basket & kitchenware\_beige\_utensil\_holder \\
frl\_apartment\_book\_01 & books\_dark\_red\_hardcover \\
frl\_apartment\_book\_02 & books\_navy\_blue\_hardcover \\
frl\_apartment\_book\_03 & books\_brown\_hardcover\_journal \\
frl\_apartment\_book\_04 & books\_black\_hardcover \\
frl\_apartment\_book\_05 & electronics\_white\_router\_box \\
frl\_apartment\_book\_06 & electronics\_navy\_blue\_printer \\
frl\_apartment\_bowl\_06 & kitchenware\_small\_white\_bowl \\
frl\_apartment\_bowl\_07 & kitchenware\_white\_porcelain\_bowl \\
frl\_apartment\_box & storage\_brown\_wooden\_box \\
frl\_apartment\_camera\_02 & electronics\_white\_security\_camera \\
frl\_apartment\_choppingboard\_02 & decor\_wooden\_cube \\
frl\_apartment\_clock & decor\_mantel\_clock \\
frl\_apartment\_cup\_01 & kitchenware\_small\_white\_mug \\
frl\_apartment\_cup\_02 & kitchenware\_blue\_floral\_mug \\
frl\_apartment\_cup\_03 & kitchenware\_beige\_tea\_cup \\
frl\_apartment\_cushion\_03 & furniture\_light\_blue\_throw\_pillows \\
\hline
\end{tabular}
\caption{\textbf{Renaming Results of ReplicaCAD.}}
\label{tab:ReplicaCADItemNames}
\end{table}



\noindent\textbf{Removal of Overlapped Objects.}We further observe that in certain scenes (e.g., SUN-RGBD), objects located on the same surface exhibit significant bounding box overlap. During preprocessing, we remove such objects to ensure data cleanliness.

\subsubsection{Generating Tasks of Controlled Complexity and Length} \label{appendix_subsubsec:atomic_task_generation}

After preprocessing the input scene data, we build the scene graph and use it to generate the tasks, as described in Sec.~\ref{subsec:task_gen}.
We first generate atomic actions using the patterns provided in Tab.~\ref{tab:AtomicTaskGenerationLogic}.
After that, we enumerate the single-step pick-and-place candidates admitted by these patterns and the represented geometric constraints (Levels 1 and 2). For more complex, longer-horizon multi-step pick-and-place tasks, we limited them to two-step sequences and sampled 5,000 instances for ReplicaCAD and AI2THOR, while only sampling SUN-RGBD to 500 instances (due to fewer objects and platforms in the scene of SUN-RGBD). Similarly, for outcome-based tasks, we generated constrained quantities for different scenes, with specific numbers detailed in Tab.\ref{tab:task level-distribution}. Note that, for AI2THOR scenes, no Level-2 tasks were generated since there were no instances of multiple objects belonging to the same category existing on the same surface after the renaming process.

\begin{table}[htbp]
\centering
\crtablespacing
\begin{tabular}{|p{3.6cm}|p{\dimexpr\textwidth-3.6cm-4\tabcolsep-3\arrayrulewidth\relax}|}
\hline
\textbf{Atomic Action Pattern} & \textbf{Generation Logic} \\
\hline
Move to a named surface & If the object can fit on the surface. \\
\hline
Move to a location around a specific object & If there exists a place to fit this object, which only involves receptacle regions of one specific object. \\
\hline
Move to a location with specific direction relative to a specific object & If there exists a place to fit this object, which only involves receptacle regions of specific direction relative to the specific object. \\
\hline
Move to a location between two objects & If there exists a place to fit this object, and the covered receptacle regions correspond to complementary directional pairs from two different objects (e.g., left side of object A and right side of object B; front-left of object A and back-right of object B). \\
\hline
\end{tabular}
\caption{\textbf{Patterns and Logics for Generating Atomic Actions.}}
\label{tab:AtomicTaskGenerationLogic}
\end{table}

\newpage

\subsubsection{Controlled GPTTaskGen Comparison and Original Validity Results}
\label{appendix_subsubsec:controlled_baseline}

To further assess the validity of the generated tasks under different method, we compare the four methods on the same ReplicaCAD scene, using 100 evaluated or retained single-step tasks per method.

\noindent\textbf{Scene and inputs.} GPTTaskGen uses GPT-4o with object/platform names and scene images. GPTTaskGen-SG additionally receives the Receptacle-Aware 3D Scene Graph, exposing the objects, platforms, support relations, and free placement regions available to ManiTaskGen. GPTTaskGen-SG-validated selects valid candidates from this scene-graph-conditioned generator. Table~\ref{tab:controlled_baseline_full} reports validity and entity coverage for all four methods. Table~\ref{tab:task_val} compares task validity before post-generation filtering, while Table~\ref{tab:controlled_baseline_full} additionally reports entity coverage and the results of GPTTaskGen-SG-validated.

\begin{table}[htbp]
\centering

\small
\setlength{\tabcolsep}{4pt}
\begin{tabular*}{\linewidth}{@{\extracolsep{\fill}}lccccc@{}}
\toprule
Method & Validity & \multicolumn{4}{c}{Normalized entity coverage (\%)} \\
\cmidrule(lr){3-6}
 & (\%) & Source plat. & Target plat. & Moved obj. & Anchor obj. \\
\midrule
GPTTaskGen & 29 & 47.2 & 48.8 & 45.6 & 42.5 \\
GPTTaskGen-SG & 68 & 42.2 & 41.7 & \textbf{73.2} & \textbf{71.3} \\
GPTTaskGen-SG-validated & \textbf{100} & 34.5 & 36.1 & 61.2 & 60.5 \\
ManiTaskGen & 94 & \textbf{85.4} & \textbf{87.3} & 72.9 & 70.9 \\
\bottomrule
\end{tabular*}
\caption{\textbf{Validity and entity coverage of single-step tasks generated by four methods.} }
\label{tab:controlled_baseline_full}
\end{table}

\noindent\textbf{Validation and generation budget.} GPTTaskGen-SG-validated is constructed by applying human verification to candidates generated by GPTTaskGen-SG and retaining 100 accepted tasks. Its 100\% validity therefore characterizes the retained set. The comparison uses equal evaluated-set sizes; generation calls, token usage, and computational cost are not controlled.

\noindent\textbf{Coverage definition.} For role $r\in\{\text{source platform},\text{target platform},\text{moved object},\text{anchor object}\}$, normalized unique coverage is
\[
 C_r(D)=\frac{|\{e\in E_r:\ e\text{ is mentioned in role }r\text{ by some }t\in D\}|}{|E_r|}\times100\% ,
\]
where $E_r$ is the common scene-specific eligible-entity set for that role and $D$ is the task set used for measurement. Repeated mentions do not add coverage; a task without an anchor contributes no anchor entity. For each role, all methods use the same eligible-entity set $E_r$. Coverage is computed over all evaluated tasks for each method. 

\noindent\textbf{Interpretation.} Table~\ref{tab:controlled_baseline_full} shows that adding the scene graph improves GPTTaskGen validity from 29\% to 68\%, while ManiTaskGen achieves 94\%. Although the scene-graph-conditioned baseline attains object coverage comparable to ManiTaskGen, its platform coverage remains lower. ManiTaskGen covers 85.4\%/87.3\% of source/target platforms, versus 47.2\%/48.8\% for GPTTaskGen and 42.2\%/41.7\% for GPTTaskGen-SG. The validated baseline covers 34.5\%/36.1\%, showing that selecting valid tasks does not close the platform-coverage gap in this comparison. Thus, the main coverage advantage is broader source- and target-platform coverage, rather than uniformly higher coverage of every entity type.



\clearpage
\subsection{Reliability of Feasibility Filtering and Execution Judging}
\label{appendix_subsec:evaluation_discrepancy}

We distinguish generation-time feasibility filtering from post-execution success judging through two supplementary 100-instance studies. We evaluate three VLM judges accessed through the OpenRouter API. These supplementary studies evaluate judge reliability independently of the ensemble used for dataset generation and the agent benchmark reported in Table~\ref{tab:vlm_results}. Table~\ref{tab:vlm_reliability_full} summarizes ensemble and per-judge agreement with human consensus, while Table~\ref{tab:filtering_vote_bins} breaks filtering results down by feasible-vote count.

\noindent\textbf{Samples and decisions.} For feasibility filtering, we instantiate each of 20 outcome-based templates five times in one ReplicaCAD scene, yielding 100 candidate tasks. Each candidate receives zero to three feasible votes, and the decision is accept if at least two judges vote feasible. Candidates with zero or one feasible vote are also human-assessed; studying only accepted tasks would conceal false negatives. For execution judging, we evaluate 100 outcome-task episodes, with five episodes for each of 20 templates. Three VLM judges independently assess the task and its before/after scene images; an episode is labeled successful if at least two judges vote successful. These are different sample sets addressing different questions.

\noindent\textbf{Human reference and error definitions.} Three human annotators label each study's samples, and the reported human consensus is the reference for agreement and errors. The positive class is \emph{feasible} in filtering and \emph{successful} in execution judging. A false positive accepts an infeasible candidate or marks an unsuccessful episode successful; a false negative rejects a feasible candidate or marks a successful episode unsuccessful. Thus FP/FN are counts relative to human consensus, not rates conditional on the predicted class. These are not comparisons between human and robot task-execution performance.

Human--human reliability is reported separately using Fleiss' $\kappa$ and mean pairwise agreement (MPA). MPA is the mean fraction of labels on which a pair of annotators agrees, averaged over the three pairs; Fleiss' $\kappa$ adjusts multi-rater agreement for agreement expected from label prevalence. Filtering has $\kappa=0.902$, MPA $=0.953$; execution judging has $\kappa=0.851$, MPA $=0.940$. These high values do not make the human reference infallible, but suggest that judge disagreement is not explained solely by disagreement among human raters.

\begin{table}[htbp]
\centering

\small
\setlength{\tabcolsep}{4pt}
\begin{tabular*}{\linewidth}{@{\extracolsep{\fill}}llrrrcc@{}}
\toprule
Stage & Evaluator & Agreement & FP & FN & Human $\kappa$ & MPA \\
\midrule
Filtering & Three-VLM vote, $\geq 2$ accepts & 92\% & 7 & 1 & 0.902 & 0.953 \\
\midrule
Execution & GPT-5.6-Luna & 79\% & 13 & 8 & --- & --- \\
Execution & Claude-Haiku-4.5 & 75\% & 11 & 14 & --- & --- \\
Execution & Gemini-3.5-Flash & 84\% & 6 & 10 & --- & --- \\
Execution & Three-VLM majority & 85\% & 5 & 10 & 0.851 & 0.940 \\
\bottomrule
\end{tabular*}
\caption{\textbf{Agreement between VLM judgments and human consensus for feasibility filtering and execution evaluation.}  Each study contains 100 instances; all execution rows use the same episode set and human reference. Human reliability is repeated only on the ensemble row.}
\label{tab:vlm_reliability_full}
\label{tab:vlm_reliability}
\end{table}

\begin{table}[htbp]
\centering

\begin{tabular}{cccrr}
\toprule
Feasible votes & Decision & Agreement with humans & FP & FN \\
\midrule
0 & Reject & 15/15 & 0 & 0 \\
1 & Reject & 17/18 & 0 & 1 \\
2 & Accept & 29/34 & 5 & 0 \\
3 & Accept & 31/33 & 2 & 0 \\
\midrule
Total & $\geq 2$ accepts & 92/100 & 7 & 1 \\
\bottomrule
\end{tabular}
\caption{\textbf{Feasibility-filtering agreement by the number of feasible votes.}}
\label{tab:filtering_vote_bins}
\end{table}

The 92\% filtering agreement is measured on the 100 evaluated candidates, whose feasible-vote distribution is reported in Table~\ref{tab:filtering_vote_bins}.

\noindent\textbf{Interpretation.} The reported ensemble agreement is lower for execution judging than for feasibility filtering, despite high human--human agreement in both studies. Filtering asks whether required objects, sufficient space, and a potentially achievable target state exist. Execution judging additionally requires detailed before/after comparison and interpretation of semantic outcome goals. This difference is a plausible explanation for the lower execution agreement, rather than a controlled causal finding. Repeated VLM calls may reduce sampling variance without eliminating systematic spatial or semantic errors. We therefore retain human evaluation as the primary Level-4 benchmark measure, use VLM judging only as an auxiliary scalability analysis, and retain structured automatic evaluation for Levels 1--3.


\noindent\textbf{Supplementary analysis of the original benchmark evaluations.}
To complement the agreement results in Table~\ref{tab:vlm_reliability_full}, we retain the original benchmark's comparison of human- and VLM-based success rates and its qualitative judging-error example below. These results concern episodes produced by GPT-5.2, Gemini-3.1-pro, and Claude-4.5-sonnet agents in the original benchmark, rather than the separate 100-episode execution-judging study above.

As discussed in Sec.~\ref{subsubsec:benchmark_results} of the main paper, outcome-based tasks (Level 4) involve subjective goal specifications that pose challenges for automated evaluation. Tab.~\ref{tab:human_vlm_gap} presents the success rates (SR) as judged by either humans or VLM evaluators. We observed an obvious discrepancy between the two sets of results, and the primary failure modes of VLM evaluator include insufficient spatial reasoning and semantic over-permissiveness. The table's rows identify the agents that executed the tasks, not the judges in Table~\ref{tab:vlm_reliability_full}; its SR gap is an aggregate percentage-point difference, not the fraction of episodes with disagreeing judgments.

\begin{table}[htbp]
  \centering
  \small
  \vspace{5mm}
    \begin{tabular}{lccc}
    \toprule
    \textbf{Model} & \textbf{Human-based SR} & \textbf{VLM-based SR} & \textbf{Discrepancy} \\
    \midrule
    GPT-5.2 & 30\% & 34\% & 4 \\
    Gemini-3.1-pro & 33\% & 38\% & 5\\
    Claude-4.5-sonnet & 35\% & 40\% & 5\\
    \bottomrule
    \end{tabular}%
    \vspace{2mm}

     \caption{\textbf{Comparison between Human and VLM-based evaluation on Level-4 tasks.}
    }
    \vspace{5mm}
    \label{tab:human_vlm_gap} 
\end{table}

The original benchmark analysis reported human--VLM disagreement on 22.7\% of evaluated episodes from these three agents. Among the divergent cases, 47.1\% were identified as VLM errors upon secondary manual inspection; the remaining divergent cases were deemed ambiguous but acceptable. This conditional manual-review summary is distinct from the FP/FN counts relative to human consensus in the supplementary studies above.

\begin{figure}[htbp]
  \centering
  \vspace{5mm}
  \begin{minipage}[b]{0.45\textwidth}
    \includegraphics[width=\textwidth]{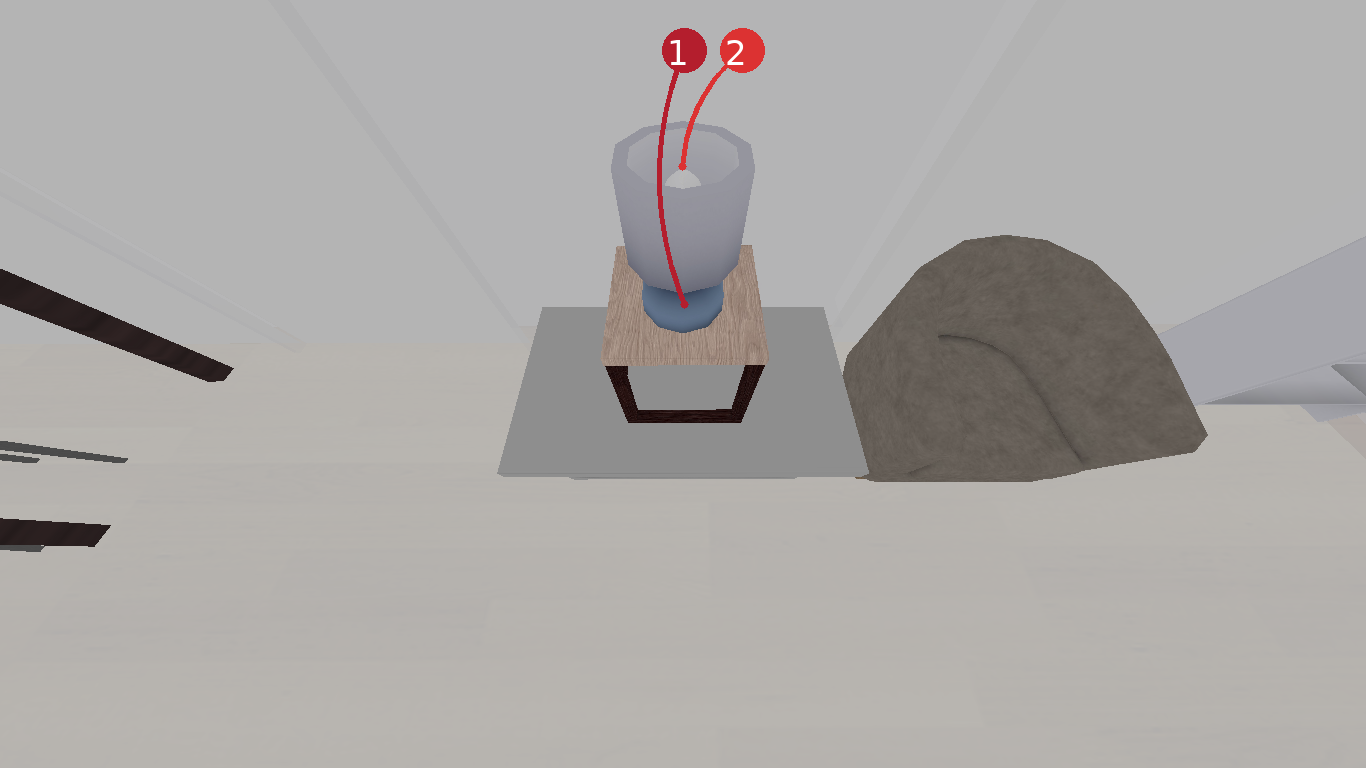}

    \label{fig:init}
  \end{minipage}
  \hfill
  \begin{minipage}[b]{0.45\textwidth}
    \includegraphics[width=\textwidth]{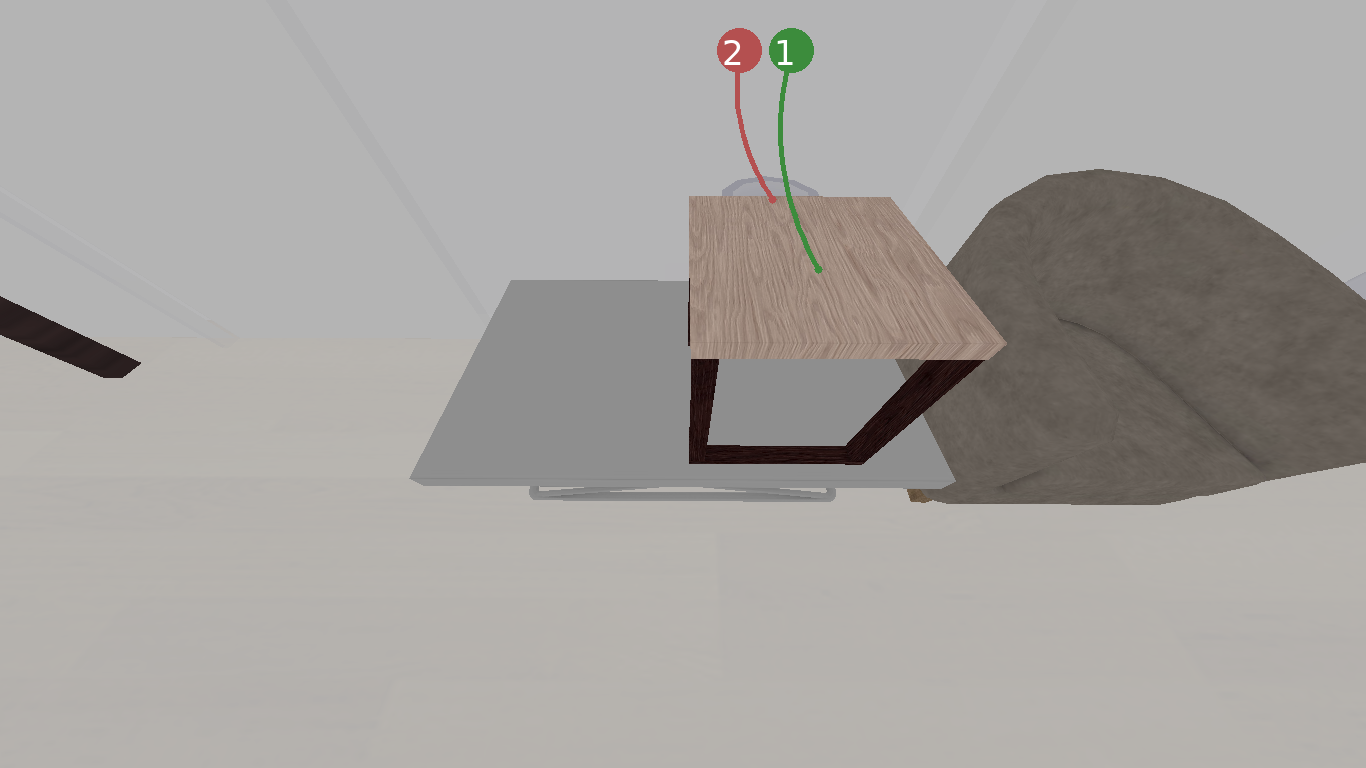}
 
    \label{fig:finish}
  \end{minipage}
  \caption{\textbf{An example of VLM evaluation error. Left: Initial state with objects at the center of table. Right: Final state where objects mistakenly placed on the right instead of left. }}
  \label{fig:obt_failure_example}
\end{figure}

To further illustrate the failure modes, Fig.~\ref{fig:obt_failure_example} provides a qualitative example of a VLM evaluation error. In this instance, the agent was tasked with \textit{"Move the objects from the right side of table and place them on the left."}. The VLM evaluator marked this task as successful, which reveals its insufficient spatial understanding. This example is retained from the original analysis as an illustration of a judging failure, not presented as a sample from either new 100-instance study.

\newpage

\subsection{Examples of Complete Benchmark Episodes}
\label{appendix_subsec:benchmark_example}

Below we display two full episode histories of benchmarking experiments elaborated in Sec.~\ref{subsubsec:benchmark_results}. In Sec.~\ref{appendix_subsubsec:success_episode}, we present a successful episode performed by Claude-3.7-sonnet in the ReplicaCAD scene. In Sec.~\ref{appendix_subsubsec:fail_episode}, we show a failed episode performed by Gemini-2.5-pro in the AI2THOR scene.

\subsubsection{Successful Episode}
\label{appendix_subsubsec:success_episode}
The following episode shows how claude-3.7-sonnet performed a task successfully in ReplicaCAD. When encountered perceptual ambiguity,  it repeatedly picking up the possible objects, and immediately use \textit{place\_r} to place it down when recognized. Eventually it picks up the correct item at the fourth attempt. Then it went to the destination, asked for the receptacle pictures and placed it down successfully.

\vspace{1cm}
\begin{figure}[h]
    \centering
    \includegraphics[width=\textwidth]{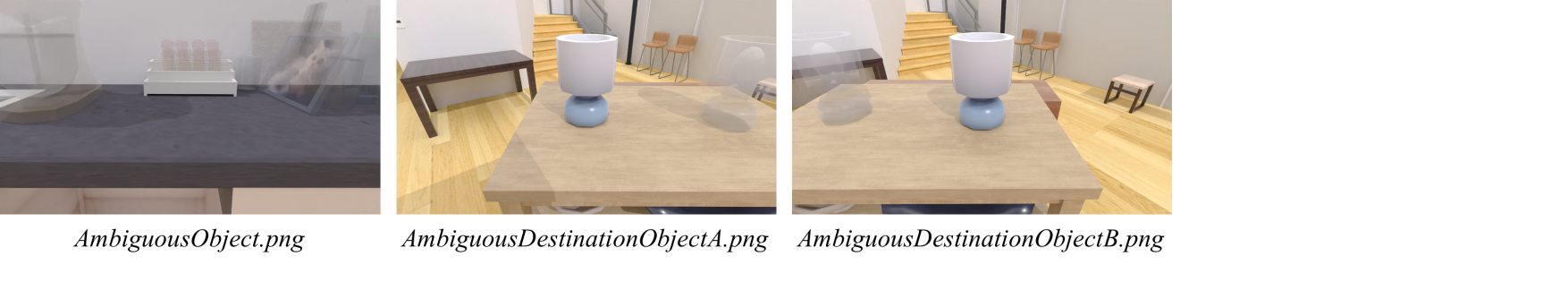}
     \vspace{-1.5em}
    \caption{\textbf{Images Given at the Step 0 of the Successful Episode.}}
    \label{fig:success_step0}
\end{figure}

\vspace{1cm}
\begin{figure}[h]
    \centering
    \includegraphics[width=\textwidth]{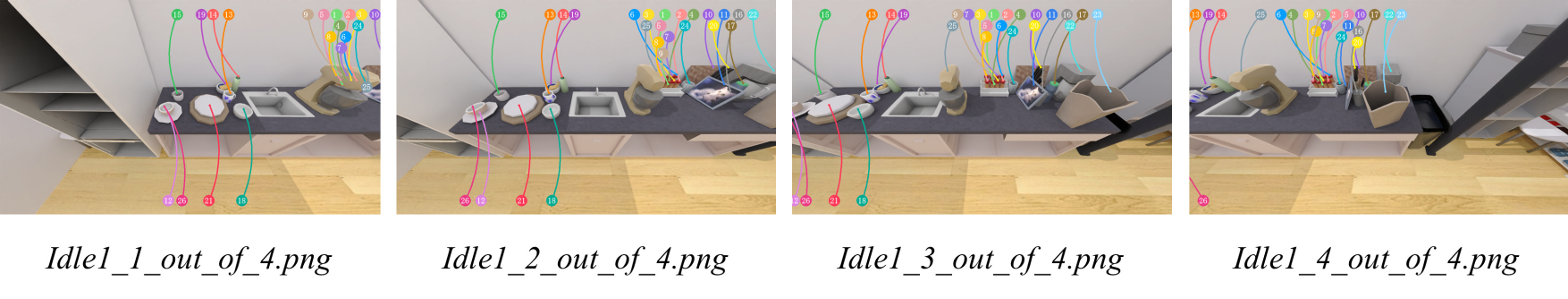}
     \vspace{-1.5em}
    \caption{\textbf{Images Given after Step 1 of the Successful Episode.}}
    \label{fig:success_step1}
\end{figure}

\vspace{1cm}
\begin{figure}[h]
    \centering
    \includegraphics[width=\textwidth]{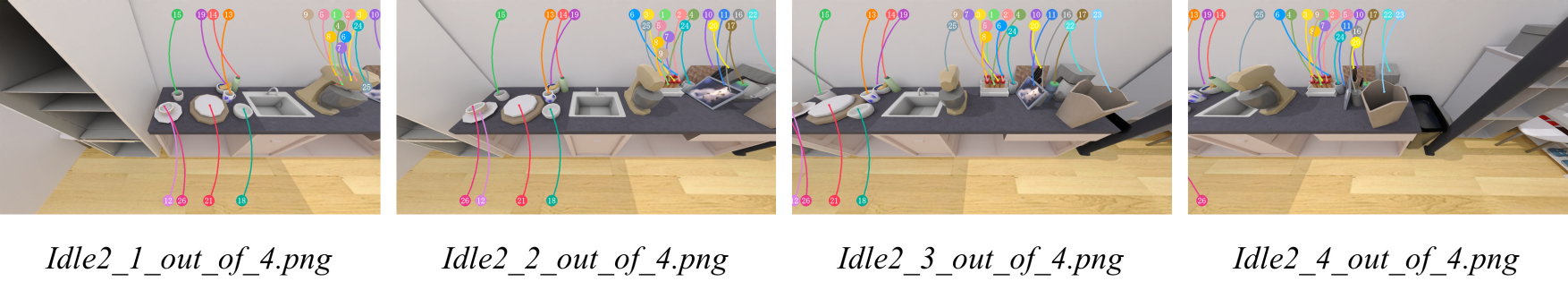}
     \vspace{-1.5em}
    \caption{\textbf{Images Given after Step 2 of the Successful Episode.}}
    \label{fig:success_step2}
\end{figure}

\begin{figure}
    \centering
    \includegraphics[width=\textwidth]{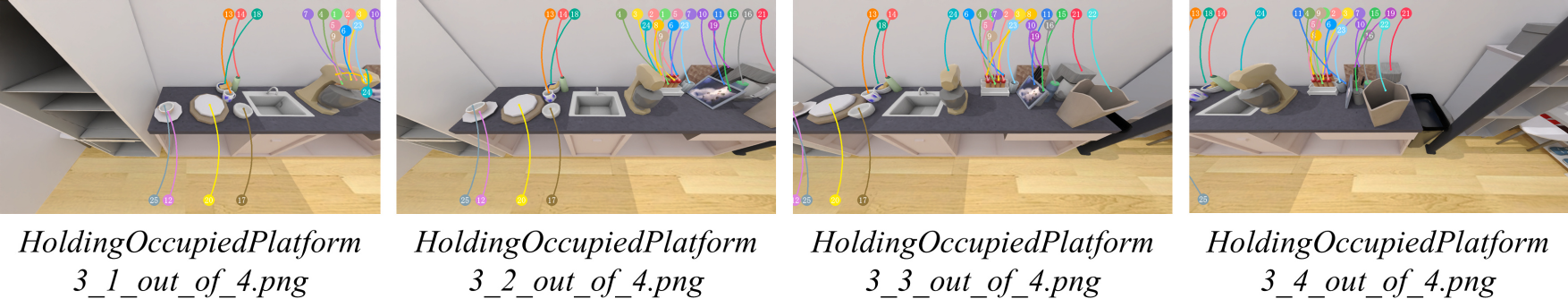}
     \vspace{-1.5em}
    \caption{\textbf{Images Given after Step 3 of the Successful Episode.}}
    \label{fig:success_step3}
\end{figure}

\begin{figure}
    \centering
    \includegraphics[width=\textwidth]{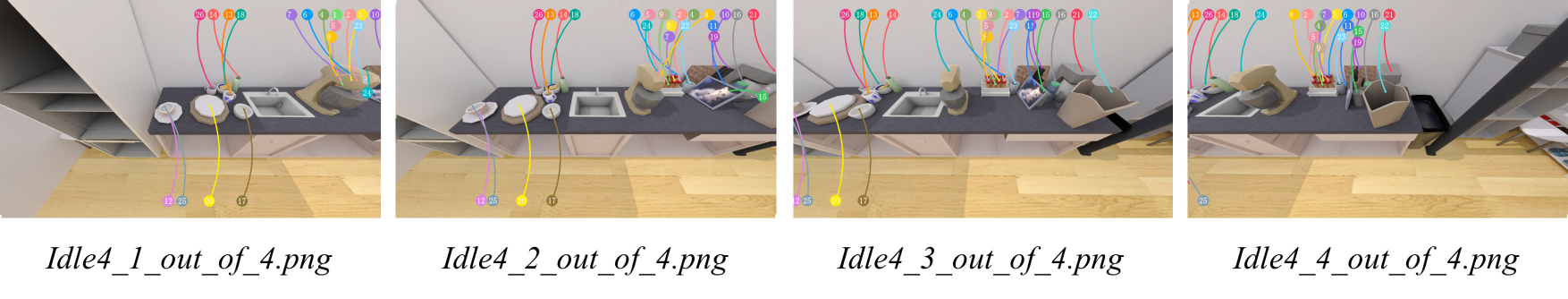}
     \vspace{-1.5em}
    \caption{\textbf{Images Given after Step 4 of the Successful Episode.}}
    \label{fig:success_step4}
\end{figure}

\begin{figure}
    \centering
    \includegraphics[width=\textwidth]{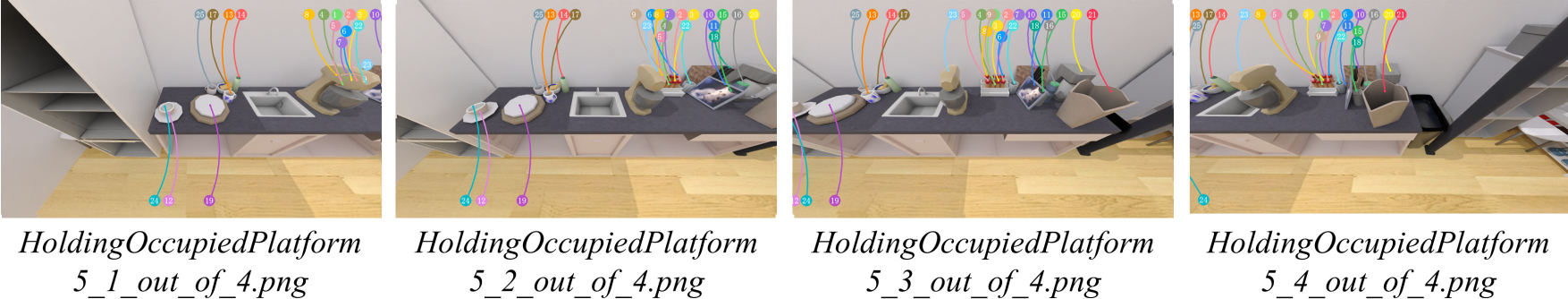}
     \vspace{-1.5em}
    \caption{\textbf{Images Given after Step 5 of the Successful Episode.}}
    \label{fig:success_step5}
\end{figure}

\begin{figure}
    \centering
    \includegraphics[width=\textwidth]{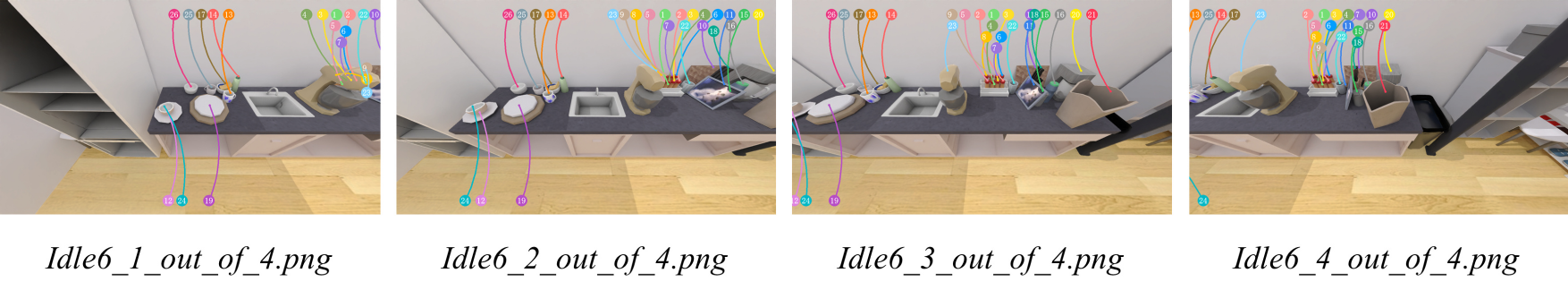}
     \vspace{-1.5em}
    \caption{\textbf{Images Given after Step 6 of the Successful Episode.}}
    \label{fig:success_step6}
\end{figure}

\begin{figure}
    \centering
    \includegraphics[width=\textwidth]{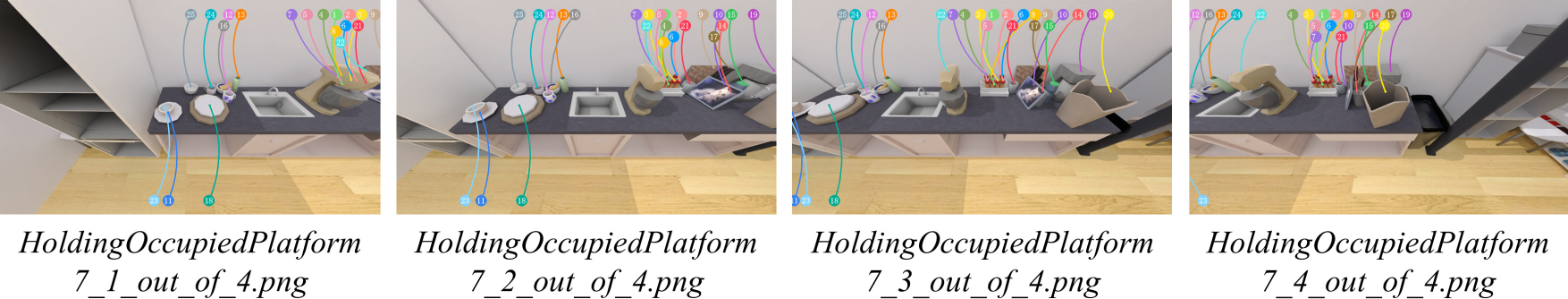}
     \vspace{-1.5em}
    \caption{\textbf{Images Given after Step 7 of the Successful Episode.}}
    \label{fig:success_step7}
\end{figure}

\begin{figure}
    \centering
    \includegraphics[width=\textwidth]{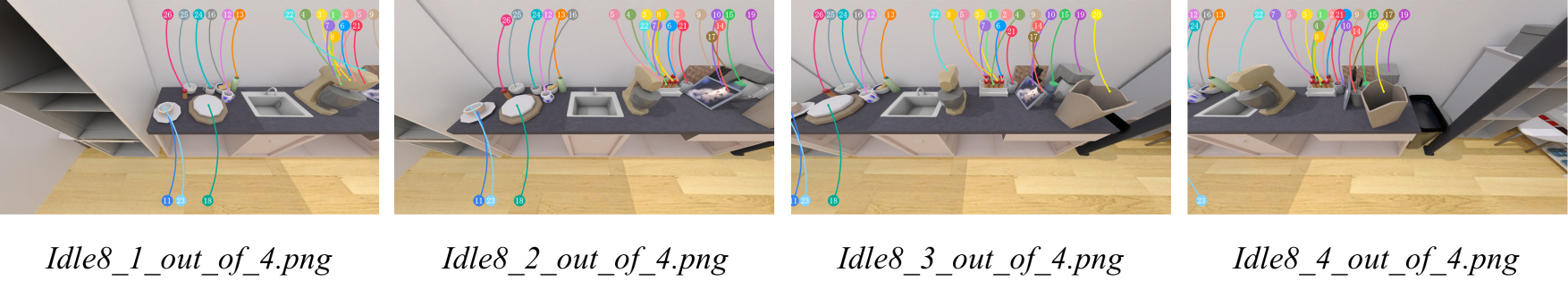}
     \vspace{-1.5em}
    \caption{\textbf{Images Given after Step 8 of the Successful Episode.}}
    \label{fig:success_step8}
\end{figure}

\begin{figure}
    \centering
    \includegraphics[width=\textwidth]{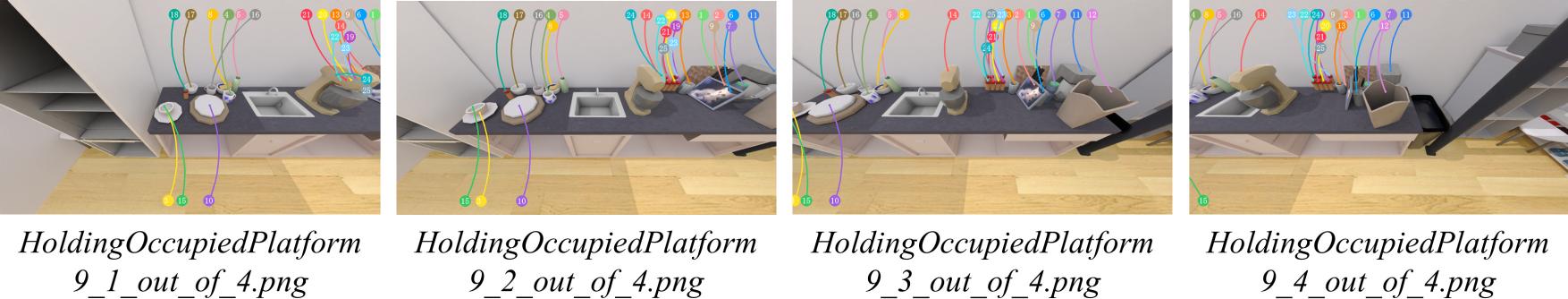}
     \vspace{-1.5em}
    \caption{\textbf{Images Given after Step 9 of the Successful Episode.}}
    \label{fig:success_step9}
\end{figure}

\begin{figure}
    \centering
    \includegraphics[width=\textwidth]{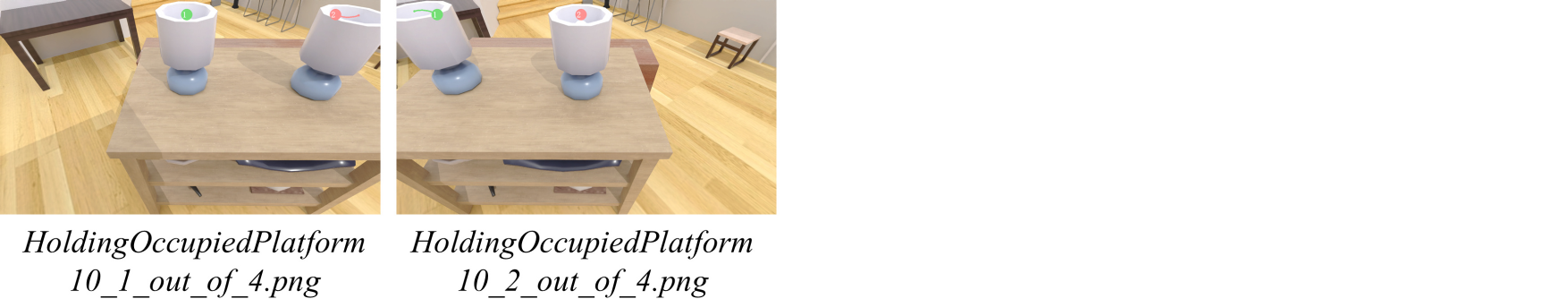}
     \vspace{-1.5em}
    \caption{\textbf{Images Given after Step 10 of the Successful Episode.}}
    \label{fig:success_step10}
\end{figure}

\begin{figure}
    \centering
    \includegraphics[width=\textwidth]{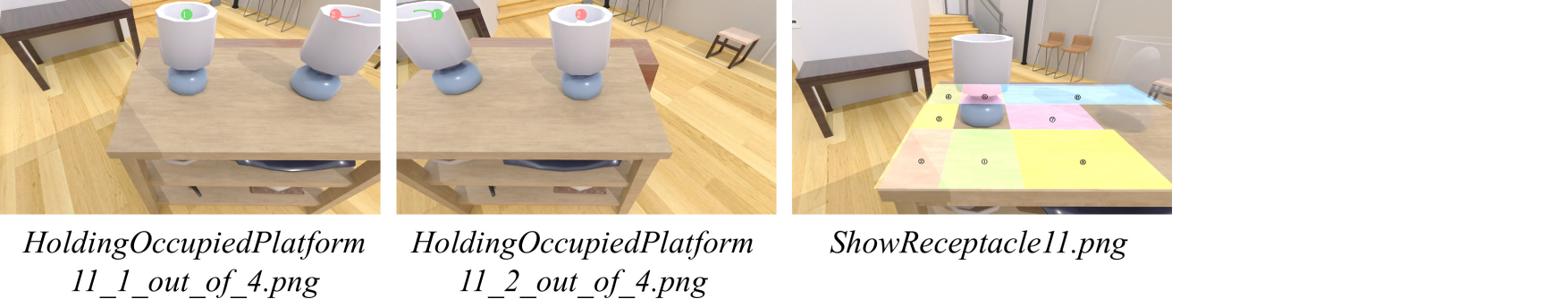}
     \vspace{-1.5em}
    \caption{\textbf{Images Given after Step 11 of the Successful Episode.}}
    \label{fig:success_step11}
\end{figure}

\begin{figure}
    \centering
    \includegraphics[width=\textwidth]{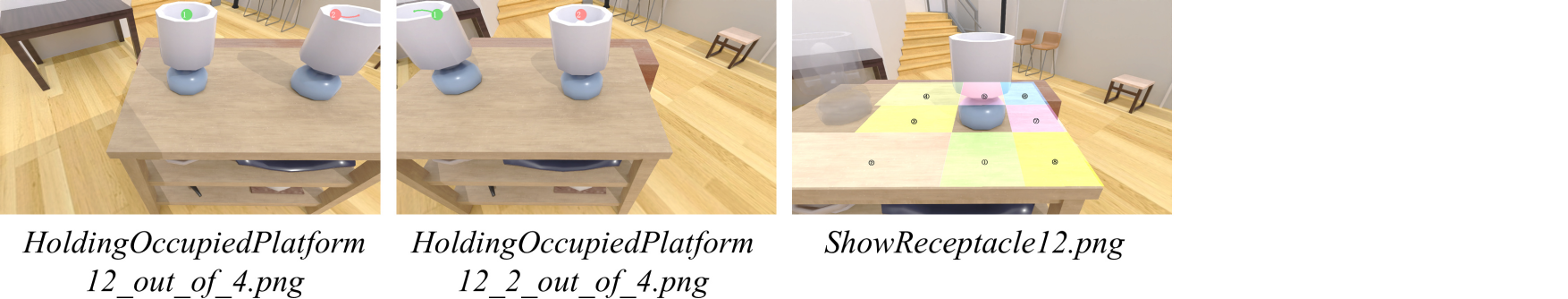}
     \vspace{-1.5em}
    \caption{\textbf{Images Given after Step 12 of the Successful Episode.}}
    \label{fig:success_step12}
\end{figure}

\begin{figure}
    \centering
    \includegraphics[width=\textwidth]{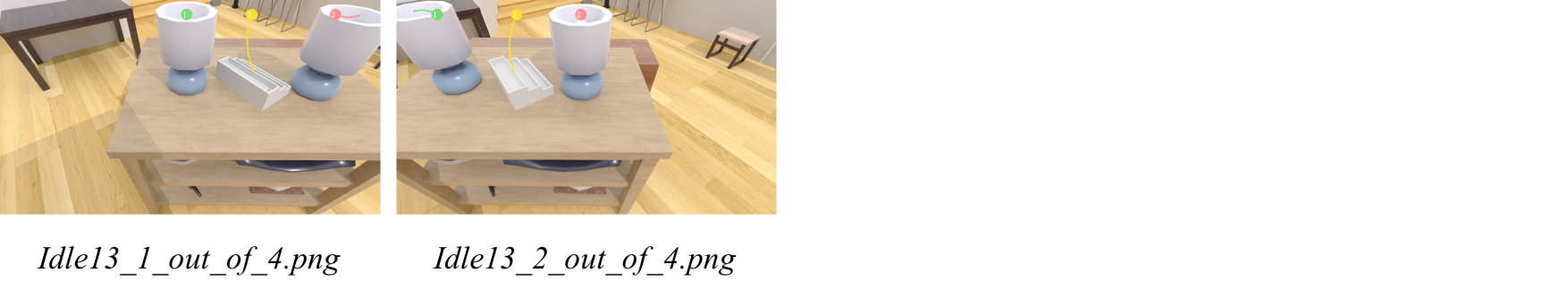}
    \vspace{-1.5em}
    \caption{\textbf{Images Given after Step 13 of the Successful Episode.}}
    \label{fig:success_step13}
\end{figure}

\clearpage

\onecolumn
\begin{tcolorbox}
[
  title={Success Episode by Claude-3.7-sonnet},
  width=\textwidth,
  breakable,
  boxrule=0.5pt,
  fonttitle=\bfseries,
  halign=flush left
]

System Message:You are a household assistant robot equipped with a vision system and manipulation capabilities. You need to complete a task in a 3D environment, mainly involving moving, picking up, and placing objects. Your goal is to move an object from one place to another correctly according to instructions. Please strictly follow the given Action Space for operations.

At the beginning of each task, you will be informed of a specific task to complete. Throughout the task execution, you will transition between multiple states and have access to a set of possible actions (action space).

You may receive one of the following task types involving one object and another object or a platform. ({an object/another object/a platform} are the name of objects or platforms, {dir} is one of the 8 directions):

- move {an object} to {a platform}.
- move {an object} to the {dir} of {a platform}.
- move {an object} around {another object}.
- move {an object} to the {dir} of {another object}.
- move {an object} between {another object} and {yet another object}.

Understanding directions:
Directions like 'front', 'rear', 'left', and 'right' are defined relative to the robot's perspective and the heading of the object or platform you are interacting with. The heading of a ground object is not fixed and depends on where the robot is standing and facing.

Empty Surfaces (Platforms): When interacting with an empty platform, imagine it divided into a 3x3 grid. The direction aligned with the heading (also, the FARTHEST middle space to you, NOT THE NEAREST MIDDLE SPACE) is labeled "front," with the remaining regions proceeding counterclockwise as "front-left," "left," "rear-left," "rear," "rear-right," "right," and "front-right."  The grid is conceptually numbered as follows, based on these directions:

  [front-left, front, front-right],
  
  [left, center, right],
  
  [rear-left, rear, rear-right]

 Occupied Surfaces (Objects on Platforms): For objects on platforms, imagine extending the edges of the object's bounding box to the platform's boundaries. The resulting regions define the object's 8 receptacles, and the direction is defined similarly.
   Task success criteria:

- move {an object} to {a platform}: The task is considered successful only when {an object} is placed anywhere on the specified platform.

- move {an object} to the {dir} of {a platform}: The task is considered successful only when {an object} is placed on the platform and the geometric center of the placement is in the {dir} direction relative to the platform. 

- move {an object} around {another object}: The task is considered successful only when {an object} is placed in any of the receptacles surrounding {another object}.

- move {an object} to the {dir} of {another object}: The task is considered successful only when {an object} is placed in the receptacles surrounding {another object} and overlaps with the {dir} direction.

- move {an object} between {another object} and {yet another object}: The task is considered successful only when {an object} is placed such that it intersects with receptacles in a pair of opposite directions between {another object} and {yet another object}.

Your action space, includes walking to platforms, picking up items, putting items down, rotating direction, getting information about directions around an object; CALL\_END, placing an item in a specific direction on a platform, in a direction relative to another object, or placing it anywhere. Whenever you're about to choose an action, I will provide your current action space. If you choose an action that cannot be performed, such as trying to pick up an object repeatedly or attempting to place an object in a space that's too small, you'll receive a notification.

Picking up an item: Corresponds to the command "pick\_object\_\{object number\}\_of\_current\_platform".

Walking to a platform: Corresponds to the command "go\_to\_{platform\_name}".

Rotating direction: Corresponds to the command "change\_view".

Getting an image of an object: Corresponds to the command "show\_receptacle\_of\_object\_\{object\_number\}\_of\_current\_platform".

Putting down your holding item on empty platform: Corresponds to the command "place\_s\_[({region\_idx1},{region\_idx2})...]" where {dir\_idx1}, {dir\_idx2} are NUMBERS, indicating the index of the receptacles; Putting down your item on an occupied platform: Corresponds to the command "place\_s\_[({object\_idx1},{dir\_idx1})...]",where {object\_idx1}, {dir\_idx1} are NUMBERS, indicating the object and its corresponding receptacles. For example,  "place\_s\_[(1,2), (2,5)]" means putting down your item at the union of object 1's 2nd receptacles and object 2's 5th receptacles.Note that for placement action, the system will attempt to find a valid placement area that intersects with all the regions you selected. If it fails, it will place the object anywhere within your chosen space and you'll receive a notification.

You may be shown various types of images:

0) At the beginning of a task, if the task involves objects on a platform that contains other objects of the same type (e.g., the object is a book on a bookshelf with many books), you will receive a notification and images of these objects to help you distinguish them.

1) Whenever you move to a platform, you will see an image of the platform from your current perspective (or multiple images for longer platforms, indicating you can move horizontally to better observe the platform).
If the platform has no objects, a 3x3 grid will be marked on the platform to help you place objects.
If there are objects on the platform, guiding lines will indicate the object numbers.

2) If there are objects on a platform, you can use "show\_receptable\_of\_object\_x\_of\_current\_platform" to get a focused image of that object with numbered receptacles around it. This allows you to see the available receptacles around the object.

Here are some useful suggestions for you to better finish the task: 

1. Your extra actions do not affect the success of the task, you only need to keep your steps within the total step limit.

2. When you want to place an object but cannot find enough space, you can try changing your standing position to find the correct orientation, try combining multiple receptacles, or remove objects that are blocking your way first(although you generally do not need to do this). Specifically, for the task asking you put object to empty platforms, try combining adjacent receptacles may be very useful.

3. For tasks that involve placing objects near other objects, please remember to get a focused image of the target object and examine it carefully. This is important because the receptacles may not be intuitive. 

4. Important: Don't output anything other than the action you want to take. You will only receive the same hint informing you your invalid action. If you make mistakes, just start output correct actions from the next prompt without any apologize.

[*See Fig.~\ref{fig:success_step0}]

\textbf{The task involves identical objects. To help you distinguish them, we have provided images of the relevant objects.}

\textcolor{purple}
{Here are 1 images, named [AmbiguousObject.png'] respectively.
Their description is as follows: This image, AmbiguousObject.png, is an image of the object you are going to move, with its adjacent objects transparentized. The object is kitchenware\_white\_paper\_towel\_holder.
The object is located on object\_kitchen\_counter\_1\_body\_platform\_1.}

\textcolor{purple}
{Here are 1 images, named [AmbiguousDestinationObjectA.png'] respectively.
Their description is as follows: This image, AmbiguousDestinationObjectA.png, is an image of the object you are going to place an object nearby, with its adjacent objects transparentized. The object is lighting\_blue\_base\_table\_lamp.
The object is located on object\_table\_02\_40\_platform\_2.}

\textcolor{purple}
{Here are 1 images, named [AmbiguousDestinationObjectB.png'] respectively.
Their description is as follows: This image, AmbiguousDestinationObjectB.png, is an image of the object you are going to place an object nearby, with its adjacent objects transparentized. The object is lighting\_blue\_base\_table\_lamp.
The object is located on object\_table\_02\_40\_platform\_2.}

Current task:

Move kitchenware\_white\_paper\_towel\_holder between lighting\_blue\_base\_table\_lamp and lighting\_blue\_base\_table\_lamp

Initially, kitchenware\_white\_paper\_towel\_holder is on object\_kitchen\_counter\_1\_body\_platform\_1, and lighting\_blue\_base\_table\_lamp is on object\_table\_02\_40\_platform\_2, together with lighting\_blue\_base\_table\_lamp.

Steps used: 0/20. You can only take at most 20 steps, so hurry if you've almost used all of them!

Your available action space:

- `CALL\_END`: Call end (terminate the task).
- ['go\_to\_object\_sofa\_10\_platform\_0', 'go\_to\_object\_sofa\_10\_platform\_1', 'go\_to\_object\_sofa\_10\_platform\_2', 'go\_to\_object\_table\_04\_13\_platform\_0', 'go\_to\_object\_chair\_04\_46\_platform\_0', 'go\_to\_object\_chair\_04\_47\_platform\_0', 'go\_to\_object\_chair\_05\_7\_platform\_0', 'go\_to\_object\_chair\_05\_8\_platform\_0', 'go\_to\_object\_stool\_02\_18\_platform\_0', 'go\_to\_object\_stool\_02\_6\_platform\_0', 'go\_to\_object\_rack\_01\_75\_platform\_0', 'go\_to\_object\_rack\_01\_75\_platform\_1', 'go\_to\_object\_rack\_01\_75\_platform\_2', 'go\_to\_object\_kitchen\_counter\_1\_body\_platform\_0', 'go\_to\_object\_kitchen\_counter\_1\_body\_platform\_1', 'go\_to\_object\_fridge\_0\_body\_platform\_1', 'go\_to\_object\_fridge\_0\_body\_platform\_3', 'go\_to\_object\_fridge\_0\_body\_platform\_4', 'go\_to\_object\_fridge\_0\_body\_platform\_5', 'go\_to\_object\_fridge\_0\_body\_platform\_6', 'go\_to\_object\_fridge\_0\_body\_platform\_7', 'go\_to\_object\_bin\_03\_3\_platform\_0', 'go\_to\_object\_wall\_cabinet\_01\_4\_platform\_0', 'go\_to\_object\_wall\_cabinet\_01\_4\_platform\_1', 'go\_to\_object\_wall\_cabinet\_01\_4\_platform\_2', 'go\_to\_object\_wall\_cabinet\_01\_4\_platform\_3', 'go\_to\_object\_wall\_cabinet\_01\_4\_platform\_4', 'go\_to\_object\_wall\_cabinet\_01\_4\_platform\_5', 'go\_to\_object\_wall\_cabinet\_01\_4\_platform\_6', 'go\_to\_object\_table\_03\_14\_platform\_1', 'go\_to\_object\_table\_01\_48\_platform\_0', 'go\_to\_object\_wall\_cabinet\_02\_21\_platform\_0', 'go\_to\_object\_wall\_cabinet\_02\_21\_platform\_1', 'go\_to\_object\_wall\_cabinet\_02\_21\_platform\_2', 'go\_to\_object\_wall\_cabinet\_02\_21\_platform\_3', 'go\_to\_object\_wall\_cabinet\_02\_21\_platform\_4', 'go\_to\_object\_wall\_cabinet\_02\_21\_platform\_5', 'go\_to\_object\_wall\_cabinet\_02\_21\_platform\_6', 'go\_to\_object\_chestOfDrawers\_01\_2\_body\_platform\_0', 'go\_to\_object\_chestOfDrawers\_01\_2\_body\_platform\_1', 'go\_to\_object\_chestOfDrawers\_01\_2\_body\_platform\_2', 'go\_to\_object\_chestOfDrawers\_01\_2\_body\_platform\_3', 'go\_to\_object\_chestOfDrawers\_01\_2\_body\_platform\_4', 'go\_to\_object\_chestOfDrawers\_01\_2\_body\_platform\_5', 'go\_to\_object\_chestOfDrawers\_01\_2\_body\_platform\_6', 'go\_to\_object\_table\_02\_40\_platform\_0', 'go\_to\_object\_table\_02\_40\_platform\_1', 'go\_to\_object\_table\_02\_40\_platform\_2', 'go\_to\_object\_tvstand\_88\_platform\_0', 'go\_to\_object\_cabinet\_3\_body\_platform\_0', 'go\_to\_object\_cabinet\_3\_body\_platform\_1']. Please choose one of the available actions.

All your current available actions are listed as above (all are without quotes and square brackets). Please don't mix up it with the action space of previous steps.

\hdashrule{\linewidth}{0.4pt}{2mm}
\textcolor{blue}{
VLM Response: go\_to\_object\_kitchen\_counter\_1\_body\_platform\_1}

\hdashrule{\linewidth}{0.4pt}{2mm}

User Message:

You are currently at object\_kitchen\_counter\_1\_body\_platform\_1.

And you are holding nothing.

Your available action space:

- ['go\_to\_object\_sofa\_10\_platform\_0', 'go\_to\_object\_sofa\_10\_platform\_1', 'go\_to\_object\_sofa\_10\_platform\_2', 'go\_to\_object\_table\_04\_13\_platform\_0', 'go\_to\_object\_chair\_04\_46\_platform\_0', 'go\_to\_object\_chair\_04\_47\_platform\_0', 'go\_to\_object\_chair\_05\_7\_platform\_0', 'go\_to\_object\_chair\_05\_8\_platform\_0', 'go\_to\_object\_stool\_02\_18\_platform\_0', 'go\_to\_object\_stool\_02\_6\_platform\_0', 'go\_to\_object\_rack\_01\_75\_platform\_0', 'go\_to\_object\_rack\_01\_75\_platform\_1', 'go\_to\_object\_rack\_01\_75\_platform\_2', 'go\_to\_object\_kitchen\_counter\_1\_body\_platform\_0', 'go\_to\_object\_kitchen\_counter\_1\_body\_platform\_1', 'go\_to\_object\_fridge\_0\_body\_platform\_1', 'go\_to\_object\_fridge\_0\_body\_platform\_3', 'go\_to\_object\_fridge\_0\_body\_platform\_4', 'go\_to\_object\_fridge\_0\_body\_platform\_5', 'go\_to\_object\_fridge\_0\_body\_platform\_6', 'go\_to\_object\_fridge\_0\_body\_platform\_7', 'go\_to\_object\_bin\_03\_3\_platform\_0', 'go\_to\_object\_wall\_cabinet\_01\_4\_platform\_0', 'go\_to\_object\_wall\_cabinet\_01\_4\_platform\_1', 'go\_to\_object\_wall\_cabinet\_01\_4\_platform\_2', 'go\_to\_object\_wall\_cabinet\_01\_4\_platform\_3', 'go\_to\_object\_wall\_cabinet\_01\_4\_platform\_4', 'go\_to\_object\_wall\_cabinet\_01\_4\_platform\_5', 'go\_to\_object\_wall\_cabinet\_01\_4\_platform\_6', 'go\_to\_object\_table\_03\_14\_platform\_1', 'go\_to\_object\_table\_01\_48\_platform\_0', 'go\_to\_object\_wall\_cabinet\_02\_21\_platform\_0', 'go\_to\_object\_wall\_cabinet\_02\_21\_platform\_1', 'go\_to\_object\_wall\_cabinet\_02\_21\_platform\_2', 'go\_to\_object\_wall\_cabinet\_02\_21\_platform\_3', 'go\_to\_object\_wall\_cabinet\_02\_21\_platform\_4', 'go\_to\_object\_wall\_cabinet\_02\_21\_platform\_5', 'go\_to\_object\_wall\_cabinet\_02\_21\_platform\_6', 'go\_to\_object\_chestOfDrawers\_01\_2\_body\_platform\_0', 'go\_to\_object\_chestOfDrawers\_01\_2\_body\_platform\_1', 'go\_to\_object\_chestOfDrawers\_01\_2\_body\_platform\_2', 'go\_to\_object\_chestOfDrawers\_01\_2\_body\_platform\_3', 'go\_to\_object\_chestOfDrawers\_01\_2\_body\_platform\_4', 'go\_to\_object\_chestOfDrawers\_01\_2\_body\_platform\_5', 'go\_to\_object\_chestOfDrawers\_01\_2\_body\_platform\_6', 'go\_to\_object\_table\_02\_40\_platform\_0', 'go\_to\_object\_table\_02\_40\_platform\_1', 'go\_to\_object\_table\_02\_40\_platform\_2', 'go\_to\_object\_tvstand\_88\_platform\_0', 'go\_to\_object\_cabinet\_3\_body\_platform\_0', 'go\_to\_object\_cabinet\_3\_body\_platform\_1']
- ['pick\_object\_1\_of\_current\_platform', 'pick\_object\_2\_of\_current\_platform', 'pick\_object\_3\_of\_current\_platform', 'pick\_object\_4\_of\_current\_platform', 'pick\_object\_5\_of\_current\_platform', 'pick\_object\_6\_of\_current\_platform', 'pick\_object\_7\_of\_current\_platform', 'pick\_object\_8\_of\_current\_platform', 'pick\_object\_9\_of\_current\_platform', 'pick\_object\_10\_of\_current\_platform', 'pick\_object\_11\_of\_current\_platform', 'pick\_object\_12\_of\_current\_platform', 'pick\_object\_13\_of\_current\_platform', 'pick\_object\_14\_of\_current\_platform', 'pick\_object\_15\_of\_current\_platform', 'pick\_object\_16\_of\_current\_platform', 'pick\_object\_17\_of\_current\_platform', 'pick\_object\_18\_of\_current\_platform', 'pick\_object\_19\_of\_current\_platform', 'pick\_object\_20\_of\_current\_platform', 'pick\_object\_21\_of\_current\_platform', 'pick\_object\_22\_of\_current\_platform', 'pick\_object\_23\_of\_current\_platform', 'pick\_object\_24\_of\_current\_platform', 'pick\_object\_25\_of\_current\_platform', 'pick\_object\_26\_of\_current\_platform']
- ['show\_receptacle\_of\_object\_1\_of\_current\_platform', 'show\_receptacle\_of\_object\_2\_of\_current\_platform', 'show\_receptacle\_of\_object\_3\_of\_current\_platform', 'show\_receptacle\_of\_object\_4\_of\_current\_platform', 'show\_receptacle\_of\_object\_5\_of\_current\_platform', 'show\_receptacle\_of\_object\_6\_of\_current\_platform', 'show\_receptacle\_of\_object\_7\_of\_current\_platform', 'show\_receptacle\_of\_object\_8\_of\_current\_platform', 'show\_receptacle\_of\_object\_9\_of\_current\_platform', 'show\_receptacle\_of\_object\_10\_of\_current\_platform', 'show\_receptacle\_of\_object\_11\_of\_current\_platform', 'show\_receptacle\_of\_object\_12\_of\_current\_platform', 'show\_receptacle\_of\_object\_13\_of\_current\_platform', 'show\_receptacle\_of\_object\_14\_of\_current\_platform', 'show\_receptacle\_of\_object\_15\_of\_current\_platform', 'show\_receptacle\_of\_object\_16\_of\_current\_platform', 'show\_receptacle\_of\_object\_17\_of\_current\_platform', 'show\_receptacle\_of\_object\_18\_of\_current\_platform', 'show\_receptacle\_of\_object\_19\_of\_current\_platform', 'show\_receptacle\_of\_object\_20\_of\_current\_platform', 'show\_receptacle\_of\_object\_21\_of\_current\_platform', 'show\_receptacle\_of\_object\_22\_of\_current\_platform', 'show\_receptacle\_of\_object\_23\_of\_current\_platform', 'show\_receptacle\_of\_object\_24\_of\_current\_platform', 'show\_receptacle\_of\_object\_25\_of\_current\_platform', 'show\_receptacle\_of\_object\_26\_of\_current\_platform']
- `CALL\_END`: Call end (terminate the task).
- `change\_view`: Change the viewing angle to observe the scene from another perspective.

Choose one of the available actions.

All your current available actions are listed as above (all are without quotes and square brackets). Please don't mix up it with the action space of previous steps.

Here are some useful suggestions for you to better finish the task: 

1. Your extra actions do not affect the success of the task, you only need to keep your steps within the total step limit.

2. When you want to place an object but cannot find enough space, you can try changing your standing position to find the correct orientation, try combining multiple receptacles, or remove objects that are blocking your way first(although you generally do not need to do this). Specifically, for the task asking you put object to empty platforms, try combining adjacent receptacles may be very useful.

3. For tasks that involve placing objects near other objects, please remember to get a focused image of the target object and examine it carefully. This is important because the receptacles may not be intuitive. 

4. Important: Don't output anything other than the action you want to take. You will only receive the same hint informing you your invalid action. If you make mistakes, just start output correct actions from the next prompt without any apologize.

Current task:

Move kitchenware\_white\_paper\_towel\_holder between lighting\_blue\_base\_table\_lamp and lighting\_blue\_base\_table\_lamp

Steps used: 1/20. You can only take at most 20 steps, so hurry if you've almost used all of them!

[*See Fig.~\ref{fig:success_step1}]

\textcolor{purple}
{Here are 4 images, named ['Idle\_1\_1\_out\_of\_4.png', 'Idle\_1\_2\_out\_of\_4.png', 'Idle\_1\_3\_out\_of\_4.png', 'Idle\_1\_4\_out\_of\_4.png'] respectively.}
\textcolor{purple}
{Their description is as follows: These images show the platform you are currently on in one view, with regions or objects marked with numbers for you to place objects accurately. The platform is object\_kitchen\_counter\_1\_body\_platform\_1.}
\textcolor{purple}
{If the platform has no objects, a 3x3 grid will be marked on the platform to help you place objects, with each grid marked in numbers from 1 to 9. If there are objects on the platform, guiding lines will indicate the object numbers. Note that  if you see multiple images, it means the platform is too long to be captured in one image, and it's guaranteed that the same object will marked in THE SAME COLOR GUIDELINES AND THE SAME NUMBER. If you think you can't determine the number of an item in one picture, you can go through other pictures and see if this item is marked more clearly.
}

\hdashrule{\linewidth}{0.4pt}{2mm}

\textcolor{blue}{VLM Response: change\_view}

\hdashrule{\linewidth}{0.4pt}{2mm}

User Message:

\textbf{Unable to rotate to another view. The platform you at only have 1 walkable place for you.}

You are currently at object\_kitchen\_counter\_1\_body\_platform\_1.

And you are holding nothing.

Your available action space:

- ['go\_to\_object\_sofa\_10\_platform\_0', 'go\_to\_object\_sofa\_10\_platform\_1', 'go\_to\_object\_sofa\_10\_platform\_2', 'go\_to\_object\_table\_04\_13\_platform\_0', 'go\_to\_object\_chair\_04\_46\_platform\_0', 'go\_to\_object\_chair\_04\_47\_platform\_0', 'go\_to\_object\_chair\_05\_7\_platform\_0', 'go\_to\_object\_chair\_05\_8\_platform\_0', 'go\_to\_object\_stool\_02\_18\_platform\_0', 'go\_to\_object\_stool\_02\_6\_platform\_0', 'go\_to\_object\_rack\_01\_75\_platform\_0', 'go\_to\_object\_rack\_01\_75\_platform\_1', 'go\_to\_object\_rack\_01\_75\_platform\_2', 'go\_to\_object\_kitchen\_counter\_1\_body\_platform\_0', 'go\_to\_object\_kitchen\_counter\_1\_body\_platform\_1', 'go\_to\_object\_fridge\_0\_body\_platform\_1', 'go\_to\_object\_fridge\_0\_body\_platform\_3', 'go\_to\_object\_fridge\_0\_body\_platform\_4', 'go\_to\_object\_fridge\_0\_body\_platform\_5', 'go\_to\_object\_fridge\_0\_body\_platform\_6', 'go\_to\_object\_fridge\_0\_body\_platform\_7', 'go\_to\_object\_bin\_03\_3\_platform\_0', 'go\_to\_object\_wall\_cabinet\_01\_4\_platform\_0', 'go\_to\_object\_wall\_cabinet\_01\_4\_platform\_1', 'go\_to\_object\_wall\_cabinet\_01\_4\_platform\_2', 'go\_to\_object\_wall\_cabinet\_01\_4\_platform\_3', 'go\_to\_object\_wall\_cabinet\_01\_4\_platform\_4', 'go\_to\_object\_wall\_cabinet\_01\_4\_platform\_5', 'go\_to\_object\_wall\_cabinet\_01\_4\_platform\_6', 'go\_to\_object\_table\_03\_14\_platform\_1', 'go\_to\_object\_table\_01\_48\_platform\_0', 'go\_to\_object\_wall\_cabinet\_02\_21\_platform\_0', 'go\_to\_object\_wall\_cabinet\_02\_21\_platform\_1', 'go\_to\_object\_wall\_cabinet\_02\_21\_platform\_2', 'go\_to\_object\_wall\_cabinet\_02\_21\_platform\_3', 'go\_to\_object\_wall\_cabinet\_02\_21\_platform\_4', 'go\_to\_object\_wall\_cabinet\_02\_21\_platform\_5', 'go\_to\_object\_wall\_cabinet\_02\_21\_platform\_6', 'go\_to\_object\_chestOfDrawers\_01\_2\_body\_platform\_0', 'go\_to\_object\_chestOfDrawers\_01\_2\_body\_platform\_1', 'go\_to\_object\_chestOfDrawers\_01\_2\_body\_platform\_2', 'go\_to\_object\_chestOfDrawers\_01\_2\_body\_platform\_3', 'go\_to\_object\_chestOfDrawers\_01\_2\_body\_platform\_4', 'go\_to\_object\_chestOfDrawers\_01\_2\_body\_platform\_5', 'go\_to\_object\_chestOfDrawers\_01\_2\_body\_platform\_6', 'go\_to\_object\_table\_02\_40\_platform\_0', 'go\_to\_object\_table\_02\_40\_platform\_1', 'go\_to\_object\_table\_02\_40\_platform\_2', 'go\_to\_object\_tvstand\_88\_platform\_0', 'go\_to\_object\_cabinet\_3\_body\_platform\_0', 'go\_to\_object\_cabinet\_3\_body\_platform\_1']
- ['pick\_object\_1\_of\_current\_platform', 'pick\_object\_2\_of\_current\_platform', 'pick\_object\_3\_of\_current\_platform', 'pick\_object\_4\_of\_current\_platform', 'pick\_object\_5\_of\_current\_platform', 'pick\_object\_6\_of\_current\_platform', 'pick\_object\_7\_of\_current\_platform', 'pick\_object\_8\_of\_current\_platform', 'pick\_object\_9\_of\_current\_platform', 'pick\_object\_10\_of\_current\_platform', 'pick\_object\_11\_of\_current\_platform', 'pick\_object\_12\_of\_current\_platform', 'pick\_object\_13\_of\_current\_platform', 'pick\_object\_14\_of\_current\_platform', 'pick\_object\_15\_of\_current\_platform', 'pick\_object\_16\_of\_current\_platform', 'pick\_object\_17\_of\_current\_platform', 'pick\_object\_18\_of\_current\_platform', 'pick\_object\_19\_of\_current\_platform', 'pick\_object\_20\_of\_current\_platform', 'pick\_object\_21\_of\_current\_platform', 'pick\_object\_22\_of\_current\_platform', 'pick\_object\_23\_of\_current\_platform', 'pick\_object\_24\_of\_current\_platform', 'pick\_object\_25\_of\_current\_platform', 'pick\_object\_26\_of\_current\_platform']
- ['show\_receptacle\_of\_object\_1\_of\_current\_platform', 'show\_receptacle\_of\_object\_2\_of\_current\_platform', 'show\_receptacle\_of\_object\_3\_of\_current\_platform', 'show\_receptacle\_of\_object\_4\_of\_current\_platform', 'show\_receptacle\_of\_object\_5\_of\_current\_platform', 'show\_receptacle\_of\_object\_6\_of\_current\_platform', 'show\_receptacle\_of\_object\_7\_of\_current\_platform', 'show\_receptacle\_of\_object\_8\_of\_current\_platform', 'show\_receptacle\_of\_object\_9\_of\_current\_platform', 'show\_receptacle\_of\_object\_10\_of\_current\_platform', 'show\_receptacle\_of\_object\_11\_of\_current\_platform', 'show\_receptacle\_of\_object\_12\_of\_current\_platform', 'show\_receptacle\_of\_object\_13\_of\_current\_platform', 'show\_receptacle\_of\_object\_14\_of\_current\_platform', 'show\_receptacle\_of\_object\_15\_of\_current\_platform', 'show\_receptacle\_of\_object\_16\_of\_current\_platform', 'show\_receptacle\_of\_object\_17\_of\_current\_platform', 'show\_receptacle\_of\_object\_18\_of\_current\_platform', 'show\_receptacle\_of\_object\_19\_of\_current\_platform', 'show\_receptacle\_of\_object\_20\_of\_current\_platform', 'show\_receptacle\_of\_object\_21\_of\_current\_platform', 'show\_receptacle\_of\_object\_22\_of\_current\_platform', 'show\_receptacle\_of\_object\_23\_of\_current\_platform', 'show\_receptacle\_of\_object\_24\_of\_current\_platform', 'show\_receptacle\_of\_object\_25\_of\_current\_platform', 'show\_receptacle\_of\_object\_26\_of\_current\_platform']
- `CALL\_END`: Call end (terminate the task).
- `change\_view`: Change the viewing angle to observe the scene from another perspective.

Choose one of the available actions.

All your current available actions are listed as above (all are without quotes and square brackets). Please don't mix up it with the action space of previous steps.

Here are some useful suggestions for you to better finish the task: 

1. Your extra actions do not affect the success of the task, you only need to keep your steps within the total step limit.

2. When you want to place an object but cannot find enough space, you can try changing your standing position to find the correct orientation, try combining multiple receptacles, or remove objects that are blocking your way first(although you generally do not need to do this). Specifically, for the task asking you put object to empty platforms, try combining adjacent receptacles may be very useful.

3. For tasks that involve placing objects near other objects, please remember to get a focused image of the target object and examine it carefully. This is important because the receptacles may not be intuitive. 

4. Important: Don't output anything other than the action you want to take. You will only receive the same hint informing you your invalid action. If you make mistakes, just start output correct actions from the next prompt without any apologize.

Current task:

Move kitchenware\_white\_paper\_towel\_holder between lighting\_blue\_base\_table\_lamp and lighting\_blue\_base\_table\_lamp

Steps used: 2/20. You can only take at most 20 steps, so hurry if you've almost used all of them!

[*See Fig.~\ref{fig:success_step2}]

\textcolor{purple}
{Here are 4 images, named ['Idle\_2\_1\_out\_of\_4.png', 'Idle\_2\_2\_out\_of\_4.png', 'Idle\_2\_3\_out\_of\_4.png', 'Idle\_2\_4\_out\_of\_4.png'] respectively.}

\textcolor{purple}
{Their description is as follows: These images show the platform you are currently on in one view, with regions or objects marked with numbers for you to place objects accurately. The platform is object\_kitchen\_counter\_1\_body\_platform\_1.}

\textcolor{purple}
{If the platform has no objects, a 3x3 grid will be marked on the platform to help you place objects, with each grid marked in numbers from 1 to 9. If there are objects on the platform, guiding lines will indicate the object numbers. Note that  if you see multiple images, it means the platform is too long to be captured in one image, and it's guaranteed that the same object will marked in THE SAME COLOR GUIDELINES AND THE SAME NUMBER. If you think you can't determine the number of an item in one picture, you can go through other pictures and see if this item is marked more clearly.}

\hdashrule{\linewidth}{0.4pt}{2mm}

\textcolor{blue}
{VLM\_response: pick\_15\_of\_current\_platform}

\hdashrule{\linewidth}{0.4pt}{2mm}

User Message:

You are currently at object\_kitchen\_counter\_1\_body\_platform\_1.

And you are holding kitchenware\_small\_white\_mug.

Your available action space:

-['go\_to\_object\_sofa\_10\_platform\_0', 'go\_to\_object\_sofa\_10\_platform\_1', 'go\_to\_object\_sofa\_10\_platform\_2', 'go\_to\_object\_table\_04\_13\_platform\_0', 'go\_to\_object\_chair\_04\_46\_platform\_0', 'go\_to\_object\_chair\_04\_47\_platform\_0', 'go\_to\_object\_chair\_05\_7\_platform\_0', 'go\_to\_object\_chair\_05\_8\_platform\_0', 'go\_to\_object\_stool\_02\_18\_platform\_0', 'go\_to\_object\_stool\_02\_6\_platform\_0', 'go\_to\_object\_rack\_01\_75\_platform\_0', 'go\_to\_object\_rack\_01\_75\_platform\_1', 'go\_to\_object\_rack\_01\_75\_platform\_2', 'go\_to\_object\_kitchen\_counter\_1\_body\_platform\_0', 'go\_to\_object\_kitchen\_counter\_1\_body\_platform\_1', 'go\_to\_object\_fridge\_0\_body\_platform\_1', 'go\_to\_object\_fridge\_0\_body\_platform\_3', 'go\_to\_object\_fridge\_0\_body\_platform\_4', 'go\_to\_object\_fridge\_0\_body\_platform\_5', 'go\_to\_object\_fridge\_0\_body\_platform\_6', 'go\_to\_object\_fridge\_0\_body\_platform\_7', 'go\_to\_object\_bin\_03\_3\_platform\_0', 'go\_to\_object\_wall\_cabinet\_01\_4\_platform\_0', 'go\_to\_object\_wall\_cabinet\_01\_4\_platform\_1', 'go\_to\_object\_wall\_cabinet\_01\_4\_platform\_2', 'go\_to\_object\_wall\_cabinet\_01\_4\_platform\_3', 'go\_to\_object\_wall\_cabinet\_01\_4\_platform\_4', 'go\_to\_object\_wall\_cabinet\_01\_4\_platform\_5', 'go\_to\_object\_wall\_cabinet\_01\_4\_platform\_6', 'go\_to\_object\_table\_03\_14\_platform\_1', 'go\_to\_object\_table\_01\_48\_platform\_0', 'go\_to\_object\_wall\_cabinet\_02\_21\_platform\_0', 'go\_to\_object\_wall\_cabinet\_02\_21\_platform\_1', 'go\_to\_object\_wall\_cabinet\_02\_21\_platform\_2', 'go\_to\_object\_wall\_cabinet\_02\_21\_platform\_3', 'go\_to\_object\_wall\_cabinet\_02\_21\_platform\_4', 'go\_to\_object\_wall\_cabinet\_02\_21\_platform\_5', 'go\_to\_object\_wall\_cabinet\_02\_21\_platform\_6', 'go\_to\_object\_chestOfDrawers\_01\_2\_body\_platform\_0', 'go\_to\_object\_chestOfDrawers\_01\_2\_body\_platform\_1', 'go\_to\_object\_chestOfDrawers\_01\_2\_body\_platform\_2', 'go\_to\_object\_chestOfDrawers\_01\_2\_body\_platform\_3', 'go\_to\_object\_chestOfDrawers\_01\_2\_body\_platform\_4', 'go\_to\_object\_chestOfDrawers\_01\_2\_body\_platform\_5', 'go\_to\_object\_chestOfDrawers\_01\_2\_body\_platform\_6', 'go\_to\_object\_table\_02\_40\_platform\_0', 'go\_to\_object\_table\_02\_40\_platform\_1', 'go\_to\_object\_table\_02\_40\_platform\_2', 'go\_to\_object\_tvstand\_88\_platform\_0', 'go\_to\_object\_cabinet\_3\_body\_platform\_0', 'go\_to\_object\_cabinet\_3\_body\_platform\_1']
- ['show\_receptacle\_of\_object\_1\_of\_current\_platform', 'show\_receptacle\_of\_object\_2\_of\_current\_platform', 'show\_receptacle\_of\_object\_3\_of\_current\_platform', 'show\_receptacle\_of\_object\_4\_of\_current\_platform', 'show\_receptacle\_of\_object\_5\_of\_current\_platform', 'show\_receptacle\_of\_object\_6\_of\_current\_platform', 'show\_receptacle\_of\_object\_7\_of\_current\_platform', 'show\_receptacle\_of\_object\_8\_of\_current\_platform', 'show\_receptacle\_of\_object\_9\_of\_current\_platform', 'show\_receptacle\_of\_object\_10\_of\_current\_platform', 'show\_receptacle\_of\_object\_11\_of\_current\_platform', 'show\_receptacle\_of\_object\_12\_of\_current\_platform', 'show\_receptacle\_of\_object\_13\_of\_current\_platform', 'show\_receptacle\_of\_object\_14\_of\_current\_platform', 'show\_receptacle\_of\_object\_15\_of\_current\_platform', 'show\_receptacle\_of\_object\_16\_of\_current\_platform', 'show\_receptacle\_of\_object\_17\_of\_current\_platform', 'show\_receptacle\_of\_object\_18\_of\_current\_platform', 'show\_receptacle\_of\_object\_19\_of\_current\_platform', 'show\_receptacle\_of\_object\_20\_of\_current\_platform', 'show\_receptacle\_of\_object\_21\_of\_current\_platform', 'show\_receptacle\_of\_object\_22\_of\_current\_platform', 'show\_receptacle\_of\_object\_23\_of\_current\_platform', 'show\_receptacle\_of\_object\_24\_of\_current\_platform', 'show\_receptacle\_of\_object\_25\_of\_current\_platform']
- `CALL\_END`: Call end (terminate the task).
- `change\_view`: Change the viewing angle to observe the scene from another perspective.
- `place\_r`:Place the object in an arbitrary place on the platform. - `place\_s\_[({object\_idx1},{dir\_idx1})...]`: Place the object in the union area of regions with indices {(object\_idx1,region\_idx1)}(object\_idx1's region\_idx1 space)  etc. All available pairs are in [(1, 1), (1, 2), (1, 3), (1, 4), (2, 1), (5, 1), (6, 1), (8, 1), (10, 1), (11, 1), (11, 2), (11, 3), (12, 1), (12, 2), (12, 3), (12, 4), (12, 5), (12, 6), (12, 7), (12, 8), (13, 1), (14, 1), (14, 2), (14, 3), (14, 4), (14, 5), (16, 1), (16, 2), (16, 3), (17, 1), (17, 2), (17, 3), (17, 4), (17, 5), (17, 6), (18, 1), (19, 1), (20, 1), (20, 2), (20, 3), (20, 4), (20, 5), (21, 1), (21, 2), (21, 3), (21, 4), (21, 5), (22, 1), (22, 2), (22, 3), (23, 1), (24, 1), (24, 2), (24, 3), (24, 4), (24, 5), (24, 6), (24, 7), (24, 8), (25, 1), (25, 2)], and a valid choice should be a subset of it.

All your current available actions are listed as above (all are without quotes and square brackets). Please don't mix up it with the action space of previous steps.

Understanding directions:
Directions like 'front', 'rear', 'left', and 'right' are defined relative to the robot's perspective and the heading of the object or platform you are interacting with. The heading of a ground object is not fixed and depends on where the robot is standing and facing.

Empty Surfaces (Platforms): When interacting with an empty platform, imagine it divided into a 3x3 grid. The direction aligned with the heading (also, the FARTHEST middle space to you, NOT THE NEAREST MIDDLE SPACE) is labeled "front," with the remaining regions proceeding counterclockwise as "front-left," "left," "rear-left," "rear," "rear-right," "right," and "front-right."  The grid is conceptually numbered as follows, based on these directions:

    [front-left, front, front-right],
    [left, center, right],
    [rear-left, rear, rear-right]

Occupied Surfaces (Objects on Platforms): For objects on platforms, imagine extending the edges of the object's bounding box to the platform's boundaries. The resulting regions define the object's 8 receptacles, and the direction is defined similarly.
Here are some useful suggestions for you to better finish the task: 

1. Your extra actions do not affect the success of the task, you only need to keep your steps within the total step limit.

2. When you want to place an object but cannot find enough space, you can try changing your standing position to find the correct orientation, try combining multiple receptacles, or remove objects that are blocking your way first(although you generally do not need to do this). Specifically, for the task asking you put object to empty platforms, try combining adjacent receptacles may be very useful.

3. For tasks that involve placing objects near other objects, please remember to get a focused image of the target object and examine it carefully. This is important because the receptacles may not be intuitive. 

4. Important: Don't output anything other than the action you want to take. You will only receive the same hint informing you your invalid action. If you make mistakes, just start output correct actions from the next prompt without any apologize.

Current task:

Move kitchenware\_white\_paper\_towel\_holder between lighting\_blue\_base\_table\_lamp and lighting\_blue\_base\_table\_lamp

Steps used: 3/20. You can only take at most 20 steps, so hurry if you've almost used all of them!

[*See Fig.~\ref{fig:success_step3}]

\textcolor{purple}{Here are 4 images, named ['HoldingOccupiedPlatform3\_1\_out\_of\_4.png', 'HoldingOccupiedPlatform3\_2\_out\_of\_4.png', 'HoldingOccupiedPlatform3\_3\_out\_of\_4.png', 'HoldingOccupiedPlatform3\_4\_out\_of\_4.png'] respectively.}

\textcolor{purple}{Their description is as follows: These image show the platform you are currently on in one view, with regions or objects marked with numbers for you to place objects accurately. The platform is object\_kitchen\_counter\_1\_body\_platform\_1.}

\textcolor{purple}{If the platform has no objects, a 3x3 grid will be marked on the platform to help you place objects, with each grid marked in numbers from 1 to 9. If there are objects on the platform, guiding lines will indicate the object numbers. Note that  if you see multiple images, it means the platform is too long to be captured in one image, and it's guaranteed that the same object will marked in THE SAME COLOR GUIDELINES AND THE SAME NUMBER. If you think you can't determine the number of an item in one picture, you can go through other pictures and see if this item is marked more clearly.}

\hdashrule{\linewidth}{0.4pt}{2mm}

\textcolor{blue}
{VLM Response: place\_r}

\hdashrule{\linewidth}{0.4pt}{2mm}

User Message:

\textbf{
The object has been placed successfully. You can now call end if you think you've finished the task correctly, or can also pick up the object again if you think the placement is not correct.}

You are currently at object\_kitchen\_counter\_1\_body\_platform\_1.

And you are holding nothing.

Your available action space:

- ['go\_to\_object\_sofa\_10\_platform\_0', 'go\_to\_object\_sofa\_10\_platform\_1', 'go\_to\_object\_sofa\_10\_platform\_2', 'go\_to\_object\_table\_04\_13\_platform\_0', 'go\_to\_object\_chair\_04\_46\_platform\_0', 'go\_to\_object\_chair\_04\_47\_platform\_0', 'go\_to\_object\_chair\_05\_7\_platform\_0', 'go\_to\_object\_chair\_05\_8\_platform\_0', 'go\_to\_object\_stool\_02\_18\_platform\_0', 'go\_to\_object\_stool\_02\_6\_platform\_0', 'go\_to\_object\_rack\_01\_75\_platform\_0', 'go\_to\_object\_rack\_01\_75\_platform\_1', 'go\_to\_object\_rack\_01\_75\_platform\_2', 'go\_to\_object\_kitchen\_counter\_1\_body\_platform\_0', 'go\_to\_object\_kitchen\_counter\_1\_body\_platform\_1', 'go\_to\_object\_fridge\_0\_body\_platform\_1', 'go\_to\_object\_fridge\_0\_body\_platform\_3', 'go\_to\_object\_fridge\_0\_body\_platform\_4', 'go\_to\_object\_fridge\_0\_body\_platform\_5', 'go\_to\_object\_fridge\_0\_body\_platform\_6', 'go\_to\_object\_fridge\_0\_body\_platform\_7', 'go\_to\_object\_bin\_03\_3\_platform\_0', 'go\_to\_object\_wall\_cabinet\_01\_4\_platform\_0', 'go\_to\_object\_wall\_cabinet\_01\_4\_platform\_1', 'go\_to\_object\_wall\_cabinet\_01\_4\_platform\_2', 'go\_to\_object\_wall\_cabinet\_01\_4\_platform\_3', 'go\_to\_object\_wall\_cabinet\_01\_4\_platform\_4', 'go\_to\_object\_wall\_cabinet\_01\_4\_platform\_5', 'go\_to\_object\_wall\_cabinet\_01\_4\_platform\_6', 'go\_to\_object\_table\_03\_14\_platform\_1', 'go\_to\_object\_table\_01\_48\_platform\_0', 'go\_to\_object\_wall\_cabinet\_02\_21\_platform\_0', 'go\_to\_object\_wall\_cabinet\_02\_21\_platform\_1', 'go\_to\_object\_wall\_cabinet\_02\_21\_platform\_2', 'go\_to\_object\_wall\_cabinet\_02\_21\_platform\_3', 'go\_to\_object\_wall\_cabinet\_02\_21\_platform\_4', 'go\_to\_object\_wall\_cabinet\_02\_21\_platform\_5', 'go\_to\_object\_wall\_cabinet\_02\_21\_platform\_6', 'go\_to\_object\_chestOfDrawers\_01\_2\_body\_platform\_0', 'go\_to\_object\_chestOfDrawers\_01\_2\_body\_platform\_1', 'go\_to\_object\_chestOfDrawers\_01\_2\_body\_platform\_2', 'go\_to\_object\_chestOfDrawers\_01\_2\_body\_platform\_3', 'go\_to\_object\_chestOfDrawers\_01\_2\_body\_platform\_4', 'go\_to\_object\_chestOfDrawers\_01\_2\_body\_platform\_5', 'go\_to\_object\_chestOfDrawers\_01\_2\_body\_platform\_6', 'go\_to\_object\_table\_02\_40\_platform\_0', 'go\_to\_object\_table\_02\_40\_platform\_1', 'go\_to\_object\_table\_02\_40\_platform\_2', 'go\_to\_object\_tvstand\_88\_platform\_0', 'go\_to\_object\_cabinet\_3\_body\_platform\_0', 'go\_to\_object\_cabinet\_3\_body\_platform\_1']
- ['pick\_object\_1\_of\_current\_platform', 'pick\_object\_2\_of\_current\_platform', 'pick\_object\_3\_of\_current\_platform', 'pick\_object\_4\_of\_current\_platform', 'pick\_object\_5\_of\_current\_platform', 'pick\_object\_6\_of\_current\_platform', 'pick\_object\_7\_of\_current\_platform', 'pick\_object\_8\_of\_current\_platform', 'pick\_object\_9\_of\_current\_platform', 'pick\_object\_10\_of\_current\_platform', 'pick\_object\_11\_of\_current\_platform', 'pick\_object\_12\_of\_current\_platform', 'pick\_object\_13\_of\_current\_platform', 'pick\_object\_14\_of\_current\_platform', 'pick\_object\_15\_of\_current\_platform', 'pick\_object\_16\_of\_current\_platform', 'pick\_object\_17\_of\_current\_platform', 'pick\_object\_18\_of\_current\_platform', 'pick\_object\_19\_of\_current\_platform', 'pick\_object\_20\_of\_current\_platform', 'pick\_object\_21\_of\_current\_platform', 'pick\_object\_22\_of\_current\_platform', 'pick\_object\_23\_of\_current\_platform', 'pick\_object\_24\_of\_current\_platform', 'pick\_object\_25\_of\_current\_platform', 'pick\_object\_26\_of\_current\_platform']
- ['show\_receptacle\_of\_object\_1\_of\_current\_platform', 'show\_receptacle\_of\_object\_2\_of\_current\_platform', 'show\_receptacle\_of\_object\_3\_of\_current\_platform', 'show\_receptacle\_of\_object\_4\_of\_current\_platform', 'show\_receptacle\_of\_object\_5\_of\_current\_platform', 'show\_receptacle\_of\_object\_6\_of\_current\_platform', 'show\_receptacle\_of\_object\_7\_of\_current\_platform', 'show\_receptacle\_of\_object\_8\_of\_current\_platform', 'show\_receptacle\_of\_object\_9\_of\_current\_platform', 'show\_receptacle\_of\_object\_10\_of\_current\_platform', 'show\_receptacle\_of\_object\_11\_of\_current\_platform', 'show\_receptacle\_of\_object\_12\_of\_current\_platform', 'show\_receptacle\_of\_object\_13\_of\_current\_platform', 'show\_receptacle\_of\_object\_14\_of\_current\_platform', 'show\_receptacle\_of\_object\_15\_of\_current\_platform', 'show\_receptacle\_of\_object\_16\_of\_current\_platform', 'show\_receptacle\_of\_object\_17\_of\_current\_platform', 'show\_receptacle\_of\_object\_18\_of\_current\_platform', 'show\_receptacle\_of\_object\_19\_of\_current\_platform', 'show\_receptacle\_of\_object\_20\_of\_current\_platform', 'show\_receptacle\_of\_object\_21\_of\_current\_platform', 'show\_receptacle\_of\_object\_22\_of\_current\_platform', 'show\_receptacle\_of\_object\_23\_of\_current\_platform', 'show\_receptacle\_of\_object\_24\_of\_current\_platform', 'show\_receptacle\_of\_object\_25\_of\_current\_platform', 'show\_receptacle\_of\_object\_26\_of\_current\_platform']
- `CALL\_END`: Call end (terminate the task).
- `change\_view`: Change the viewing angle to observe the scene from another perspective.

Choose one of the available actions.

All your current available actions are listed as above (all are without quotes and square brackets). Please don't mix up it with the action space of previous steps.

Here are some useful suggestions for you to better finish the task: 

1. Your extra actions do not affect the success of the task, you only need to keep your steps within the total step limit.

2. When you want to place an object but cannot find enough space, you can try changing your standing position to find the correct orientation, try combining multiple receptacles, or remove objects that are blocking your way first(although you generally do not need to do this). Specifically, for the task asking you put object to empty platforms, try combining adjacent receptacles may be very useful.

3. For tasks that involve placing objects near other objects, please remember to get a focused image of the target object and examine it carefully. This is important because the receptacles may not be intuitive. 

4. Important: Don't output anything other than the action you want to take. You will only receive the same hint informing you your invalid action. If you make mistakes, just start output correct actions from the next prompt without any apologize.

Current task:

Move kitchenware\_white\_paper\_towel\_holder between lighting\_blue\_base\_table\_lamp and lighting\_blue\_base\_table\_lamp

Steps used: 4/20. You can only take at most 20 steps, so hurry if you've almost used all of them!

[*See Fig.~\ref{fig:success_step4}]

\textcolor{purple}
{Here are 4 images, named ['Idle\_4\_1\_out\_of\_4.png', 'Idle\_4\_2\_out\_of\_4.png', 'Idle\_4\_3\_out\_of\_4.png', 'Idle\_4\_4\_out\_of\_4.png'] respectively.}

\textcolor{purple}
{Their description is as follows: These images show the platform you are currently on in one view, with regions or objects marked with numbers for you to place objects accurately. The platform is object\_kitchen\_counter\_1\_body\_platform\_1.}

\textcolor{purple}
{If the platform has no objects, a 3x3 grid will be marked on the platform to help you place objects, with each grid marked in numbers from 1 to 9. If there are objects on the platform, guiding lines will indicate the object numbers. Note that  if you see multiple images, it means the platform is too long to be captured in one image, and it's guaranteed that the same object will marked in THE SAME COLOR GUIDELINES AND THE SAME NUMBER. If you think you can't determine the number of an item in one picture, you can go through other pictures and see if this item is marked more clearly.}

\hdashrule{\linewidth}{0.4pt}{2mm}

\textcolor{blue}
{VLM\_response: pick\_object\_17\_of\_current\_platform}

\hdashrule{\linewidth}{0.4pt}{2mm}

User Message:

You are currently at object\_kitchen\_counter\_1\_body\_platform\_1.

And you are holding kitchenware\_small\_white\_bowl.

Your available action space:
-['go\_to\_object\_sofa\_10\_platform\_0', 'go\_to\_object\_sofa\_10\_platform\_1', 'go\_to\_object\_sofa\_10\_platform\_2', 'go\_to\_object\_table\_04\_13\_platform\_0', 'go\_to\_object\_chair\_04\_46\_platform\_0', 'go\_to\_object\_chair\_04\_47\_platform\_0', 'go\_to\_object\_chair\_05\_7\_platform\_0', 'go\_to\_object\_chair\_05\_8\_platform\_0', 'go\_to\_object\_stool\_02\_18\_platform\_0', 'go\_to\_object\_stool\_02\_6\_platform\_0', 'go\_to\_object\_rack\_01\_75\_platform\_0', 'go\_to\_object\_rack\_01\_75\_platform\_1', 'go\_to\_object\_rack\_01\_75\_platform\_2', 'go\_to\_object\_kitchen\_counter\_1\_body\_platform\_0', 'go\_to\_object\_kitchen\_counter\_1\_body\_platform\_1', 'go\_to\_object\_fridge\_0\_body\_platform\_1', 'go\_to\_object\_fridge\_0\_body\_platform\_3', 'go\_to\_object\_fridge\_0\_body\_platform\_4', 'go\_to\_object\_fridge\_0\_body\_platform\_5', 'go\_to\_object\_fridge\_0\_body\_platform\_6', 'go\_to\_object\_fridge\_0\_body\_platform\_7', 'go\_to\_object\_bin\_03\_3\_platform\_0', 'go\_to\_object\_wall\_cabinet\_01\_4\_platform\_0', 'go\_to\_object\_wall\_cabinet\_01\_4\_platform\_1', 'go\_to\_object\_wall\_cabinet\_01\_4\_platform\_2', 'go\_to\_object\_wall\_cabinet\_01\_4\_platform\_3', 'go\_to\_object\_wall\_cabinet\_01\_4\_platform\_4', 'go\_to\_object\_wall\_cabinet\_01\_4\_platform\_5', 'go\_to\_object\_wall\_cabinet\_01\_4\_platform\_6', 'go\_to\_object\_table\_03\_14\_platform\_1', 'go\_to\_object\_table\_01\_48\_platform\_0', 'go\_to\_object\_wall\_cabinet\_02\_21\_platform\_0', 'go\_to\_object\_wall\_cabinet\_02\_21\_platform\_1', 'go\_to\_object\_wall\_cabinet\_02\_21\_platform\_2', 'go\_to\_object\_wall\_cabinet\_02\_21\_platform\_3', 'go\_to\_object\_wall\_cabinet\_02\_21\_platform\_4', 'go\_to\_object\_wall\_cabinet\_02\_21\_platform\_5', 'go\_to\_object\_wall\_cabinet\_02\_21\_platform\_6', 'go\_to\_object\_chestOfDrawers\_01\_2\_body\_platform\_0', 'go\_to\_object\_chestOfDrawers\_01\_2\_body\_platform\_1', 'go\_to\_object\_chestOfDrawers\_01\_2\_body\_platform\_2', 'go\_to\_object\_chestOfDrawers\_01\_2\_body\_platform\_3', 'go\_to\_object\_chestOfDrawers\_01\_2\_body\_platform\_4', 'go\_to\_object\_chestOfDrawers\_01\_2\_body\_platform\_5', 'go\_to\_object\_chestOfDrawers\_01\_2\_body\_platform\_6', 'go\_to\_object\_table\_02\_40\_platform\_0', 'go\_to\_object\_table\_02\_40\_platform\_1', 'go\_to\_object\_table\_02\_40\_platform\_2', 'go\_to\_object\_tvstand\_88\_platform\_0', 'go\_to\_object\_cabinet\_3\_body\_platform\_0', 'go\_to\_object\_cabinet\_3\_body\_platform\_1']
- ['show\_receptacle\_of\_object\_1\_of\_current\_platform', 'show\_receptacle\_of\_object\_2\_of\_current\_platform', 'show\_receptacle\_of\_object\_3\_of\_current\_platform', 'show\_receptacle\_of\_object\_4\_of\_current\_platform', 'show\_receptacle\_of\_object\_5\_of\_current\_platform', 'show\_receptacle\_of\_object\_6\_of\_current\_platform', 'show\_receptacle\_of\_object\_7\_of\_current\_platform', 'show\_receptacle\_of\_object\_8\_of\_current\_platform', 'show\_receptacle\_of\_object\_9\_of\_current\_platform', 'show\_receptacle\_of\_object\_10\_of\_current\_platform', 'show\_receptacle\_of\_object\_11\_of\_current\_platform', 'show\_receptacle\_of\_object\_12\_of\_current\_platform', 'show\_receptacle\_of\_object\_13\_of\_current\_platform', 'show\_receptacle\_of\_object\_14\_of\_current\_platform', 'show\_receptacle\_of\_object\_15\_of\_current\_platform', 'show\_receptacle\_of\_object\_16\_of\_current\_platform', 'show\_receptacle\_of\_object\_17\_of\_current\_platform', 'show\_receptacle\_of\_object\_18\_of\_current\_platform', 'show\_receptacle\_of\_object\_19\_of\_current\_platform', 'show\_receptacle\_of\_object\_20\_of\_current\_platform', 'show\_receptacle\_of\_object\_21\_of\_current\_platform', 'show\_receptacle\_of\_object\_22\_of\_current\_platform', 'show\_receptacle\_of\_object\_23\_of\_current\_platform', 'show\_receptacle\_of\_object\_24\_of\_current\_platform', 'show\_receptacle\_of\_object\_25\_of\_current\_platform']
- `CALL\_END`: Call end (terminate the task).
- `change\_view`: Change the viewing angle to observe the scene from another perspective.
- `place\_r`:Place the object in an arbitrary place on the platform. - `place\_s\_[({object\_idx1},{dir\_idx1})...]`: Place the object in the union area of regions with indices {(object\_idx1,region\_idx1)}(object\_idx1's region\_idx1 space)  etc. All available pairs are in [(1, 1), (1, 2), (1, 3), (1, 4), (2, 1), (5, 1), (6, 1), (8, 1), (10, 1), (11, 1), (11, 2), (11, 3), (12, 1), (12, 2), (12, 3), (12, 4), (12, 5), (12, 6), (12, 7), (12, 8), (13, 1), (13, 2), (14, 1), (14, 2), (14, 3), (14, 4), (14, 5), (16, 1), (16, 2), (16, 3), (17, 1), (18, 1), (19, 1), (19, 2), (19, 3), (19, 4), (19, 5), (20, 1), (20, 2), (20, 3), (20, 4), (20, 5), (21, 1), (21, 2), (21, 3), (22, 1), (23, 1), (23, 2), (23, 3), (23, 4), (23, 5), (23, 6), (23, 7), (23, 8), (24, 1), (24, 2), (25, 1), (25, 2), (25, 3)], and a valid choice should be a subset of it. 

All your current available actions are listed as above (all are without quotes and square brackets). Please don't mix up it with the action space of previous steps.

Understanding directions:
Directions like 'front', 'rear', 'left', and 'right' are defined relative to the robot's perspective and the heading of the object or platform you are interacting with. The heading of a ground object is not fixed and depends on where the robot is standing and facing.

Empty Surfaces (Platforms): When interacting with an empty platform, imagine it divided into a 3x3 grid. The direction aligned with the heading (also, the FARTHEST middle space to you, NOT THE NEAREST MIDDLE SPACE) is labeled "front," with the remaining regions proceeding counterclockwise as "front-left," "left," "rear-left," "rear," "rear-right," "right," and "front-right."  The grid is conceptually numbered as follows, based on these directions:

    [front-left, front, front-right],
    [left, center, right],
    [rear-left, rear, rear-right]

Occupied Surfaces (Objects on Platforms): For objects on platforms, imagine extending the edges of the object's bounding box to the platform's boundaries. The resulting regions define the object's 8 receptacles, and the direction is defined similarly.
Here are some useful suggestions for you to better finish the task: 

1. Your extra actions do not affect the success of the task, you only need to keep your steps within the total step limit.

2. When you want to place an object but cannot find enough space, you can try changing your standing position to find the correct orientation, try combining multiple receptacles, or remove objects that are blocking your way first(although you generally do not need to do this). Specifically, for the task asking you put object to empty platforms, try combining adjacent receptacles may be very useful.

3. For tasks that involve placing objects near other objects, please remember to get a focused image of the target object and examine it carefully. This is important because the receptacles may not be intuitive. 

4. Important: Don't output anything other than the action you want to take. You will only receive the same hint informing you your invalid action. If you make mistakes, just start output correct actions from the next prompt without any apologize.

Current task:

Move kitchenware\_white\_paper\_towel\_holder between lighting\_blue\_base\_table\_lamp and lighting\_blue\_base\_table\_lamp

Steps used: 5/20. You can only take at most 20 steps, so hurry if you've almost used all of them!

[*See Fig.~\ref{fig:success_step5}]

\textcolor{purple}{Here are 4 images, named ['HoldingOccupiedPlatform5\_1\_out\_of\_4.png', 'HoldingOccupiedPlatform5\_2\_out\_of\_4.png', 'HoldingOccupiedPlatform5\_3\_out\_of\_4.png', 'HoldingOccupiedPlatform5\_4\_out\_of\_4.png'] respectively.}

\textcolor{purple}{Their description is as follows: These image show the platform you are currently on in one view, with regions or objects marked with numbers for you to place objects accurately. The platform is object\_kitchen\_counter\_1\_body\_platform\_1.}

\textcolor{purple}{If the platform has no objects, a 3x3 grid will be marked on the platform to help you place objects, with each grid marked in numbers from 1 to 9. If there are objects on the platform, guiding lines will indicate the object numbers. Note that  if you see multiple images, it means the platform is too long to be captured in one image, and it's guaranteed that the same object will marked in THE SAME COLOR GUIDELINES AND THE SAME NUMBER. If you think you can't determine the number of an item in one picture, you can go through other pictures and see if this item is marked more clearly.}

\hdashrule{\linewidth}{0.4pt}{2mm}

\textcolor{blue}
{VLM Response: place\_r}

\hdashrule{\linewidth}{0.4pt}{2mm}

User Message:

\textbf{
The object has been placed successfully. You can now call end if you think you've finished the task correctly, or can also pick up the object again if you think the placement is not correct.}

You are currently at object\_kitchen\_counter\_1\_body\_platform\_1.

And you are holding nothing.

Your available action space:
- ['go\_to\_object\_sofa\_10\_platform\_0', 'go\_to\_object\_sofa\_10\_platform\_1', 'go\_to\_object\_sofa\_10\_platform\_2', 'go\_to\_object\_table\_04\_13\_platform\_0', 'go\_to\_object\_chair\_04\_46\_platform\_0', 'go\_to\_object\_chair\_04\_47\_platform\_0', 'go\_to\_object\_chair\_05\_7\_platform\_0', 'go\_to\_object\_chair\_05\_8\_platform\_0', 'go\_to\_object\_stool\_02\_18\_platform\_0', 'go\_to\_object\_stool\_02\_6\_platform\_0', 'go\_to\_object\_rack\_01\_75\_platform\_0', 'go\_to\_object\_rack\_01\_75\_platform\_1', 'go\_to\_object\_rack\_01\_75\_platform\_2', 'go\_to\_object\_kitchen\_counter\_1\_body\_platform\_0', 'go\_to\_object\_kitchen\_counter\_1\_body\_platform\_1', 'go\_to\_object\_fridge\_0\_body\_platform\_1', 'go\_to\_object\_fridge\_0\_body\_platform\_3', 'go\_to\_object\_fridge\_0\_body\_platform\_4', 'go\_to\_object\_fridge\_0\_body\_platform\_5', 'go\_to\_object\_fridge\_0\_body\_platform\_6', 'go\_to\_object\_fridge\_0\_body\_platform\_7', 'go\_to\_object\_bin\_03\_3\_platform\_0', 'go\_to\_object\_wall\_cabinet\_01\_4\_platform\_0', 'go\_to\_object\_wall\_cabinet\_01\_4\_platform\_1', 'go\_to\_object\_wall\_cabinet\_01\_4\_platform\_2', 'go\_to\_object\_wall\_cabinet\_01\_4\_platform\_3', 'go\_to\_object\_wall\_cabinet\_01\_4\_platform\_4', 'go\_to\_object\_wall\_cabinet\_01\_4\_platform\_5', 'go\_to\_object\_wall\_cabinet\_01\_4\_platform\_6', 'go\_to\_object\_table\_03\_14\_platform\_1', 'go\_to\_object\_table\_01\_48\_platform\_0', 'go\_to\_object\_wall\_cabinet\_02\_21\_platform\_0', 'go\_to\_object\_wall\_cabinet\_02\_21\_platform\_1', 'go\_to\_object\_wall\_cabinet\_02\_21\_platform\_2', 'go\_to\_object\_wall\_cabinet\_02\_21\_platform\_3', 'go\_to\_object\_wall\_cabinet\_02\_21\_platform\_4', 'go\_to\_object\_wall\_cabinet\_02\_21\_platform\_5', 'go\_to\_object\_wall\_cabinet\_02\_21\_platform\_6', 'go\_to\_object\_chestOfDrawers\_01\_2\_body\_platform\_0', 'go\_to\_object\_chestOfDrawers\_01\_2\_body\_platform\_1', 'go\_to\_object\_chestOfDrawers\_01\_2\_body\_platform\_2', 'go\_to\_object\_chestOfDrawers\_01\_2\_body\_platform\_3', 'go\_to\_object\_chestOfDrawers\_01\_2\_body\_platform\_4', 'go\_to\_object\_chestOfDrawers\_01\_2\_body\_platform\_5', 'go\_to\_object\_chestOfDrawers\_01\_2\_body\_platform\_6', 'go\_to\_object\_table\_02\_40\_platform\_0', 'go\_to\_object\_table\_02\_40\_platform\_1', 'go\_to\_object\_table\_02\_40\_platform\_2', 'go\_to\_object\_tvstand\_88\_platform\_0', 'go\_to\_object\_cabinet\_3\_body\_platform\_0', 'go\_to\_object\_cabinet\_3\_body\_platform\_1']
- ['pick\_object\_1\_of\_current\_platform', 'pick\_object\_2\_of\_current\_platform', 'pick\_object\_3\_of\_current\_platform', 'pick\_object\_4\_of\_current\_platform', 'pick\_object\_5\_of\_current\_platform', 'pick\_object\_6\_of\_current\_platform', 'pick\_object\_7\_of\_current\_platform', 'pick\_object\_8\_of\_current\_platform', 'pick\_object\_9\_of\_current\_platform', 'pick\_object\_10\_of\_current\_platform', 'pick\_object\_11\_of\_current\_platform', 'pick\_object\_12\_of\_current\_platform', 'pick\_object\_13\_of\_current\_platform', 'pick\_object\_14\_of\_current\_platform', 'pick\_object\_15\_of\_current\_platform', 'pick\_object\_16\_of\_current\_platform', 'pick\_object\_17\_of\_current\_platform', 'pick\_object\_18\_of\_current\_platform', 'pick\_object\_19\_of\_current\_platform', 'pick\_object\_20\_of\_current\_platform', 'pick\_object\_21\_of\_current\_platform', 'pick\_object\_22\_of\_current\_platform', 'pick\_object\_23\_of\_current\_platform', 'pick\_object\_24\_of\_current\_platform', 'pick\_object\_25\_of\_current\_platform', 'pick\_object\_26\_of\_current\_platform']
- ['show\_receptacle\_of\_object\_1\_of\_current\_platform', 'show\_receptacle\_of\_object\_2\_of\_current\_platform', 'show\_receptacle\_of\_object\_3\_of\_current\_platform', 'show\_receptacle\_of\_object\_4\_of\_current\_platform', 'show\_receptacle\_of\_object\_5\_of\_current\_platform', 'show\_receptacle\_of\_object\_6\_of\_current\_platform', 'show\_receptacle\_of\_object\_7\_of\_current\_platform', 'show\_receptacle\_of\_object\_8\_of\_current\_platform', 'show\_receptacle\_of\_object\_9\_of\_current\_platform', 'show\_receptacle\_of\_object\_10\_of\_current\_platform', 'show\_receptacle\_of\_object\_11\_of\_current\_platform', 'show\_receptacle\_of\_object\_12\_of\_current\_platform', 'show\_receptacle\_of\_object\_13\_of\_current\_platform', 'show\_receptacle\_of\_object\_14\_of\_current\_platform', 'show\_receptacle\_of\_object\_15\_of\_current\_platform', 'show\_receptacle\_of\_object\_16\_of\_current\_platform', 'show\_receptacle\_of\_object\_17\_of\_current\_platform', 'show\_receptacle\_of\_object\_18\_of\_current\_platform', 'show\_receptacle\_of\_object\_19\_of\_current\_platform', 'show\_receptacle\_of\_object\_20\_of\_current\_platform', 'show\_receptacle\_of\_object\_21\_of\_current\_platform', 'show\_receptacle\_of\_object\_22\_of\_current\_platform', 'show\_receptacle\_of\_object\_23\_of\_current\_platform', 'show\_receptacle\_of\_object\_24\_of\_current\_platform', 'show\_receptacle\_of\_object\_25\_of\_current\_platform', 'show\_receptacle\_of\_object\_26\_of\_current\_platform']
- `CALL\_END`: Call end (terminate the task).
- `change\_view`: Change the viewing angle to observe the scene from another perspective.

Choose one of the available actions.

All your current available actions are listed as above (all are without quotes and square brackets). Please don't mix up it with the action space of previous steps.

Here are some useful suggestions for you to better finish the task: 

1. Your extra actions do not affect the success of the task, you only need to keep your steps within the total step limit.

2. When you want to place an object but cannot find enough space, you can try changing your standing position to find the correct orientation, try combining multiple receptacles, or remove objects that are blocking your way first(although you generally do not need to do this). Specifically, for the task asking you put object to empty platforms, try combining adjacent receptacles may be very useful.

3. For tasks that involve placing objects near other objects, please remember to get a focused image of the target object and examine it carefully. This is important because the receptacles may not be intuitive. 

4. Important: Don't output anything other than the action you want to take. You will only receive the same hint informing you your invalid action. If you make mistakes, just start output correct actions from the next prompt without any apologize.

Current task:

Move kitchenware\_white\_paper\_towel\_holder between lighting\_blue\_base\_table\_lamp and lighting\_blue\_base\_table\_lamp

Steps used: 6/20. You can only take at most 20 steps, so hurry if you've almost used all of them!

[*See Fig.~\ref{fig:success_step6}]

\textcolor{purple}
{Here are 4 images, named ['Idle\_6\_1\_out\_of\_4.png', 'Idle\_6\_2\_out\_of\_4.png', 'Idle\_6\_3\_out\_of\_4.png', 'Idle\_6\_4\_out\_of\_4.png'] respectively.}

\textcolor{purple}
{Their description is as follows: These images show the platform you are currently on in one view, with regions or objects marked with numbers for you to place objects accurately. The platform is object\_kitchen\_counter\_1\_body\_platform\_1.}

\textcolor{purple}
{If the platform has no objects, a 3x3 grid will be marked on the platform to help you place objects, with each grid marked in numbers from 1 to 9. If there are objects on the platform, guiding lines will indicate the object numbers. Note that  if you see multiple images, it means the platform is too long to be captured in one image, and it's guaranteed that the same object will marked in THE SAME COLOR GUIDELINES AND THE SAME NUMBER. If you think you can't determine the number of an item in one picture, you can go through other pictures and see if this item is marked more clearly.}

\hdashrule{\linewidth}{0.4pt}{2mm}

\textcolor{blue}
{VLM\_response: pick\_object\_7\_of\_current\_platform}

\hdashrule{\linewidth}{0.4pt}{2mm}

User Message:

You are currently at object\_kitchen\_counter\_1\_body\_platform\_1.

And you are holding kitchenware\_spice\_container.

Your available action space:
-['go\_to\_object\_sofa\_10\_platform\_0', 'go\_to\_object\_sofa\_10\_platform\_1', 'go\_to\_object\_sofa\_10\_platform\_2', 'go\_to\_object\_table\_04\_13\_platform\_0', 'go\_to\_object\_chair\_04\_46\_platform\_0', 'go\_to\_object\_chair\_04\_47\_platform\_0', 'go\_to\_object\_chair\_05\_7\_platform\_0', 'go\_to\_object\_chair\_05\_8\_platform\_0', 'go\_to\_object\_stool\_02\_18\_platform\_0', 'go\_to\_object\_stool\_02\_6\_platform\_0', 'go\_to\_object\_rack\_01\_75\_platform\_0', 'go\_to\_object\_rack\_01\_75\_platform\_1', 'go\_to\_object\_rack\_01\_75\_platform\_2', 'go\_to\_object\_kitchen\_counter\_1\_body\_platform\_0', 'go\_to\_object\_kitchen\_counter\_1\_body\_platform\_1', 'go\_to\_object\_fridge\_0\_body\_platform\_1', 'go\_to\_object\_fridge\_0\_body\_platform\_3', 'go\_to\_object\_fridge\_0\_body\_platform\_4', 'go\_to\_object\_fridge\_0\_body\_platform\_5', 'go\_to\_object\_fridge\_0\_body\_platform\_6', 'go\_to\_object\_fridge\_0\_body\_platform\_7', 'go\_to\_object\_bin\_03\_3\_platform\_0', 'go\_to\_object\_wall\_cabinet\_01\_4\_platform\_0', 'go\_to\_object\_wall\_cabinet\_01\_4\_platform\_1', 'go\_to\_object\_wall\_cabinet\_01\_4\_platform\_2', 'go\_to\_object\_wall\_cabinet\_01\_4\_platform\_3', 'go\_to\_object\_wall\_cabinet\_01\_4\_platform\_4', 'go\_to\_object\_wall\_cabinet\_01\_4\_platform\_5', 'go\_to\_object\_wall\_cabinet\_01\_4\_platform\_6', 'go\_to\_object\_table\_03\_14\_platform\_1', 'go\_to\_object\_table\_01\_48\_platform\_0', 'go\_to\_object\_wall\_cabinet\_02\_21\_platform\_0', 'go\_to\_object\_wall\_cabinet\_02\_21\_platform\_1', 'go\_to\_object\_wall\_cabinet\_02\_21\_platform\_2', 'go\_to\_object\_wall\_cabinet\_02\_21\_platform\_3', 'go\_to\_object\_wall\_cabinet\_02\_21\_platform\_4', 'go\_to\_object\_wall\_cabinet\_02\_21\_platform\_5', 'go\_to\_object\_wall\_cabinet\_02\_21\_platform\_6', 'go\_to\_object\_chestOfDrawers\_01\_2\_body\_platform\_0', 'go\_to\_object\_chestOfDrawers\_01\_2\_body\_platform\_1', 'go\_to\_object\_chestOfDrawers\_01\_2\_body\_platform\_2', 'go\_to\_object\_chestOfDrawers\_01\_2\_body\_platform\_3', 'go\_to\_object\_chestOfDrawers\_01\_2\_body\_platform\_4', 'go\_to\_object\_chestOfDrawers\_01\_2\_body\_platform\_5', 'go\_to\_object\_chestOfDrawers\_01\_2\_body\_platform\_6', 'go\_to\_object\_table\_02\_40\_platform\_0', 'go\_to\_object\_table\_02\_40\_platform\_1', 'go\_to\_object\_table\_02\_40\_platform\_2', 'go\_to\_object\_tvstand\_88\_platform\_0', 'go\_to\_object\_cabinet\_3\_body\_platform\_0', 'go\_to\_object\_cabinet\_3\_body\_platform\_1']
- ['show\_receptacle\_of\_object\_1\_of\_current\_platform', 'show\_receptacle\_of\_object\_2\_of\_current\_platform', 'show\_receptacle\_of\_object\_3\_of\_current\_platform', 'show\_receptacle\_of\_object\_4\_of\_current\_platform', 'show\_receptacle\_of\_object\_5\_of\_current\_platform', 'show\_receptacle\_of\_object\_6\_of\_current\_platform', 'show\_receptacle\_of\_object\_7\_of\_current\_platform', 'show\_receptacle\_of\_object\_8\_of\_current\_platform', 'show\_receptacle\_of\_object\_9\_of\_current\_platform', 'show\_receptacle\_of\_object\_10\_of\_current\_platform', 'show\_receptacle\_of\_object\_11\_of\_current\_platform', 'show\_receptacle\_of\_object\_12\_of\_current\_platform', 'show\_receptacle\_of\_object\_13\_of\_current\_platform', 'show\_receptacle\_of\_object\_14\_of\_current\_platform', 'show\_receptacle\_of\_object\_15\_of\_current\_platform', 'show\_receptacle\_of\_object\_16\_of\_current\_platform', 'show\_receptacle\_of\_object\_17\_of\_current\_platform', 'show\_receptacle\_of\_object\_18\_of\_current\_platform', 'show\_receptacle\_of\_object\_19\_of\_current\_platform', 'show\_receptacle\_of\_object\_20\_of\_current\_platform', 'show\_receptacle\_of\_object\_21\_of\_current\_platform', 'show\_receptacle\_of\_object\_22\_of\_current\_platform', 'show\_receptacle\_of\_object\_23\_of\_current\_platform', 'show\_receptacle\_of\_object\_24\_of\_current\_platform', 'show\_receptacle\_of\_object\_25\_of\_current\_platform']
- `CALL\_END`: Call end (terminate the task).
- `change\_view`: Change the viewing angle to observe the scene from another perspective.
- `place\_r`:Place the object in an arbitrary place on the platform. - `place\_s\_[({object\_idx1},{dir\_idx1})...]`: Place the object in the union area of regions with indices {(object\_idx1,region\_idx1)}(object\_idx1's region\_idx1 space)  etc. All available pairs are in  [(1, 1), (1, 2), (1, 3), (1, 4), (2, 1), (5, 1), (6, 1), (6, 2), (7, 1), (8, 1), (9, 1), (10, 1), (10, 2), (10, 3), (11, 1), (11, 2), (11, 3), (11, 4), (11, 5), (11, 6), (11, 7), (12, 1), (12, 2), (13, 1), (13, 2), (13, 3), (13, 4), (13, 5), (15, 1), (15, 2), (15, 3), (16, 1), (17, 1), (18, 1), (18, 2), (18, 3), (18, 4), (19, 1), (19, 2), (19, 3), (19, 4), (19, 5), (20, 1), (20, 2), (20, 3), (21, 1), (22, 1), (22, 2), (22, 3), (22, 4), (22, 5), (22, 6), (22, 7), (22, 8), (23, 1), (23, 2), (24, 1), (24, 2), (24, 3), (25, 1), (25, 2), (25, 3), (25, 4), (25, 5)], and a valid choice should be a subset of it. 

All your current available actions are listed as above (all are without quotes and square brackets). Please don't mix up it with the action space of previous steps.

Understanding directions:
Directions like 'front', 'rear', 'left', and 'right' are defined relative to the robot's perspective and the heading of the object or platform you are interacting with. The heading of a ground object is not fixed and depends on where the robot is standing and facing.

Empty Surfaces (Platforms): When interacting with an empty platform, imagine it divided into a 3x3 grid. The direction aligned with the heading (also, the FARTHEST middle space to you, NOT THE NEAREST MIDDLE SPACE) is labeled "front," with the remaining regions proceeding counterclockwise as "front-left," "left," "rear-left," "rear," "rear-right," "right," and "front-right."  The grid is conceptually numbered as follows, based on these directions:

    [front-left, front, front-right],
    [left, center, right],
    [rear-left, rear, rear-right]

Occupied Surfaces (Objects on Platforms): For objects on platforms, imagine extending the edges of the object's bounding box to the platform's boundaries. The resulting regions define the object's 8 receptacles, and the direction is defined similarly.
Here are some useful suggestions for you to better finish the task: 

1. Your extra actions do not affect the success of the task, you only need to keep your steps within the total step limit.

2. When you want to place an object but cannot find enough space, you can try changing your standing position to find the correct orientation, try combining multiple receptacles, or remove objects that are blocking your way first(although you generally do not need to do this). Specifically, for the task asking you put object to empty platforms, try combining adjacent receptacles may be very useful.

3. For tasks that involve placing objects near other objects, please remember to get a focused image of the target object and examine it carefully. This is important because the receptacles may not be intuitive. 

4. Important: Don't output anything other than the action you want to take. You will only receive the same hint informing you your invalid action. If you make mistakes, just start output correct actions from the next prompt without any apologize.

Current task:

Move kitchenware\_white\_paper\_towel\_holder between lighting\_blue\_base\_table\_lamp and lighting\_blue\_base\_table\_lamp

Steps used: 7/20. You can only take at most 20 steps, so hurry if you've almost used all of them!

[*See Fig.~\ref{fig:success_step7}]

\textcolor{purple}{Here are 4 images, named ['HoldingOccupiedPlatform7\_1\_out\_of\_4.png', 'HoldingOccupiedPlatform7\_2\_out\_of\_4.png', 'HoldingOccupiedPlatform7\_3\_out\_of\_4.png', 'HoldingOccupiedPlatform7\_4\_out\_of\_4.png'] respectively.}

\textcolor{purple}{Their description is as follows: These image show the platform you are currently on in one view, with regions or objects marked with numbers for you to place objects accurately. The platform is object\_kitchen\_counter\_1\_body\_platform\_1.}

\textcolor{purple}{If the platform has no objects, a 3x3 grid will be marked on the platform to help you place objects, with each grid marked in numbers from 1 to 9. If there are objects on the platform, guiding lines will indicate the object numbers. Note that  if you see multiple images, it means the platform is too long to be captured in one image, and it's guaranteed that the same object will marked in THE SAME COLOR GUIDELINES AND THE SAME NUMBER. If you think you can't determine the number of an item in one picture, you can go through other pictures and see if this item is marked more clearly.}

\hdashrule{\linewidth}{0.4pt}{2mm}

\textcolor{blue}
{VLM Response: place\_r}

\hdashrule{\linewidth}{0.4pt}{2mm}

User Message:

\textbf{
The object has been placed successfully. You can now call end if you think you've finished the task correctly, or can also pick up the object again if you think the placement is not correct.
}

You are currently at object\_kitchen\_counter\_1\_body\_platform\_1.

And you are holding nothing.

Your available action space:

- ['go\_to\_object\_sofa\_10\_platform\_0', 'go\_to\_object\_sofa\_10\_platform\_1', 'go\_to\_object\_sofa\_10\_platform\_2', 'go\_to\_object\_table\_04\_13\_platform\_0', 'go\_to\_object\_chair\_04\_46\_platform\_0', 'go\_to\_object\_chair\_04\_47\_platform\_0', 'go\_to\_object\_chair\_05\_7\_platform\_0', 'go\_to\_object\_chair\_05\_8\_platform\_0', 'go\_to\_object\_stool\_02\_18\_platform\_0', 'go\_to\_object\_stool\_02\_6\_platform\_0', 'go\_to\_object\_rack\_01\_75\_platform\_0', 'go\_to\_object\_rack\_01\_75\_platform\_1', 'go\_to\_object\_rack\_01\_75\_platform\_2', 'go\_to\_object\_kitchen\_counter\_1\_body\_platform\_0', 'go\_to\_object\_kitchen\_counter\_1\_body\_platform\_1', 'go\_to\_object\_fridge\_0\_body\_platform\_1', 'go\_to\_object\_fridge\_0\_body\_platform\_3', 'go\_to\_object\_fridge\_0\_body\_platform\_4', 'go\_to\_object\_fridge\_0\_body\_platform\_5', 'go\_to\_object\_fridge\_0\_body\_platform\_6', 'go\_to\_object\_fridge\_0\_body\_platform\_7', 'go\_to\_object\_bin\_03\_3\_platform\_0', 'go\_to\_object\_wall\_cabinet\_01\_4\_platform\_0', 'go\_to\_object\_wall\_cabinet\_01\_4\_platform\_1', 'go\_to\_object\_wall\_cabinet\_01\_4\_platform\_2', 'go\_to\_object\_wall\_cabinet\_01\_4\_platform\_3', 'go\_to\_object\_wall\_cabinet\_01\_4\_platform\_4', 'go\_to\_object\_wall\_cabinet\_01\_4\_platform\_5', 'go\_to\_object\_wall\_cabinet\_01\_4\_platform\_6', 'go\_to\_object\_table\_03\_14\_platform\_1', 'go\_to\_object\_table\_01\_48\_platform\_0', 'go\_to\_object\_wall\_cabinet\_02\_21\_platform\_0', 'go\_to\_object\_wall\_cabinet\_02\_21\_platform\_1', 'go\_to\_object\_wall\_cabinet\_02\_21\_platform\_2', 'go\_to\_object\_wall\_cabinet\_02\_21\_platform\_3', 'go\_to\_object\_wall\_cabinet\_02\_21\_platform\_4', 'go\_to\_object\_wall\_cabinet\_02\_21\_platform\_5', 'go\_to\_object\_wall\_cabinet\_02\_21\_platform\_6', 'go\_to\_object\_chestOfDrawers\_01\_2\_body\_platform\_0', 'go\_to\_object\_chestOfDrawers\_01\_2\_body\_platform\_1', 'go\_to\_object\_chestOfDrawers\_01\_2\_body\_platform\_2', 'go\_to\_object\_chestOfDrawers\_01\_2\_body\_platform\_3', 'go\_to\_object\_chestOfDrawers\_01\_2\_body\_platform\_4', 'go\_to\_object\_chestOfDrawers\_01\_2\_body\_platform\_5', 'go\_to\_object\_chestOfDrawers\_01\_2\_body\_platform\_6', 'go\_to\_object\_table\_02\_40\_platform\_0', 'go\_to\_object\_table\_02\_40\_platform\_1', 'go\_to\_object\_table\_02\_40\_platform\_2', 'go\_to\_object\_tvstand\_88\_platform\_0', 'go\_to\_object\_cabinet\_3\_body\_platform\_0', 'go\_to\_object\_cabinet\_3\_body\_platform\_1']
- ['pick\_object\_1\_of\_current\_platform', 'pick\_object\_2\_of\_current\_platform', 'pick\_object\_3\_of\_current\_platform', 'pick\_object\_4\_of\_current\_platform', 'pick\_object\_5\_of\_current\_platform', 'pick\_object\_6\_of\_current\_platform', 'pick\_object\_7\_of\_current\_platform', 'pick\_object\_8\_of\_current\_platform', 'pick\_object\_9\_of\_current\_platform', 'pick\_object\_10\_of\_current\_platform', 'pick\_object\_11\_of\_current\_platform', 'pick\_object\_12\_of\_current\_platform', 'pick\_object\_13\_of\_current\_platform', 'pick\_object\_14\_of\_current\_platform', 'pick\_object\_15\_of\_current\_platform', 'pick\_object\_16\_of\_current\_platform', 'pick\_object\_17\_of\_current\_platform', 'pick\_object\_18\_of\_current\_platform', 'pick\_object\_19\_of\_current\_platform', 'pick\_object\_20\_of\_current\_platform', 'pick\_object\_21\_of\_current\_platform', 'pick\_object\_22\_of\_current\_platform', 'pick\_object\_23\_of\_current\_platform', 'pick\_object\_24\_of\_current\_platform', 'pick\_object\_25\_of\_current\_platform', 'pick\_object\_26\_of\_current\_platform']
- ['show\_receptacle\_of\_object\_1\_of\_current\_platform', 'show\_receptacle\_of\_object\_2\_of\_current\_platform', 'show\_receptacle\_of\_object\_3\_of\_current\_platform', 'show\_receptacle\_of\_object\_4\_of\_current\_platform', 'show\_receptacle\_of\_object\_5\_of\_current\_platform', 'show\_receptacle\_of\_object\_6\_of\_current\_platform', 'show\_receptacle\_of\_object\_7\_of\_current\_platform', 'show\_receptacle\_of\_object\_8\_of\_current\_platform', 'show\_receptacle\_of\_object\_9\_of\_current\_platform', 'show\_receptacle\_of\_object\_10\_of\_current\_platform', 'show\_receptacle\_of\_object\_11\_of\_current\_platform', 'show\_receptacle\_of\_object\_12\_of\_current\_platform', 'show\_receptacle\_of\_object\_13\_of\_current\_platform', 'show\_receptacle\_of\_object\_14\_of\_current\_platform', 'show\_receptacle\_of\_object\_15\_of\_current\_platform', 'show\_receptacle\_of\_object\_16\_of\_current\_platform', 'show\_receptacle\_of\_object\_17\_of\_current\_platform', 'show\_receptacle\_of\_object\_18\_of\_current\_platform', 'show\_receptacle\_of\_object\_19\_of\_current\_platform', 'show\_receptacle\_of\_object\_20\_of\_current\_platform', 'show\_receptacle\_of\_object\_21\_of\_current\_platform', 'show\_receptacle\_of\_object\_22\_of\_current\_platform', 'show\_receptacle\_of\_object\_23\_of\_current\_platform', 'show\_receptacle\_of\_object\_24\_of\_current\_platform', 'show\_receptacle\_of\_object\_25\_of\_current\_platform', 'show\_receptacle\_of\_object\_26\_of\_current\_platform']
- `CALL\_END`: Call end (terminate the task).
- `change\_view`: Change the viewing angle to observe the scene from another perspective.

Choose one of the available actions.

All your current available actions are listed as above (all are without quotes and square brackets). Please don't mix up it with the action space of previous steps.

Here are some useful suggestions for you to better finish the task: 

1. Your extra actions do not affect the success of the task, you only need to keep your steps within the total step limit.

2. When you want to place an object but cannot find enough space, you can try changing your standing position to find the correct orientation, try combining multiple receptacles, or remove objects that are blocking your way first(although you generally do not need to do this). Specifically, for the task asking you put object to empty platforms, try combining adjacent receptacles may be very useful.

3. For tasks that involve placing objects near other objects, please remember to get a focused image of the target object and examine it carefully. This is important because the receptacles may not be intuitive. 

4. Important: Don't output anything other than the action you want to take. You will only receive the same hint informing you your invalid action. If you make mistakes, just start output correct actions from the next prompt without any apologize.

Current task:

Move kitchenware\_white\_paper\_towel\_holder between lighting\_blue\_base\_table\_lamp and lighting\_blue\_base\_table\_lamp

Steps used: 8/20. You can only take at most 20 steps, so hurry if you've almost used all of them!

[*See Fig.~\ref{fig:success_step8}]

\textcolor{purple}
{Here are 4 images, named ['Idle\_8\_1\_out\_of\_4.png', 'Idle\_8\_2\_out\_of\_4.png', 'Idle\_8\_3\_out\_of\_4.png', 'Idle\_8\_4\_out\_of\_4.png'] respectively.}

\textcolor{purple}
{Their description is as follows: These images show the platform you are currently on in one view, with regions or objects marked with numbers for you to place objects accurately. The platform is object\_kitchen\_counter\_1\_body\_platform\_1.}

\textcolor{purple}
{If the platform has no objects, a 3x3 grid will be marked on the platform to help you place objects, with each grid marked in numbers from 1 to 9. If there are objects on the platform, guiding lines will indicate the object numbers. Note that  if you see multiple images, it means the platform is too long to be captured in one image, and it's guaranteed that the same object will marked in THE SAME COLOR GUIDELINES AND THE SAME NUMBER. If you think you can't determine the number of an item in one picture, you can go through other pictures and see if this item is marked more clearly.}

\hdashrule{\linewidth}{0.4pt}{2mm}

\textcolor{blue}
{VLM\_response: pick\_object\_1\_of\_current\_platform}

\hdashrule{\linewidth}{0.4pt}{2mm}

User Message:

You are currently at object\_kitchen\_counter\_1\_body\_platform\_1.

And you are holding kitchenware\_white\_paper\_towel\_holder.

Your available action space:
-['go\_to\_object\_sofa\_10\_platform\_0', 'go\_to\_object\_sofa\_10\_platform\_1', 'go\_to\_object\_sofa\_10\_platform\_2', 'go\_to\_object\_table\_04\_13\_platform\_0', 'go\_to\_object\_chair\_04\_46\_platform\_0', 'go\_to\_object\_chair\_04\_47\_platform\_0', 'go\_to\_object\_chair\_05\_7\_platform\_0', 'go\_to\_object\_chair\_05\_8\_platform\_0', 'go\_to\_object\_stool\_02\_18\_platform\_0', 'go\_to\_object\_stool\_02\_6\_platform\_0', 'go\_to\_object\_rack\_01\_75\_platform\_0', 'go\_to\_object\_rack\_01\_75\_platform\_1', 'go\_to\_object\_rack\_01\_75\_platform\_2', 'go\_to\_object\_kitchen\_counter\_1\_body\_platform\_0', 'go\_to\_object\_kitchen\_counter\_1\_body\_platform\_1', 'go\_to\_object\_fridge\_0\_body\_platform\_1', 'go\_to\_object\_fridge\_0\_body\_platform\_3', 'go\_to\_object\_fridge\_0\_body\_platform\_4', 'go\_to\_object\_fridge\_0\_body\_platform\_5', 'go\_to\_object\_fridge\_0\_body\_platform\_6', 'go\_to\_object\_fridge\_0\_body\_platform\_7', 'go\_to\_object\_bin\_03\_3\_platform\_0', 'go\_to\_object\_wall\_cabinet\_01\_4\_platform\_0', 'go\_to\_object\_wall\_cabinet\_01\_4\_platform\_1', 'go\_to\_object\_wall\_cabinet\_01\_4\_platform\_2', 'go\_to\_object\_wall\_cabinet\_01\_4\_platform\_3', 'go\_to\_object\_wall\_cabinet\_01\_4\_platform\_4', 'go\_to\_object\_wall\_cabinet\_01\_4\_platform\_5', 'go\_to\_object\_wall\_cabinet\_01\_4\_platform\_6', 'go\_to\_object\_table\_03\_14\_platform\_1', 'go\_to\_object\_table\_01\_48\_platform\_0', 'go\_to\_object\_wall\_cabinet\_02\_21\_platform\_0', 'go\_to\_object\_wall\_cabinet\_02\_21\_platform\_1', 'go\_to\_object\_wall\_cabinet\_02\_21\_platform\_2', 'go\_to\_object\_wall\_cabinet\_02\_21\_platform\_3', 'go\_to\_object\_wall\_cabinet\_02\_21\_platform\_4', 'go\_to\_object\_wall\_cabinet\_02\_21\_platform\_5', 'go\_to\_object\_wall\_cabinet\_02\_21\_platform\_6', 'go\_to\_object\_chestOfDrawers\_01\_2\_body\_platform\_0', 'go\_to\_object\_chestOfDrawers\_01\_2\_body\_platform\_1', 'go\_to\_object\_chestOfDrawers\_01\_2\_body\_platform\_2', 'go\_to\_object\_chestOfDrawers\_01\_2\_body\_platform\_3', 'go\_to\_object\_chestOfDrawers\_01\_2\_body\_platform\_4', 'go\_to\_object\_chestOfDrawers\_01\_2\_body\_platform\_5', 'go\_to\_object\_chestOfDrawers\_01\_2\_body\_platform\_6', 'go\_to\_object\_table\_02\_40\_platform\_0', 'go\_to\_object\_table\_02\_40\_platform\_1', 'go\_to\_object\_table\_02\_40\_platform\_2', 'go\_to\_object\_tvstand\_88\_platform\_0', 'go\_to\_object\_cabinet\_3\_body\_platform\_0', 'go\_to\_object\_cabinet\_3\_body\_platform\_1']
- ['show\_receptacle\_of\_object\_1\_of\_current\_platform', 'show\_receptacle\_of\_object\_2\_of\_current\_platform', 'show\_receptacle\_of\_object\_3\_of\_current\_platform', 'show\_receptacle\_of\_object\_4\_of\_current\_platform', 'show\_receptacle\_of\_object\_5\_of\_current\_platform', 'show\_receptacle\_of\_object\_6\_of\_current\_platform', 'show\_receptacle\_of\_object\_7\_of\_current\_platform', 'show\_receptacle\_of\_object\_8\_of\_current\_platform', 'show\_receptacle\_of\_object\_9\_of\_current\_platform', 'show\_receptacle\_of\_object\_10\_of\_current\_platform', 'show\_receptacle\_of\_object\_11\_of\_current\_platform', 'show\_receptacle\_of\_object\_12\_of\_current\_platform', 'show\_receptacle\_of\_object\_13\_of\_current\_platform', 'show\_receptacle\_of\_object\_14\_of\_current\_platform', 'show\_receptacle\_of\_object\_15\_of\_current\_platform', 'show\_receptacle\_of\_object\_16\_of\_current\_platform', 'show\_receptacle\_of\_object\_17\_of\_current\_platform', 'show\_receptacle\_of\_object\_18\_of\_current\_platform', 'show\_receptacle\_of\_object\_19\_of\_current\_platform', 'show\_receptacle\_of\_object\_20\_of\_current\_platform', 'show\_receptacle\_of\_object\_21\_of\_current\_platform', 'show\_receptacle\_of\_object\_22\_of\_current\_platform', 'show\_receptacle\_of\_object\_23\_of\_current\_platform', 'show\_receptacle\_of\_object\_24\_of\_current\_platform', 'show\_receptacle\_of\_object\_25\_of\_current\_platform']
- `CALL\_END`: Call end (terminate the task).
- `change\_view`: Change the viewing angle to observe the scene from another perspective.
- `place\_r`:Place the object in an arbitrary place on the platform. - `place\_s\_[({object\_idx1},{dir\_idx1})...]`: Place the object in the union area of regions with indices {(object\_idx1,region\_idx1)}(object\_idx1's region\_idx1 space)  etc. All available pairs are in  [(1, 1), (2, 1), (2, 2), (2, 3), (3, 1), (3, 2), (3, 3), (3, 4), (3, 5), (3, 6), (3, 7), (4, 1), (4, 2), (5, 1), (5, 2), (5, 3), (5, 4), (5, 5), (7, 1), (7, 2), (7, 3), (8, 1), (9, 1), (10, 1), (10, 2), (10, 3), (11, 1), (11, 2), (11, 3), (11, 4), (11, 5), (12, 1), (12, 2), (12, 3), (13, 1), (13, 2), (13, 3), (13, 4), (14, 1), (14, 2), (14, 3), (14, 4), (14, 5), (14, 6), (14, 7), (14, 8), (15, 1), (15, 2), (16, 1), (16, 2), (16, 3), (17, 1), (17, 2), (17, 3), (17, 4), (17, 5), (18, 1), (18, 2), (18, 3), (18, 4), (19, 1), (19, 2), (19, 3), (19, 4), (19, 5), (19, 6), (20, 1), (20, 2), (20, 3), (20, 4), (21, 1), (21, 2), (21, 3), (21, 4), (22, 1), (22, 2), (22, 3), (22, 4), (23, 1), (23, 2), (23, 3), (24, 1), (24, 2), (25, 1)], and a valid choice should be a subset of it. 

All your current available actions are listed as above (all are without quotes and square brackets). Please don't mix up it with the action space of previous steps.

Understanding directions:
Directions like 'front', 'rear', 'left', and 'right' are defined relative to the robot's perspective and the heading of the object or platform you are interacting with. The heading of a ground object is not fixed and depends on where the robot is standing and facing.

Empty Surfaces (Platforms): When interacting with an empty platform, imagine it divided into a 3x3 grid. The direction aligned with the heading (also, the FARTHEST middle space to you, NOT THE NEAREST MIDDLE SPACE) is labeled "front," with the remaining regions proceeding counterclockwise as "front-left," "left," "rear-left," "rear," "rear-right," "right," and "front-right."  The grid is conceptually numbered as follows, based on these directions:

    [front-left, front, front-right],
    [left, center, right],
    [rear-left, rear, rear-right]

Occupied Surfaces (Objects on Platforms): For objects on platforms, imagine extending the edges of the object's bounding box to the platform's boundaries. The resulting regions define the object's 8 receptacles, and the direction is defined similarly.
Here are some useful suggestions for you to better finish the task: 

1. Your extra actions do not affect the success of the task, you only need to keep your steps within the total step limit.

2. When you want to place an object but cannot find enough space, you can try changing your standing position to find the correct orientation, try combining multiple receptacles, or remove objects that are blocking your way first(although you generally do not need to do this). Specifically, for the task asking you put object to empty platforms, try combining adjacent receptacles may be very useful.

3. For tasks that involve placing objects near other objects, please remember to get a focused image of the target object and examine it carefully. This is important because the receptacles may not be intuitive. 

4. Important: Don't output anything other than the action you want to take. You will only receive the same hint informing you your invalid action. If you make mistakes, just start output correct actions from the next prompt without any apologize.

Current task:

Move kitchenware\_white\_paper\_towel\_holder between lighting\_blue\_base\_table\_lamp and lighting\_blue\_base\_table\_lamp

Steps used: 9/20. You can only take at most 20 steps, so hurry if you've almost used all of them!

[*See Fig.~\ref{fig:success_step9}]

\textcolor{purple}{Here are 4 images, named ['HoldingOccupiedPlatform9\_1\_out\_of\_4.png', 'HoldingOccupiedPlatform9\_2\_out\_of\_4.png', 'HoldingOccupiedPlatform9\_3\_out\_of\_4.png', 'HoldingOccupiedPlatform9\_4\_out\_of\_4.png'] respectively.}

\textcolor{purple}{Their description is as follows: These image show the platform you are currently on in one view, with regions or objects marked with numbers for you to place objects accurately. The platform is object\_kitchen\_counter\_1\_body\_platform\_1.}

\textcolor{purple}{If the platform has no objects, a 3x3 grid will be marked on the platform to help you place objects, with each grid marked in numbers from 1 to 9. If there are objects on the platform, guiding lines will indicate the object numbers. Note that  if you see multiple images, it means the platform is too long to be captured in one image, and it's guaranteed that the same object will marked in THE SAME COLOR GUIDELINES AND THE SAME NUMBER. If you think you can't determine the number of an item in one picture, you can go through other pictures and see if this item is marked more clearly.}

\hdashrule{\linewidth}{0.4pt}{2mm}

\textcolor{blue}
{VLM Response: go\_to\_object\_table\_02\_40\_platform\_2}

\hdashrule{\linewidth}{0.4pt}{2mm}

User Message:

You are currently at object\_table\_02\_40\_platform\_2.

And you are holding kitchenware\_white\_paper\_towel\_holder.

Your available action space:

-['go\_to\_object\_sofa\_10\_platform\_0', 'go\_to\_object\_sofa\_10\_platform\_1', 'go\_to\_object\_sofa\_10\_platform\_2', 'go\_to\_object\_table\_04\_13\_platform\_0', 'go\_to\_object\_chair\_04\_46\_platform\_0', 'go\_to\_object\_chair\_04\_47\_platform\_0', 'go\_to\_object\_chair\_05\_7\_platform\_0', 'go\_to\_object\_chair\_05\_8\_platform\_0', 'go\_to\_object\_stool\_02\_18\_platform\_0', 'go\_to\_object\_stool\_02\_6\_platform\_0', 'go\_to\_object\_rack\_01\_75\_platform\_0', 'go\_to\_object\_rack\_01\_75\_platform\_1', 'go\_to\_object\_rack\_01\_75\_platform\_2', 'go\_to\_object\_kitchen\_counter\_1\_body\_platform\_0', 'go\_to\_object\_kitchen\_counter\_1\_body\_platform\_1', 'go\_to\_object\_fridge\_0\_body\_platform\_1', 'go\_to\_object\_fridge\_0\_body\_platform\_3', 'go\_to\_object\_fridge\_0\_body\_platform\_4', 'go\_to\_object\_fridge\_0\_body\_platform\_5', 'go\_to\_object\_fridge\_0\_body\_platform\_6', 'go\_to\_object\_fridge\_0\_body\_platform\_7', 'go\_to\_object\_bin\_03\_3\_platform\_0', 'go\_to\_object\_wall\_cabinet\_01\_4\_platform\_0', 'go\_to\_object\_wall\_cabinet\_01\_4\_platform\_1', 'go\_to\_object\_wall\_cabinet\_01\_4\_platform\_2', 'go\_to\_object\_wall\_cabinet\_01\_4\_platform\_3', 'go\_to\_object\_wall\_cabinet\_01\_4\_platform\_4', 'go\_to\_object\_wall\_cabinet\_01\_4\_platform\_5', 'go\_to\_object\_wall\_cabinet\_01\_4\_platform\_6', 'go\_to\_object\_table\_03\_14\_platform\_1', 'go\_to\_object\_table\_01\_48\_platform\_0', 'go\_to\_object\_wall\_cabinet\_02\_21\_platform\_0', 'go\_to\_object\_wall\_cabinet\_02\_21\_platform\_1', 'go\_to\_object\_wall\_cabinet\_02\_21\_platform\_2', 'go\_to\_object\_wall\_cabinet\_02\_21\_platform\_3', 'go\_to\_object\_wall\_cabinet\_02\_21\_platform\_4', 'go\_to\_object\_wall\_cabinet\_02\_21\_platform\_5', 'go\_to\_object\_wall\_cabinet\_02\_21\_platform\_6', 'go\_to\_object\_chestOfDrawers\_01\_2\_body\_platform\_0', 'go\_to\_object\_chestOfDrawers\_01\_2\_body\_platform\_1', 'go\_to\_object\_chestOfDrawers\_01\_2\_body\_platform\_2', 'go\_to\_object\_chestOfDrawers\_01\_2\_body\_platform\_3', 'go\_to\_object\_chestOfDrawers\_01\_2\_body\_platform\_4', 'go\_to\_object\_chestOfDrawers\_01\_2\_body\_platform\_5', 'go\_to\_object\_chestOfDrawers\_01\_2\_body\_platform\_6', 'go\_to\_object\_table\_02\_40\_platform\_0', 'go\_to\_object\_table\_02\_40\_platform\_1', 'go\_to\_object\_table\_02\_40\_platform\_2', 'go\_to\_object\_tvstand\_88\_platform\_0', 'go\_to\_object\_cabinet\_3\_body\_platform\_0', 'go\_to\_object\_cabinet\_3\_body\_platform\_1']
- ['show\_receptacle\_of\_object\_1\_of\_current\_platform', 'show\_receptacle\_of\_object\_2\_of\_current\_platform']
- `CALL\_END`: Call end (terminate the task).
- `change\_view`: Change the viewing angle to observe the scene from another perspective.
- `place\_r`:Place the object in an arbitrary place on the platform. - `place\_s\_[({object\_idx1},{dir\_idx1})...]`: Place the object in the union area of regions with indices {(object\_idx1,region\_idx1)}(object\_idx1's region\_idx1 space)  etc. All available pairs are in [(1, 1), (1, 2), (1, 3), (1, 4), (1, 5), (1, 6), (1, 7), (1, 8), (2, 1), (2, 2), (2, 3), (2, 4), (2, 5), (2, 6), (2, 7), (2, 8)], and a valid choice should be a subset of it.

All your current available actions are listed as above (all are without quotes and square brackets). Please don't mix up it with the action space of previous steps.

Understanding directions:
Directions like 'front', 'rear', 'left', and 'right' are defined relative to the robot's perspective and the heading of the object or platform you are interacting with. The heading of a ground object is not fixed and depends on where the robot is standing and facing.

Empty Surfaces (Platforms): When interacting with an empty platform, imagine it divided into a 3x3 grid. The direction aligned with the heading (also, the FARTHEST middle space to you, NOT THE NEAREST MIDDLE SPACE) is labeled "front," with the remaining regions proceeding counterclockwise as "front-left," "left," "rear-left," "rear," "rear-right," "right," and "front-right."  The grid is conceptually numbered as follows, based on these directions:

    [front-left, front, front-right],
    
    [left, center, right],
    
    [rear-left, rear, rear-right]

Occupied Surfaces (Objects on Platforms): For objects on platforms, imagine extending the edges of the object's bounding box to the platform's boundaries. The resulting regions define the object's 8 receptacles, and the direction is defined similarly.
Here are some useful suggestions for you to better finish the task: 

1. Your extra actions do not affect the success of the task, you only need to keep your steps within the total step limit.

2. When you want to place an object but cannot find enough space, you can try changing your standing position to find the correct orientation, try combining multiple receptacles, or remove objects that are blocking your way first(although you generally do not need to do this). Specifically, for the task asking you put object to empty platforms, try combining adjacent receptacles may be very useful.

3. For tasks that involve placing objects near other objects, please remember to get a focused image of the target object and examine it carefully. This is important because the receptacles may not be intuitive. 

4. Important: Don't output anything other than the action you want to take. You will only receive the same hint informing you your invalid action. If you make mistakes, just start output correct actions from the next prompt without any apologize.

Current task:

Move kitchenware\_white\_paper\_towel\_holder between lighting\_blue\_base\_table\_lamp and lighting\_blue\_base\_table\_lamp

Steps used: 10/20. You can only take at most 20 steps, so hurry if you've almost used all of them!

[*See Fig.~\ref{fig:success_step10}]

\textcolor{purple}
{Here are 2 images, named ['HoldingOccupiedPlatform10\_1\_out\_of\_2.png', 'HoldingOccupiedPlatform10\_2\_out\_of\_2.png'] respectively.}

\textcolor{purple}
{Their description are as follows: These image show the platform you are currently on in one view, with regions or objects marked with numbers for you to place objects accurately. The platform is object\_table\_02\_40\_platform\_2.}

\textcolor{purple}
{If the platform has no objects, a 3x3 grid will be marked on the platform to help you place objects, with each grid marked in numbers from 1 to 9. If there are objects on the platform, guiding lines will indicate the object numbers. Note that  if you see multiple images, it means the platform is too long to be captured in one image, and it's guaranteed that the same object will marked in THE SAME COLOR GUIDELINES AND THE SAME NUMBER. If you think you can't determine the number of an item in one picture, you can go through other pictures and see if this item is marked more clearly.}

\hdashrule{\linewidth}{0.4pt}{2mm}

\textcolor{blue}
{VLM Response: show\_receptacle\_of\_object\_1\_of\_current\_platform}

\hdashrule{\linewidth}{0.4pt}{2mm}

User Message:

[*See Fig.~\ref{fig:success_step11} \textit{ShowReceptacle\_11.png}]

\textcolor{purple}{Here are 1 images, named ['ShowReceptacle\_11.png'] respectively.}

\textcolor{purple}{Their description is as follows: This image shows the available receptacles around object 1 on the platform, you see this because you've taken "show\_receptacle" action previously. The numbers from 1 to 8 indicate the available receptacles. Not all numbers from 1 to 8 may be present, as some receptacles might be too small or blocked. To use the numbers shown in the image to specify where you want to place the object you are holding, you can include (1, {receptacles\_idx} (receptacles\_idx in [1, 8]) in the output action.}

You are currently at object\_table\_02\_40\_platform\_2.

And you are holding kitchenware\_white\_paper\_towel\_holder.

Your available action space:

-['go\_to\_object\_sofa\_10\_platform\_0', 'go\_to\_object\_sofa\_10\_platform\_1', 'go\_to\_object\_sofa\_10\_platform\_2', 'go\_to\_object\_table\_04\_13\_platform\_0', 'go\_to\_object\_chair\_04\_46\_platform\_0', 'go\_to\_object\_chair\_04\_47\_platform\_0', 'go\_to\_object\_chair\_05\_7\_platform\_0', 'go\_to\_object\_chair\_05\_8\_platform\_0', 'go\_to\_object\_stool\_02\_18\_platform\_0', 'go\_to\_object\_stool\_02\_6\_platform\_0', 'go\_to\_object\_rack\_01\_75\_platform\_0', 'go\_to\_object\_rack\_01\_75\_platform\_1', 'go\_to\_object\_rack\_01\_75\_platform\_2', 'go\_to\_object\_kitchen\_counter\_1\_body\_platform\_0', 'go\_to\_object\_kitchen\_counter\_1\_body\_platform\_1', 'go\_to\_object\_fridge\_0\_body\_platform\_1', 'go\_to\_object\_fridge\_0\_body\_platform\_3', 'go\_to\_object\_fridge\_0\_body\_platform\_4', 'go\_to\_object\_fridge\_0\_body\_platform\_5', 'go\_to\_object\_fridge\_0\_body\_platform\_6', 'go\_to\_object\_fridge\_0\_body\_platform\_7', 'go\_to\_object\_bin\_03\_3\_platform\_0', 'go\_to\_object\_wall\_cabinet\_01\_4\_platform\_0', 'go\_to\_object\_wall\_cabinet\_01\_4\_platform\_1', 'go\_to\_object\_wall\_cabinet\_01\_4\_platform\_2', 'go\_to\_object\_wall\_cabinet\_01\_4\_platform\_3', 'go\_to\_object\_wall\_cabinet\_01\_4\_platform\_4', 'go\_to\_object\_wall\_cabinet\_01\_4\_platform\_5', 'go\_to\_object\_wall\_cabinet\_01\_4\_platform\_6', 'go\_to\_object\_table\_03\_14\_platform\_1', 'go\_to\_object\_table\_01\_48\_platform\_0', 'go\_to\_object\_wall\_cabinet\_02\_21\_platform\_0', 'go\_to\_object\_wall\_cabinet\_02\_21\_platform\_1', 'go\_to\_object\_wall\_cabinet\_02\_21\_platform\_2', 'go\_to\_object\_wall\_cabinet\_02\_21\_platform\_3', 'go\_to\_object\_wall\_cabinet\_02\_21\_platform\_4', 'go\_to\_object\_wall\_cabinet\_02\_21\_platform\_5', 'go\_to\_object\_wall\_cabinet\_02\_21\_platform\_6', 'go\_to\_object\_chestOfDrawers\_01\_2\_body\_platform\_0', 'go\_to\_object\_chestOfDrawers\_01\_2\_body\_platform\_1', 'go\_to\_object\_chestOfDrawers\_01\_2\_body\_platform\_2', 'go\_to\_object\_chestOfDrawers\_01\_2\_body\_platform\_3', 'go\_to\_object\_chestOfDrawers\_01\_2\_body\_platform\_4', 'go\_to\_object\_chestOfDrawers\_01\_2\_body\_platform\_5', 'go\_to\_object\_chestOfDrawers\_01\_2\_body\_platform\_6', 'go\_to\_object\_table\_02\_40\_platform\_0', 'go\_to\_object\_table\_02\_40\_platform\_1', 'go\_to\_object\_table\_02\_40\_platform\_2', 'go\_to\_object\_tvstand\_88\_platform\_0', 'go\_to\_object\_cabinet\_3\_body\_platform\_0', 'go\_to\_object\_cabinet\_3\_body\_platform\_1']
- ['show\_receptacle\_of\_object\_1\_of\_current\_platform', 'show\_receptacle\_of\_object\_2\_of\_current\_platform']
- `CALL\_END`: Call end (terminate the task).
- `change\_view`: Change the viewing angle to observe the scene from another perspective.
- `place\_r`:Place the object in an arbitrary place on the platform. - `place\_s\_[({object\_idx1},{dir\_idx1})...]`: Place the object in the union area of regions with indices {(object\_idx1,region\_idx1)}(object\_idx1's region\_idx1 space)  etc. All available pairs are in [(1, 1), (1, 2), (1, 3), (1, 4), (1, 5), (1, 6), (1, 7), (1, 8), (2, 1), (2, 2), (2, 3), (2, 4), (2, 5), (2, 6), (2, 7), (2, 8)], and a valid choice should be a subset of it.

All your current available actions are listed as above (all are without quotes and square brackets). Please don't mix up it with the action space of previous steps.

Understanding directions:
Directions like 'front', 'rear', 'left', and 'right' are defined relative to the robot's perspective and the heading of the object or platform you are interacting with. The heading of a ground object is not fixed and depends on where the robot is standing and facing.

Empty Surfaces (Platforms): When interacting with an empty platform, imagine it divided into a 3x3 grid. The direction aligned with the heading (also, the FARTHEST middle space to you, NOT THE NEAREST MIDDLE SPACE) is labeled "front," with the remaining regions proceeding counterclockwise as "front-left," "left," "rear-left," "rear," "rear-right," "right," and "front-right."  The grid is conceptually numbered as follows, based on these directions:

    [front-left, front, front-right],
    
    [left, center, right],
    
    [rear-left, rear, rear-right]

Occupied Surfaces (Objects on Platforms): For objects on platforms, imagine extending the edges of the object's bounding box to the platform's boundaries. The resulting regions define the object's 8 receptacles, and the direction is defined similarly.
Here are some useful suggestions for you to better finish the task: 

1. Your extra actions do not affect the success of the task, you only need to keep your steps within the total step limit.

2. When you want to place an object but cannot find enough space, you can try changing your standing position to find the correct orientation, try combining multiple receptacles, or remove objects that are blocking your way first(although you generally do not need to do this). Specifically, for the task asking you put object to empty platforms, try combining adjacent receptacles may be very useful.

3. For tasks that involve placing objects near other objects, please remember to get a focused image of the target object and examine it carefully. This is important because the receptacles may not be intuitive. 

4. Important: Don't output anything other than the action you want to take. You will only receive the same hint informing you your invalid action. If you make mistakes, just start output correct actions from the next prompt without any apologize.

Current task:

Move kitchenware\_white\_paper\_towel\_holder between lighting\_blue\_base\_table\_lamp and lighting\_blue\_base\_table\_lamp

Steps used: 11/20. You can only take at most 20 steps, so hurry if you've almost used all of them!

[*See Fig.~\ref{fig:success_step11} \textit{HoldingOccupiedPlatform11\_1\_out\_of\_2.png},\textit{HoldingOccupiedPlatform11\_2\_out\_of\_2.png}]

\textcolor{purple}
{Here are 2 images, named ['HoldingOccupiedPlatform11\_1\_out\_of\_2.png', 'HoldingOccupiedPlatform11\_2\_out\_of\_2.png'] respectively.}

\textcolor{purple}
{Their description are as follows: These images show the platform you are currently on in one view, with regions or objects marked with numbers for you to place objects accurately. The platform is object\_table\_02\_40\_platform\_2.}

\textcolor{purple}
{If the platform has no objects, a 3x3 grid will be marked on the platform to help you place objects, with each grid marked in numbers from 1 to 9. If there are objects on the platform, guiding lines will indicate the object numbers. Note that  if you see multiple images, it means the platform is too long to be captured in one image, and it's guaranteed that the same object will marked in THE SAME COLOR GUIDELINES AND THE SAME NUMBER. If you think you can't determine the number of an item in one picture, you can go through other pictures and see if this item is marked more clearly.}

\hdashrule{\linewidth}{0.4pt}{2mm}

\textcolor{blue}
{VLM Response: show\_receptacle\_of\_object\_2\_of\_current\_platform}

\hdashrule{\linewidth}{0.4pt}{2mm}

User Message:

[*See Fig.~\ref{fig:success_step12} \textit{ShowReceptacle\_12.png}]

\textcolor{purple}{Here are 1 images, named ['ShowReceptacle\_12.png'] respectively.}

\textcolor{purple}{Their description is as follows: This image shows the available receptacles around object 1 on the platform, you see this because you've taken "show\_receptacle" action previously. The numbers from 1 to 8 indicate the available receptacles. Not all numbers from 1 to 8 may be present, as some receptacles might be too small or blocked. To use the numbers shown in the image to specify where you want to place the object you are holding, you can include (1, {receptacles\_idx} (receptacles\_idx in [1, 8]) in the output action.}

You are currently at object\_table\_02\_40\_platform\_2.

And you are holding kitchenware\_white\_paper\_towel\_holder.

Your available action space:

-['go\_to\_object\_sofa\_10\_platform\_0', 'go\_to\_object\_sofa\_10\_platform\_1', 'go\_to\_object\_sofa\_10\_platform\_2', 'go\_to\_object\_table\_04\_13\_platform\_0', 'go\_to\_object\_chair\_04\_46\_platform\_0', 'go\_to\_object\_chair\_04\_47\_platform\_0', 'go\_to\_object\_chair\_05\_7\_platform\_0', 'go\_to\_object\_chair\_05\_8\_platform\_0', 'go\_to\_object\_stool\_02\_18\_platform\_0', 'go\_to\_object\_stool\_02\_6\_platform\_0', 'go\_to\_object\_rack\_01\_75\_platform\_0', 'go\_to\_object\_rack\_01\_75\_platform\_1', 'go\_to\_object\_rack\_01\_75\_platform\_2', 'go\_to\_object\_kitchen\_counter\_1\_body\_platform\_0', 'go\_to\_object\_kitchen\_counter\_1\_body\_platform\_1', 'go\_to\_object\_fridge\_0\_body\_platform\_1', 'go\_to\_object\_fridge\_0\_body\_platform\_3', 'go\_to\_object\_fridge\_0\_body\_platform\_4', 'go\_to\_object\_fridge\_0\_body\_platform\_5', 'go\_to\_object\_fridge\_0\_body\_platform\_6', 'go\_to\_object\_fridge\_0\_body\_platform\_7', 'go\_to\_object\_bin\_03\_3\_platform\_0', 'go\_to\_object\_wall\_cabinet\_01\_4\_platform\_0', 'go\_to\_object\_wall\_cabinet\_01\_4\_platform\_1', 'go\_to\_object\_wall\_cabinet\_01\_4\_platform\_2', 'go\_to\_object\_wall\_cabinet\_01\_4\_platform\_3', 'go\_to\_object\_wall\_cabinet\_01\_4\_platform\_4', 'go\_to\_object\_wall\_cabinet\_01\_4\_platform\_5', 'go\_to\_object\_wall\_cabinet\_01\_4\_platform\_6', 'go\_to\_object\_table\_03\_14\_platform\_1', 'go\_to\_object\_table\_01\_48\_platform\_0', 'go\_to\_object\_wall\_cabinet\_02\_21\_platform\_0', 'go\_to\_object\_wall\_cabinet\_02\_21\_platform\_1', 'go\_to\_object\_wall\_cabinet\_02\_21\_platform\_2', 'go\_to\_object\_wall\_cabinet\_02\_21\_platform\_3', 'go\_to\_object\_wall\_cabinet\_02\_21\_platform\_4', 'go\_to\_object\_wall\_cabinet\_02\_21\_platform\_5', 'go\_to\_object\_wall\_cabinet\_02\_21\_platform\_6', 'go\_to\_object\_chestOfDrawers\_01\_2\_body\_platform\_0', 'go\_to\_object\_chestOfDrawers\_01\_2\_body\_platform\_1', 'go\_to\_object\_chestOfDrawers\_01\_2\_body\_platform\_2', 'go\_to\_object\_chestOfDrawers\_01\_2\_body\_platform\_3', 'go\_to\_object\_chestOfDrawers\_01\_2\_body\_platform\_4', 'go\_to\_object\_chestOfDrawers\_01\_2\_body\_platform\_5', 'go\_to\_object\_chestOfDrawers\_01\_2\_body\_platform\_6', 'go\_to\_object\_table\_02\_40\_platform\_0', 'go\_to\_object\_table\_02\_40\_platform\_1', 'go\_to\_object\_table\_02\_40\_platform\_2', 'go\_to\_object\_tvstand\_88\_platform\_0', 'go\_to\_object\_cabinet\_3\_body\_platform\_0', 'go\_to\_object\_cabinet\_3\_body\_platform\_1']
- ['show\_receptacle\_of\_object\_1\_of\_current\_platform', 'show\_receptacle\_of\_object\_2\_of\_current\_platform']
- `CALL\_END`: Call end (terminate the task).
- `change\_view`: Change the viewing angle to observe the scene from another perspective.
- `place\_r`:Place the object in an arbitrary place on the platform. - `place\_s\_[({object\_idx1},{dir\_idx1})...]`: Place the object in the union area of regions with indices {(object\_idx1,region\_idx1)}(object\_idx1's region\_idx1 space)  etc. All available pairs are in [(1, 1), (1, 2), (1, 3), (1, 4), (1, 5), (1, 6), (1, 7), (1, 8), (2, 1), (2, 2), (2, 3), (2, 4), (2, 5), (2, 6), (2, 7), (2, 8)], and a valid choice should be a subset of it.

All your current available actions are listed as above (all are without quotes and square brackets). Please don't mix up it with the action space of previous steps.

Understanding directions:
Directions like 'front', 'rear', 'left', and 'right' are defined relative to the robot's perspective and the heading of the object or platform you are interacting with. The heading of a ground object is not fixed and depends on where the robot is standing and facing.

Empty Surfaces (Platforms): When interacting with an empty platform, imagine it divided into a 3x3 grid. The direction aligned with the heading (also, the FARTHEST middle space to you, NOT THE NEAREST MIDDLE SPACE) is labeled "front," with the remaining regions proceeding counterclockwise as "front-left," "left," "rear-left," "rear," "rear-right," "right," and "front-right."  The grid is conceptually numbered as follows, based on these directions:

    [front-left, front, front-right],
    
    [left, center, right],
    
    [rear-left, rear, rear-right]

Occupied Surfaces (Objects on Platforms): For objects on platforms, imagine extending the edges of the object's bounding box to the platform's boundaries. The resulting regions define the object's 8 receptacles, and the direction is defined similarly.
Here are some useful suggestions for you to better finish the task: 

1. Your extra actions do not affect the success of the task, you only need to keep your steps within the total step limit.

2. When you want to place an object but cannot find enough space, you can try changing your standing position to find the correct orientation, try combining multiple receptacles, or remove objects that are blocking your way first(although you generally do not need to do this). Specifically, for the task asking you put object to empty platforms, try combining adjacent receptacles may be very useful.

3. For tasks that involve placing objects near other objects, please remember to get a focused image of the target object and examine it carefully. This is important because the receptacles may not be intuitive. 

4. Important: Don't output anything other than the action you want to take. You will only receive the same hint informing you your invalid action. If you make mistakes, just start output correct actions from the next prompt without any apologize.

Current task:

Move kitchenware\_white\_paper\_towel\_holder between lighting\_blue\_base\_table\_lamp and lighting\_blue\_base\_table\_lamp

Steps used: 12/20. You can only take at most 20 steps, so hurry if you've almost used all of them!

[*See Fig.~\ref{fig:success_step12} \textit{HoldingOccupiedPlatform12\_1\_out\_of\_2.png}, \textit{HoldingOccupiedPlatform12\_2\_out\_of\_2.png}]

\textcolor{purple}
{Here are 2 images, named ['HoldingOccupiedPlatform12\_1\_out\_of\_2.png', 'HoldingOccupiedPlatform12\_2\_out\_of\_2.png'] respectively.}

\textcolor{purple}
{Their description are as follows: These images show the platform you are currently on in one view, with regions or objects marked with numbers for you to place objects accurately. The platform is object\_table\_02\_40\_platform\_2.}

\textcolor{purple}
{If the platform has no objects, a 3x3 grid will be marked on the platform to help you place objects, with each grid marked in numbers from 1 to 9. If there are objects on the platform, guiding lines will indicate the object numbers. Note that  if you see multiple images, it means the platform is too long to be captured in one image, and it's guaranteed that the same object will marked in THE SAME COLOR GUIDELINES AND THE SAME NUMBER. If you think you can't determine the number of an item in one picture, you can go through other pictures and see if this item is marked more clearly.}

\hdashrule{\linewidth}{0.4pt}{2mm}

\textcolor{blue}
{
VLM Response: place\_s\_[(1,7),(1,8),(2,3),(2,4)]
}

\hdashrule{\linewidth}{0.4pt}{2mm}

User Message:

\textbf{
The object has been placed successfully. You can now call end if you think you've finished the task correctly, or can also pick up the object again if you think the placement is not correct.}

You are currently at object\_table\_02\_40\_platform\_2.

And you are holding nothing.

Your available action space:

- ['go\_to\_object\_sofa\_10\_platform\_0', 'go\_to\_object\_sofa\_10\_platform\_1', 'go\_to\_object\_sofa\_10\_platform\_2', 'go\_to\_object\_table\_04\_13\_platform\_0', 'go\_to\_object\_chair\_04\_46\_platform\_0', 'go\_to\_object\_chair\_04\_47\_platform\_0', 'go\_to\_object\_chair\_05\_7\_platform\_0', 'go\_to\_object\_chair\_05\_8\_platform\_0', 'go\_to\_object\_stool\_02\_18\_platform\_0', 'go\_to\_object\_stool\_02\_6\_platform\_0', 'go\_to\_object\_rack\_01\_75\_platform\_0', 'go\_to\_object\_rack\_01\_75\_platform\_1', 'go\_to\_object\_rack\_01\_75\_platform\_2', 'go\_to\_object\_kitchen\_counter\_1\_body\_platform\_0', 'go\_to\_object\_kitchen\_counter\_1\_body\_platform\_1', 'go\_to\_object\_fridge\_0\_body\_platform\_1', 'go\_to\_object\_fridge\_0\_body\_platform\_3', 'go\_to\_object\_fridge\_0\_body\_platform\_4', 'go\_to\_object\_fridge\_0\_body\_platform\_5', 'go\_to\_object\_fridge\_0\_body\_platform\_6', 'go\_to\_object\_fridge\_0\_body\_platform\_7', 'go\_to\_object\_bin\_03\_3\_platform\_0', 'go\_to\_object\_wall\_cabinet\_01\_4\_platform\_0', 'go\_to\_object\_wall\_cabinet\_01\_4\_platform\_1', 'go\_to\_object\_wall\_cabinet\_01\_4\_platform\_2', 'go\_to\_object\_wall\_cabinet\_01\_4\_platform\_3', 'go\_to\_object\_wall\_cabinet\_01\_4\_platform\_4', 'go\_to\_object\_wall\_cabinet\_01\_4\_platform\_5', 'go\_to\_object\_wall\_cabinet\_01\_4\_platform\_6', 'go\_to\_object\_table\_03\_14\_platform\_1', 'go\_to\_object\_table\_01\_48\_platform\_0', 'go\_to\_object\_wall\_cabinet\_02\_21\_platform\_0', 'go\_to\_object\_wall\_cabinet\_02\_21\_platform\_1', 'go\_to\_object\_wall\_cabinet\_02\_21\_platform\_2', 'go\_to\_object\_wall\_cabinet\_02\_21\_platform\_3', 'go\_to\_object\_wall\_cabinet\_02\_21\_platform\_4', 'go\_to\_object\_wall\_cabinet\_02\_21\_platform\_5', 'go\_to\_object\_wall\_cabinet\_02\_21\_platform\_6', 'go\_to\_object\_chestOfDrawers\_01\_2\_body\_platform\_0', 'go\_to\_object\_chestOfDrawers\_01\_2\_body\_platform\_1', 'go\_to\_object\_chestOfDrawers\_01\_2\_body\_platform\_2', 'go\_to\_object\_chestOfDrawers\_01\_2\_body\_platform\_3', 'go\_to\_object\_chestOfDrawers\_01\_2\_body\_platform\_4', 'go\_to\_object\_chestOfDrawers\_01\_2\_body\_platform\_5', 'go\_to\_object\_chestOfDrawers\_01\_2\_body\_platform\_6', 'go\_to\_object\_table\_02\_40\_platform\_0', 'go\_to\_object\_table\_02\_40\_platform\_1', 'go\_to\_object\_table\_02\_40\_platform\_2', 'go\_to\_object\_tvstand\_88\_platform\_0', 'go\_to\_object\_cabinet\_3\_body\_platform\_0', 'go\_to\_object\_cabinet\_3\_body\_platform\_1']
- ['pick\_object\_1\_of\_current\_platform', 'pick\_object\_2\_of\_current\_platform', 'pick\_object\_3\_of\_current\_platform']
- ['show\_receptacle\_of\_object\_1\_of\_current\_platform', 'show\_receptacle\_of\_object\_2\_of\_current\_platform', 'show\_receptacle\_of\_object\_3\_of\_current\_platform']
- `CALL\_END`: Call end (terminate the task).
- `change\_view`: Change the viewing angle to observe the scene from another perspective.

Choose one of the available actions.

All your current available actions are listed as above (all are without quotes and square brackets). Please don't mix up it with the action space of previous steps.

Here are some useful suggestions for you to better finish the task: 

1. Your extra actions do not affect the success of the task, you only need to keep your steps within the total step limit.

2. When you want to place an object but cannot find enough space, you can try changing your standing position to find the correct orientation, try combining multiple receptacles, or remove objects that are blocking your way first(although you generally do not need to do this). Specifically, for the task asking you put object to empty platforms, try combining adjacent receptacles may be very useful.

3. For tasks that involve placing objects near other objects, please remember to get a focused image of the target object and examine it carefully. This is important because the receptacles may not be intuitive. 

4. Important: Don't output anything other than the action you want to take. You will only receive the same hint informing you your invalid action. If you make mistakes, just start output correct actions from the next prompt without any apologize.

Current task:

Move kitchenware\_white\_paper\_towel\_holder between lighting\_blue\_base\_table\_lamp and lighting\_blue\_base\_table\_lamp

Steps used: 13/20. You can only take at most 20 steps, so hurry if you've almost used all of them!

[*See Fig.~\ref{fig:success_step13}]

\textcolor{purple}
{Here are 2 images, named ['Idle\_13\_1\_out\_of\_2.png', 'Idle\_13\_2\_out\_of\_2.png'] respectively.}

\textcolor{purple}
{Their description is as follows: These image show the platform you are currently on in one view, with regions or objects marked with numbers for you to place objects accurately. The platform is object\_table\_02\_40\_platform\_2.}

\textcolor{purple}
{If the platform has no objects, a 3x3 grid will be marked on the platform to help you place objects, with each grid marked in numbers from 1 to 9. If there are objects on the platform, guiding lines will indicate the object numbers. Note that  if you see multiple images, it means the platform is too long to be captured in one image, and it's guaranteed that the same object will marked in THE SAME COLOR GUIDELINES AND THE SAME NUMBER. If you think you can't determine the number of an item in one picture, you can go through other pictures and see if this item is marked more clearly.}

\hdashrule{\linewidth}{0.4pt}{2mm}

\textcolor{blue}{VLM Response: CALL\_END}

\hdashrule{\linewidth}{0.4pt}{2mm}

(call\_ended)

Task\_result:

Target object placed between two objects. 

Task\_status: 

Success

\end{tcolorbox}

\newpage

\subsubsection{Failed Episode}
\label{appendix_subsubsec:fail_episode}
The following episode shows how Gemini-2.5-pro failed in attempting a task in AI2THOR. It picked up the object and went to the destination correctly, but it failed to recognize the correct receptacle region. 

\vspace{1cm}
\begin{figure}[h]
    \centering
    \includegraphics[width=\textwidth]{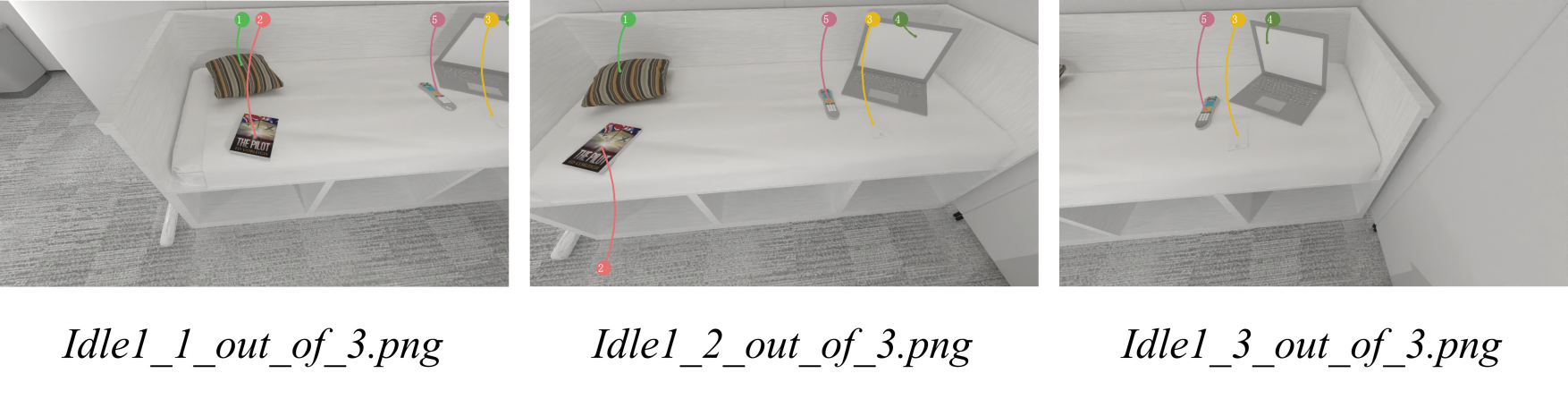}
     \vspace{-1.5em}
    \caption{\textbf{Images Given after Step 1 of the Failed Episode.}}
    \label{fig:failed_step1}
\end{figure}

\vspace{1cm}
\begin{figure}[h]
    \centering
    \includegraphics[width=\textwidth]{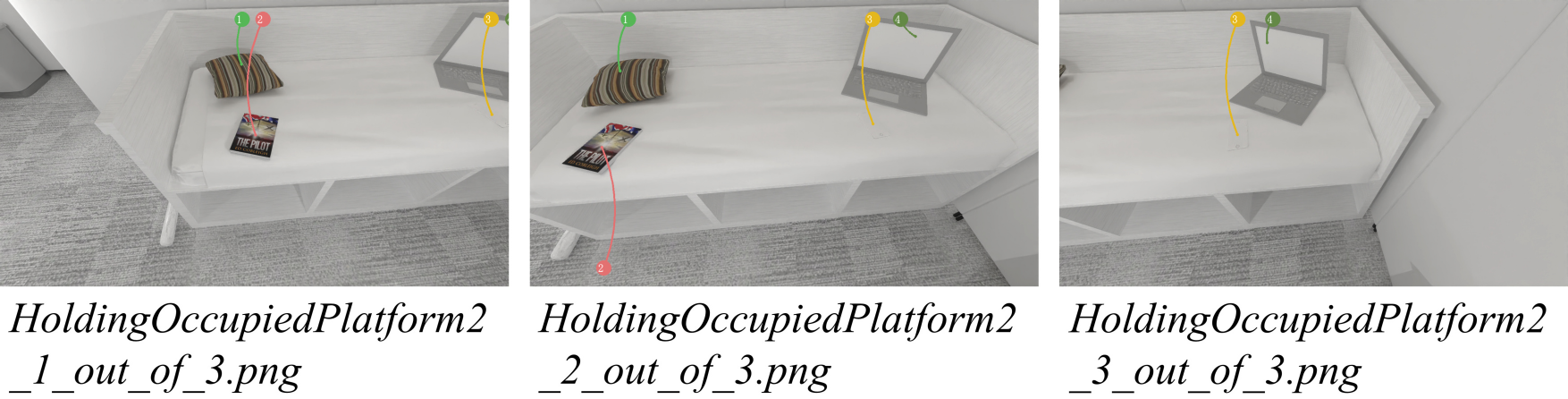}
     \vspace{-1.5em}
    \caption{\textbf{Images Given after Step 2 of the Failed Episode.}}
    \label{fig:failed_step2}
\end{figure}

\vspace{1cm}
\begin{figure}[h]
    \centering
    \includegraphics[width=\textwidth]{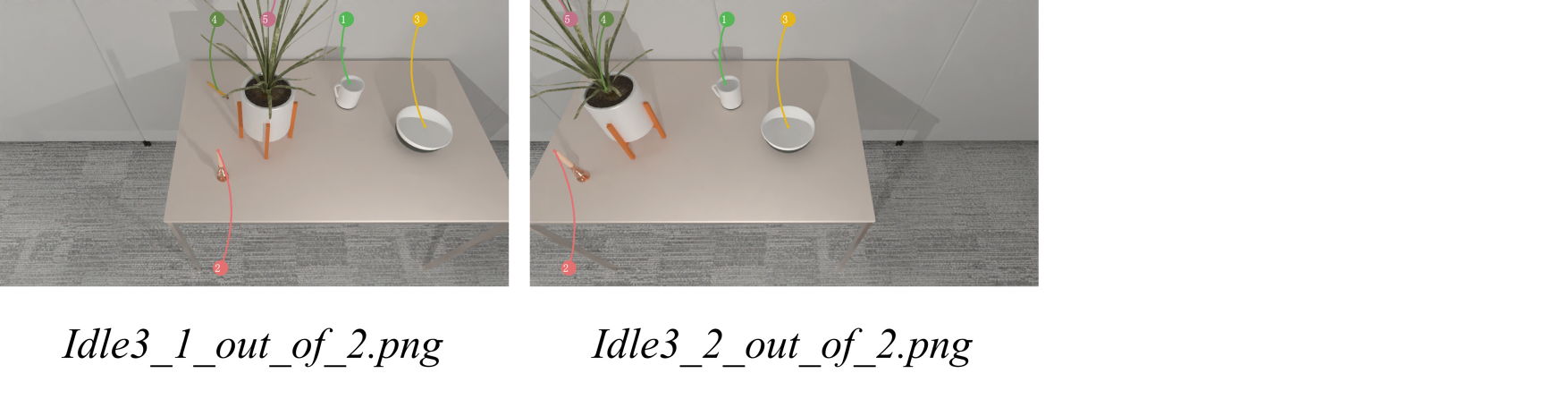}
     \vspace{-1.5em}
    \caption{\textbf{Images Given after Step 3 of the Failed Episode.}}
    \label{fig:failed_step3}
\end{figure}

\begin{figure}
    \centering
    \includegraphics[width=\textwidth]{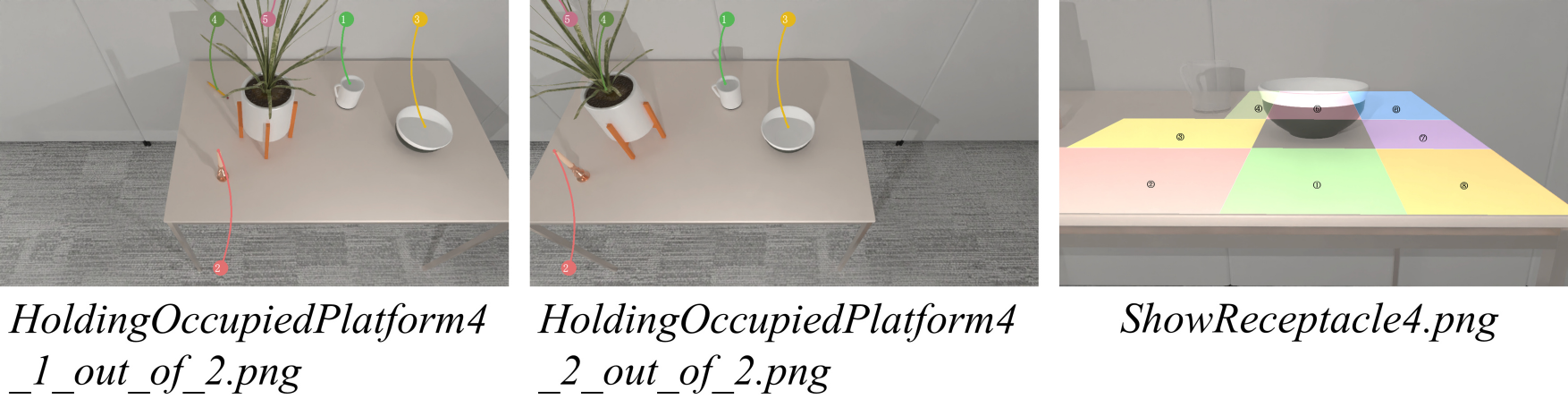}
     \vspace{-1.5em}
    \caption{\textbf{Images Given after Step 4 of the Failed Episode.}}
    \label{fig:failed_step4}
\end{figure}

\begin{figure}
    \centering
    \includegraphics[width=\textwidth]{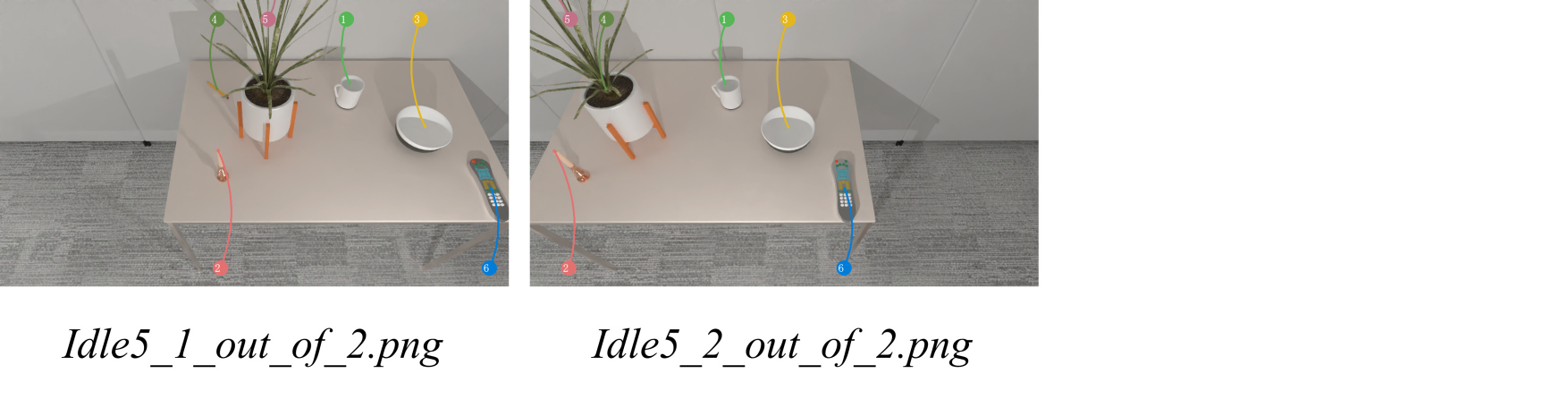}
    \vspace{-1.5em}
    \caption{\textbf{Images Given After Step 5 of the Failed Episode.}}
    \label{fig:failed_step5}
\end{figure}

\begin{tcolorbox}
[
  title={Unsuccessful Episode by Gemini-2.5-pro},
  width=\textwidth,
  breakable,
  boxrule=0.5pt,
  fonttitle=\bfseries,
  halign=flush left
]

System Message:You are a household assistant robot equipped with a vision system and manipulation capabilities. You need to complete a task in a 3D environment, mainly involving moving, picking up, and placing objects. Your goal is to move an object from one place to another correctly according to instructions. Please strictly follow the given Action Space for operations.

At the beginning of each task, you will be informed of a specific task to complete. Throughout the task execution, you will transition between multiple states and have access to a set of possible actions (action space).

You may receive one of the following task types involving one object and another object or a platform. ({an object/another object/a platform} are the name of objects or platforms, {dir} is one of the 8 directions):

- move {an object} to {a platform}.
- move {an object} to the {dir} of {a platform}.
- move {an object} around {another object}.
- move {an object} to the {dir} of {another object}.
- move {an object} between {another object} and {yet another object}.

Understanding directions:
Directions like 'front', 'rear', 'left', and 'right' are defined relative to the robot's perspective and the heading of the object or platform you are interacting with. The heading of a ground object is not fixed and depends on where the robot is standing and facing.

Empty Surfaces (Platforms): When interacting with an empty platform, imagine it divided into a 3x3 grid. The direction aligned with the heading (also, the FARTHEST middle space to you, NOT THE NEAREST MIDDLE SPACE) is labeled "front," with the remaining regions proceeding counterclockwise as "front-left," "left," "rear-left," "rear," "rear-right," "right," and "front-right."  The grid is conceptually numbered as follows, based on these directions:

  [front-left, front, front-right],

  [left, center, right],

  [rear-left, rear, rear-right]

 Occupied Surfaces (Objects on Platforms): For objects on platforms, imagine extending the edges of the object's bounding box to the platform's boundaries. The resulting regions define the object's 8 receptacles, and the direction is defined similarly.
  
Task success criteria:

- move {an object} to {a platform}: The task is considered successful only when {an object} is placed anywhere on the specified platform.

- move {an object} to the {dir} of {a platform}: The task is considered successful only when {an object} is placed on the platform and the geometric center of the placement is in the {dir} direction relative to the platform. 

- move {an object} around {another object}: The task is considered successful only when {an object} is placed in any of the receptacles surrounding {another object}.

- move {an object} to the {dir} of {another object}: The task is considered successful only when {an object} is placed in the receptacles surrounding {another object} and overlaps with the {dir} direction.

- move {an object} between {another object} and {yet another object}: The task is considered successful only when {an object} is placed such that it intersects with receptacles in a pair of opposite directions between {another object} and {yet another object}.

Your action space, includes walking to platforms, picking up items, putting items down, rotating direction, getting information about directions around an object; CALL\_END, placing an item in a specific direction on a platform, in a direction relative to another object, or placing it anywhere. Whenever you're about to choose an action, I will provide your current action space. If you choose an action that cannot be performed, such as trying to pick up an object repeatedly or attempting to place an object in a space that's too small, you'll receive a notification.

Picking up an item: Corresponds to the command "pick\_object\_{object number}\_of\_current\_platform".

Walking to a platform: Corresponds to the command "go\_to\_{platform name}".

Rotating direction: Corresponds to the command "change\_view".

Getting an image of an object: Corresponds to the command "show\_receptacle\_of\_object\_{object\_number}\_of\_current\_platform".

Putting down your holding item on empty platform: Corresponds to the command "place\_s\_[{region\_idx1},{region\_idx2}...]" where {dir\_idx1}, {dir\_idx2} are NUMBERS, indicating the index of the receptacles; Putting down your item on an occupied platform: Corresponds to the command "place\_s\_[({object\_idx1},{dir\_idx1})...]",where {object\_idx1}, {dir\_idx1} are NUMBERS, indicating the object and its corresponding receptacles. For example,  "place\_s\_[(1,2), (2,5)]" means putting down your item at the union of object 1's 2nd receptacles and object 2's 5th receptacles.Note that for placement action, the system will attempt to find a valid placement area that intersects with all the regions you selected. If it fails, it will place the object anywhere within your chosen space and you'll receive a notification.

You may be shown various types of images:

0) At the beginning of a task, if the task involves objects on a platform that contains other objects of the same type (e.g., the object is a book on a bookshelf with many books), you will receive a notification and images of these objects to help you distinguish them.

1) Whenever you move to a platform, you will see an image of the platform from your current perspective (or multiple images for longer platforms, indicating you can move horizontally to better observe the platform).
If the platform has no objects, a 3x3 grid will be marked on the platform to help you place objects.
If there are objects on the platform, guiding lines will indicate the object numbers.

2) If there are objects on a platform, you can use "show\_receptable\_of\_object\_x\_of\_current\_platform" to get a focused image of that object with numbered receptacles around it. This allows you to see the available receptacles around the object.

Here are some useful suggestions for you to better finish the task: 

1. Your extra actions do not affect the success of the task, you only need to keep your steps within the total step limit.

2. When you want to place an object but cannot find enough space, you can try changing your standing position to find the correct orientation, try combining multiple receptacles, or remove objects that are blocking your way first(although you generally do not need to do this). Specifically, for the task asking you put object to empty platforms, try combining adjacent receptacles may be very useful.

3. For tasks that involve placing objects near other objects, please remember to get a focused image of the target object and examine it carefully. This is important because the receptacles may not be intuitive. 

4. Important: Don't output anything other than the action you want to take. You will only receive the same hint informing you your invalid action. If you make mistakes, just start output correct actions from the next prompt without any apologize.

Current task:

  Move Remote to Bowl's rear-left receptacles
  Initially, Remote is on object\_hemnes\_day\_bed\_platform\_5, and Bowl is on object\_dining\_table\_mella\_platform\_0.

  Steps used: 0/20. You can only take at most 20 steps, so hurry if you've almost used all of them!

  Your available action space:

  - `CALL\_END`: Call end (terminate the task).
  - ['go\_to\_object\_hemnes\_day\_bed\_platform\_5', 'go\_to\_object\_bin\_6\_platform\_0', 'go\_to\_object\_Side\_Table\_317\_1\_platform\_1', 'go\_to\_object\_side\_table\_strind\_platform\_0', 'go\_to\_object\_side\_table\_strind\_platform\_1', 'go\_to\_object\_Side\_Table\_203\_1\_platform\_0', 'go\_to\_object\_Side\_Table\_203\_1\_platform\_1', 'go\_to\_object\_chair\_antnas\_platform\_0', 'go\_to\_object\_dresser\_birkeland\_platform\_0', 'go\_to\_object\_dresser\_birkeland\_platform\_1', 'go\_to\_object\_dresser\_birkeland\_platform\_2', 'go\_to\_object\_dresser\_birkeland\_platform\_3', 'go\_to\_object\_dresser\_birkeland\_platform\_4', 'go\_to\_object\_dresser\_birkeland\_platform\_5', 'go\_to\_object\_dresser\_birkeland\_platform\_6', 'go\_to\_object\_Basketball\_1\_platform\_0', 'go\_to\_object\_dining\_table\_mella\_platform\_0', 'go\_to\_object\_Floor\_Lamp\_13\_platform\_0', 'go\_to\_object\_armchair\_emmabo\_platform\_0', 'go\_to\_object\_BaseballBat\_2\_platform\_0']. Please choose one of the available actions.

All your current available actions are listed as above (all are without quotes and square brackets). Please don't mix up it with the action space of previous steps.

\hdashrule{\linewidth}{0.4pt}{2mm}

\textcolor{blue}{
VLM Response: go\_to\_object\_hemnes\_day\_bed\_platform\_5}

\hdashrule{\linewidth}{0.4pt}{2mm}

User Message:

You are currently at  object\_hemnes\_day\_bed\_platform\_5.

And you are holding nothing.

Your available action space:

- ['go\_to\_object\_hemnes\_day\_bed\_platform\_5', 'go\_to\_object\_bin\_6\_platform\_0', 'go\_to\_object\_Side\_Table\_317\_1\_platform\_1', 'go\_to\_object\_side\_table\_strind\_platform\_0', 'go\_to\_object\_side\_table\_strind\_platform\_1', 'go\_to\_object\_Side\_Table\_203\_1\_platform\_0', 'go\_to\_object\_Side\_Table\_203\_1\_platform\_1', 'go\_to\_object\_chair\_antnas\_platform\_0', 'go\_to\_object\_dresser\_birkeland\_platform\_0', 'go\_to\_object\_dresser\_birkeland\_platform\_1', 'go\_to\_object\_dresser\_birkeland\_platform\_2', 'go\_to\_object\_dresser\_birkeland\_platform\_3', 'go\_to\_object\_dresser\_birkeland\_platform\_4', 'go\_to\_object\_dresser\_birkeland\_platform\_5', 'go\_to\_object\_dresser\_birkeland\_platform\_6', 'go\_to\_object\_Basketball\_1\_platform\_0', 'go\_to\_object\_dining\_table\_mella\_platform\_0', 'go\_to\_object\_Floor\_Lamp\_13\_platform\_0', 'go\_to\_object\_armchair\_emmabo\_platform\_0', 'go\_to\_object\_BaseballBat\_2\_platform\_0']
- ['pick\_object\_1\_of\_current\_platform', 'pick\_object\_2\_of\_current\_platform', 'pick\_object\_3\_of\_current\_platform', 'pick\_object\_4\_of\_current\_platform', 'pick\_object\_5\_of\_current\_platform']
- ['show\_receptacles\_of\_object\_1\_of\_current\_platform', 'show\_receptacles\_of\_object\_2\_of\_current\_platform', 'show\_receptacles\_of\_object\_3\_of\_current\_platform', 'show\_receptacles\_of\_object\_4\_of\_current\_platform', 'show\_receptacles\_of\_object\_5\_of\_current\_platform']
- `CALL\_END`: Call end (terminate the task).
- `rotate\_observation\_view\_of\_current\_platform`: Change the viewing angle to observe the scene from another perspective.

Choose one of the available actions.

All your current available actions are listed as above (all are without quotes and square brackets). Please don't mix up it with the action space of previous steps.

Here are some useful suggestions for you to better finish the task: 

1. Your extra actions do not affect the success of the task, you only need to keep your steps within the total step limit.

2. When you want to place an object but cannot find enough space, you can try changing your standing position to find the correct orientation, try combining multiple receptacles, or remove objects that are blocking your way first(although you generally do not need to do this). Specifically, for the task asking you put object to empty platforms, try combining adjacent receptacles may be very useful.

3. For tasks that involve placing objects near other objects, please remember to get a focused image of the target object and examine it carefully. This is important because the receptacles may not be intuitive. 

4. Important: Don't output anything other than the action you want to take. You will only receive the same hint informing you your invalid action. If you make mistakes, just start output correct actions from the next prompt without any apologize.

Current task:

Move kitchenware\_white\_paper\_towel\_holder between lighting\_blue\_base\_table\_lamp and lighting\_blue\_base\_table\_lamp

Steps used: 1/20. You can only take at most 20 steps, so hurry if you've almost used all of them!

[*See Fig.~\ref{fig:failed_step1}]

\textcolor{purple}
{Here are 3 images, named ['Idle\_1\_1\_out\_of\_3.png', 'Idle\_1\_2\_out\_of\_3.png', 'Idle\_1\_3\_out\_of\_3.png'] respectively.}
\textcolor{purple}
{Their description is as follows: These images show the platform you are currently on in one view, with regions or objects marked with numbers for you to place objects accurately. The platform is object\_hemnes\_day\_bed\_platform\_5.}
\textcolor{purple}
{If the platform has no objects, a 3x3 grid will be marked on the platform to help you place objects, with each grid marked in numbers from 1 to 9. If there are objects on the platform, guiding lines will indicate the object numbers. Note that  if you see multiple images, it means the platform is too long to be captured in one image, and it's guaranteed that the same object will marked in THE SAME COLOR GUIDELINES AND THE SAME NUMBER. If you think you can't determine the number of an item in one picture, you can go through other pictures and see if this item is marked more clearly.
}

\hdashrule{\linewidth}{0.4pt}{2mm}

\textcolor{blue}
{VLM Response: pick\_object\_5\_of\_current\_platform}

\hdashrule{\linewidth}{0.4pt}{2mm}

You are currently at  object\_hemnes\_day\_bed\_platform\_5.

And you are holding Remote.

Your available action space:

-['go\_to\_object\_hemnes\_day\_bed\_platform\_5', 'go\_to\_object\_bin\_6\_platform\_0', 'go\_to\_object\_Side\_Table\_317\_1\_platform\_1', 'go\_to\_object\_side\_table\_strind\_platform\_0', 'go\_to\_object\_side\_table\_strind\_platform\_1', 'go\_to\_object\_Side\_Table\_203\_1\_platform\_0', 'go\_to\_object\_Side\_Table\_203\_1\_platform\_1', 'go\_to\_object\_chair\_antnas\_platform\_0', 'go\_to\_object\_dresser\_birkeland\_platform\_0', 'go\_to\_object\_dresser\_birkeland\_platform\_1', 'go\_to\_object\_dresser\_birkeland\_platform\_2', 'go\_to\_object\_dresser\_birkeland\_platform\_3', 'go\_to\_object\_dresser\_birkeland\_platform\_4', 'go\_to\_object\_dresser\_birkeland\_platform\_5', 'go\_to\_object\_dresser\_birkeland\_platform\_6', 'go\_to\_object\_Basketball\_1\_platform\_0', 'go\_to\_object\_dining\_table\_mella\_platform\_0', 'go\_to\_object\_Floor\_Lamp\_13\_platform\_0', 'go\_to\_object\_armchair\_emmabo\_platform\_0', 'go\_to\_object\_BaseballBat\_2\_platform\_0']

- ['show\_receptacles\_of\_object\_1\_of\_current\_platform', 'show\_receptacles\_of\_object\_2\_of\_current\_platform', 'show\_receptacles\_of\_object\_3\_of\_current\_platform', 'show\_receptacles\_of\_object\_4\_of\_current\_platform']
- `CALL\_END`: Call end (terminate the task).
- `rotate\_observation\_view\_of\_current\_platform`: Change the viewing angle to observe the scene from another perspective.
- `place\_r`:Place the object in an arbitrary place on the platform. - `place\_s\_[({object\_idx1},{dir\_idx1})...]`: Place the object in the union area of regions with indices {(object\_idx1,region\_idx1)}(object\_idx1's region\_idx1 space)  etc. All available pairs are in [(1, 1), (1, 2), (1, 3), (1, 4), (1, 5), (2, 1), (2, 2), (2, 3), (3, 1), (3, 2), (3, 3), (3, 4), (3, 5), (3, 6), (4, 1), (4, 2), (4, 3), (4, 4), (4, 5), (4, 6), (4, 7)], and a valid choice should be a subset of it. 

All your current available actions are listed as follows (all are without quotes and square brackets). Please don't mix up it with the action space of previous steps.

Current task:

Move Remote to Bowl's rear-left receptacles

Steps used: 2/20. You can only take at most 20 steps, so hurry if you've almost used all of them!

Understanding directions:
Directions like 'front', 'rear', 'left', and 'right' are defined relative to the robot's perspective and the heading of the object or platform you are interacting with. The heading of a ground object is not fixed and depends on where the robot is standing and facing.

Empty Surfaces (Platforms): When interacting with an empty platform, imagine it divided into a 3x3 grid. The direction aligned with the heading (also, the FARTHEST middle space to you, NOT THE NEAREST MIDDLE SPACE) is labeled "front," with the remaining regions proceeding counterclockwise as "front-left," "left," "rear-left," "rear," "rear-right," "right," and "front-right."  The grid is conceptually numbered as follows, based on these directions:

  [front-left, front, front-right],
  [left, center, right],
  [rear-left, rear, rear-right]

 Occupied Surfaces (Objects on Platforms): For objects on platforms, imagine extending the edges of the object's bounding box to the platform's boundaries. The resulting regions define the object's 8 receptacles, and the direction is defined similarly.
   Here are some useful suggestions for you to better finish the task: 

1. Your extra actions do not affect the success of the task, you only need to keep your steps within the total step limit.

2. When you want to place an object but cannot find enough space, you can try changing your standing position to find the correct orientation, try combining multiple receptacles, or remove objects that are blocking your way first(although you generally do not need to do this). Specifically, for the task asking you put object to empty platforms, try combining adjacent receptacles may be very useful.

3. For tasks that involve placing objects near other objects, please remember to get a focused image of the target object and examine it carefully. This is important because the regions may not be intuitive. 

4. Important: Don't output anything other than the action you want to take. You will only receive the same hint informing you your invalid action. If you make mistakes, just start output correct actions from the next prompt without any apologize.

[*See Fig.~\ref{fig:failed_step2}]
\textcolor{purple}
{Here are 3 images, named ['HoldingOccupiedPlatformState\_2\_1\_out\_of\_3.png', 'HoldingOccupiedPlatformState\_2\_2\_out\_of\_3.png', 'HoldingOccupiedPlatformState\_2\_3\_out\_of\_3.png'] respectively.}

\textcolor{purple}
{
Their description is as follows: This image shows the platform you are currently on in one view, with regions or objects marked with numbers for you to place objects accurately. The platform is object\_hemnes\_day\_bed\_platform\_5.}

\textcolor{purple}
{
If the platform has no objects, a 3x3 grid will be marked on the platform to help you place objects, with each grid marked in numbers from 1 to 9. If there are objects on the platform, guiding lines will indicate the object numbers. Note that  if you see multiple images, it means the platform is too long to be captured in one image, and it's guaranteed that the same object will marked in THE SAME COLOR GUILDLINES AND THE SAME NUMBER. If you think you can't determine the number of an item in one picture, you can go through other pictures and see if this item is marked more clearly.}

\hdashrule{\linewidth}{0.4pt}{2mm}

\textcolor{blue}
{VLM Response: go\_to\_object\_dining\_table\_mella\_platform\_0}

\hdashrule{\linewidth}{0.4pt}{2mm}

You are currently at object\_dining\_table\_mella\_platform\_0.

And you are holding Remote.

Your available action space:

['go\_to\_object\_hemnes\_day\_bed\_platform\_5', 'go\_to\_object\_bin\_6\_platform\_0', 'go\_to\_object\_Side\_Table\_317\_1\_platform\_1', 'go\_to\_object\_side\_table\_strind\_platform\_0', 'go\_to\_object\_side\_table\_strind\_platform\_1', 'go\_to\_object\_Side\_Table\_203\_1\_platform\_0', 'go\_to\_object\_Side\_Table\_203\_1\_platform\_1', 'go\_to\_object\_chair\_antnas\_platform\_0', 'go\_to\_object\_dresser\_birkeland\_platform\_0', 'go\_to\_object\_dresser\_birkeland\_platform\_1', 'go\_to\_object\_dresser\_birkeland\_platform\_2', 'go\_to\_object\_dresser\_birkeland\_platform\_3', 'go\_to\_object\_dresser\_birkeland\_platform\_4', 'go\_to\_object\_dresser\_birkeland\_platform\_5', 'go\_to\_object\_dresser\_birkeland\_platform\_6', 'go\_to\_object\_Basketball\_1\_platform\_0', 'go\_to\_object\_dining\_table\_mella\_platform\_0', 'go\_to\_object\_Floor\_Lamp\_13\_platform\_0', 'go\_to\_object\_armchair\_emmabo\_platform\_0', 'go\_to\_object\_BaseballBat\_2\_platform\_0']

- ['show\_receptacles\_of\_object\_1\_of\_current\_platform', 'show\_receptacles\_of\_object\_2\_of\_current\_platform', 'show\_receptacles\_of\_object\_3\_of\_current\_platform', 'show\_receptacles\_of\_object\_4\_of\_current\_platform', 'show\_receptacles\_of\_object\_5\_of\_current\_platform']
- `CALL\_END`: Call end (terminate the task).
- `rotate\_observation\_view\_of\_current\_platform`: Change the viewing angle to observe the scene from another perspective.
- `place\_r`:Place the object in an arbitrary place on the platform. - `place\_s\_[({object\_idx1},{dir\_idx1})...]`: Place the object in the union area of regions with indices {(object\_idx1,region\_idx1)}(object\_idx1's region\_idx1 space)  etc. All available pairs are in [(1, 1), (1, 2), (1, 3), (1, 4), (1, 5), (1, 6), (1, 7), (1, 8), (2, 1), (2, 2), (2, 3), (2, 4), (2, 5), (3, 1), (3, 2), (3, 3), (3, 4), (3, 5), (3, 6), (3, 7), (3, 8), (4, 1), (4, 2), (4, 3), (4, 4), (4, 5), (4, 6), (4, 7), (5, 1), (5, 2), (5, 3), (5, 4), (5, 5)], and a valid choice should be a subset of it.

All your current available actions are listed as follows (all are without quotes and square brackets). Please don't mix up it with the action space of previous steps.

Current task:

Move Remote to Bowl's rear-left receptacles

Steps used: 3/20. You can only take at most 20 steps, so hurry if you've almost used all of them!

Understanding directions:
Directions like 'front', 'rear', 'left', and 'right' are defined relative to the robot's perspective and the heading of the object or platform you are interacting with. The heading of a ground object is not fixed and depends on where the robot is standing and facing.

Empty Surfaces (Platforms): When interacting with an empty platform, imagine it divided into a 3x3 grid. The direction aligned with the heading (also, the FARTHEST middle space to you, NOT THE NEAREST MIDDLE SPACE) is labeled "front," with the remaining regions proceeding counterclockwise as "front-left," "left," "rear-left," "rear," "rear-right," "right," and "front-right."  The grid is conceptually numbered as follows, based on these directions:

  [front-left, front, front-right],
  [left, center, right],
  [rear-left, rear, rear-right]

 Occupied Surfaces (Objects on Platforms): For objects on platforms, imagine extending the edges of the object's bounding box to the platform's boundaries. The resulting regions define the object's 8 receptacles, and the direction is defined similarly.
   Here are some useful suggestions for you to better finish the task: 

1. Your extra actions do not affect the success of the task, you only need to keep your steps within the total step limit.

2. When you want to place an object but cannot find enough space, you can try changing your standing position to find the correct orientation, try combining multiple receptacles, or remove objects that are blocking your way first(although you generally do not need to do this). Specifically, for the task asking you put object to empty platforms, try combining adjacent receptacles may be very useful.

3. For tasks that involve placing objects near other objects, please remember to get a focused image of the target object and examine it carefully. This is important because the regions may not be intuitive. 

4. Important: Don't output anything other than the action you want to take. You will only receive the same hint informing you your invalid action. If you make mistakes, just start output correct actions from the next prompt without any apologize.

[*See Fig.~\ref{fig:failed_step3}]

\textcolor{purple}
{Here are 2 images, named ['HoldingOccupiedPlatform\_3\_1\_out\_of\_2.png', 'HoldingOccupiedPlatform\_3\_2\_out\_of\_2.png'] respectively.}

\textcolor{purple}
{Their description is as follows: This image shows the platform you are currently on in one view, with regions or objects marked with numbers for you to place objects accurately. The platform is object\_dining\_table\_mella\_platform\_0.}

\textcolor{purple}
{If the platform has no objects, a 3x3 grid will be marked on the platform to help you place objects, with each grid marked in numbers from 1 to 9. If there are objects on the platform, guiding lines will indicate the object numbers. Note that  if you see multiple images, it means the platform is too long to be captured in one image, and it's guaranteed that the same object will marked in THE SAME COLOR GUILDLINES AND THE SAME NUMBER. If you think you can't determine the number of an item in one picture, you can go through other pictures and see if this item is marked more clearly.}

\hdashrule{\linewidth}{0.4pt}{2mm}

\textcolor{blue}
{VLM Response: show\_receptacle\_of\_object\_5\_of\_current\_platform}

\hdashrule{\linewidth}{0.4pt}{2mm}

[* See Fig.~\ref{fig:failed_step4} \textit{ShowReceptacle\_4.png}]

Here are 1 images, named ['ShowReceptacle\_4.png'] respectively.

Their description is as follows: This image shows the available receptacles around object 3 on the platform, you see this because you've taken "show\_receptacles" action previously. The numbers from 1 to 8 indicate the available receptacles. Not all numbers from 1 to 8 may be present, as some receptacles might be too small or blocked. To use the numbers shown in the image to specify where you want to place the object you are holding, you can include (3, {receptacles\_idx} (receptacles\_idx in [1, 8]) in the output action.

You are currently at

object\_dining\_table\_mella\_platform\_0.

And you are holding Remote.

Your available action space:

-['go\_to\_object\_hemnes\_day\_bed\_platform\_5', 'go\_to\_object\_bin\_6\_platform\_0', 'go\_to\_object\_Side\_Table\_317\_1\_platform\_1', 'go\_to\_object\_side\_table\_strind\_platform\_0', 'go\_to\_object\_side\_table\_strind\_platform\_1', 'go\_to\_object\_Side\_Table\_203\_1\_platform\_0', 'go\_to\_object\_Side\_Table\_203\_1\_platform\_1', 'go\_to\_object\_chair\_antnas\_platform\_0', 'go\_to\_object\_dresser\_birkeland\_platform\_0', 'go\_to\_object\_dresser\_birkeland\_platform\_1', 'go\_to\_object\_dresser\_birkeland\_platform\_2', 'go\_to\_object\_dresser\_birkeland\_platform\_3', 'go\_to\_object\_dresser\_birkeland\_platform\_4', 'go\_to\_object\_dresser\_birkeland\_platform\_5', 'go\_to\_object\_dresser\_birkeland\_platform\_6', 'go\_to\_object\_Basketball\_1\_platform\_0', 'go\_to\_object\_dining\_table\_mella\_platform\_0', 'go\_to\_object\_Floor\_Lamp\_13\_platform\_0', 'go\_to\_object\_armchair\_emmabo\_platform\_0', 'go\_to\_object\_BaseballBat\_2\_platform\_0']

- ['show\_receptacles\_of\_object\_1\_of\_current\_platform', 'show\_receptacles\_of\_object\_2\_of\_current\_platform', 'show\_receptacles\_of\_object\_3\_of\_current\_platform', 'show\_receptacles\_of\_object\_4\_of\_current\_platform', 'show\_receptacles\_of\_object\_5\_of\_current\_platform']
- `CALL\_END`: Call end (terminate the task).
- `rotate\_observation\_view\_of\_current\_platform`: Change the viewing angle to observe the scene from another perspective.
- `place\_r`:Place the object in an arbitrary place on the platform. - `place\_s\_[({object\_idx1},{dir\_idx1})...]`: Place the object in the union area of regions with indices {(object\_idx1,region\_idx1)}(object\_idx1's region\_idx1 space)  etc. All available pairs are in [(1, 1), (1, 2), (1, 3), (1, 4), (1, 5), (1, 6), (1, 7), (1, 8), (2, 1), (2, 2), (2, 3), (2, 4), (2, 5), (3, 1), (3, 2), (3, 3), (3, 4), (3, 5), (3, 6), (3, 7), (3, 8), (4, 1), (4, 2), (4, 3), (4, 4), (4, 5), (4, 6), (4, 7), (5, 1), (5, 2), (5, 3), (5, 4), (5, 5)], and a valid choice should be a subset of it. 

All your current available actions are listed as follows (all are without quotes and square brackets). Please don't mix up it with the action space of previous steps.

Current task:

Move Remote to Bowl's rear-left receptacles

Steps used: 4/20. You can only take at most 20 steps, so hurry if you've almost used all of them!

Understanding directions:
Directions like 'front', 'rear', 'left', and 'right' are defined relative to the robot's perspective and the heading of the object or platform you are interacting with. The heading of a ground object is not fixed and depends on where the robot is standing and facing.

Empty Surfaces (Platforms): When interacting with an empty platform, imagine it divided into a 3x3 grid. The direction aligned with the heading (also, the FARTHEST middle space to you, NOT THE NEAREST MIDDLE SPACE) is labeled "front," with the remaining regions proceeding counterclockwise as "front-left," "left," "rear-left," "rear," "rear-right," "right," and "front-right."  The grid is conceptually numbered as follows, based on these directions:

  [front-left, front, front-right],
  [left, center, right],
  [rear-left, rear, rear-right]

 Occupied Surfaces (Objects on Platforms): For objects on platforms, imagine extending the edges of the object's bounding box to the platform's boundaries. The resulting regions define the object's 8 receptacles, and the direction is defined similarly.
   Here are some useful suggestions for you to better finish the task: 

1. Your extra actions do not affect the success of the task, you only need to keep your steps within the total step limit.

2. When you want to place an object but cannot find enough space, you can try changing your standing position to find the correct orientation, try combining multiple receptacles, or remove objects that are blocking your way first(although you generally do not need to do this). Specifically, for the task asking you put object to empty platforms, try combining adjacent receptacles may be very useful.

3. For tasks that involve placing objects near other objects, please remember to get a focused image of the target object and examine it carefully. This is important because the regions may not be intuitive. 

4. Important: Don't output anything other than the action you want to take. You will only receive the same hint informing you your invalid action. If you make mistakes, just start output correct actions from the next prompt without any apologize.

[* See Fig.~\ref{fig:failed_step4} \textit{HoldingOccupiedPlatform\_4\_1\_out\_of\_2.png},\textit{HoldingOccupiedPlatform\_4\_2\_out\_of\_2.png}]

\textcolor{purple}
{Here are 2 images, named ['HoldingOccupiedPlatform4\_1\_out\_of\_2.png', 'HoldingOccupiedPlatform4\_2\_out\_of\_2.png'] respectively.}

\textcolor{purple}
{Their description is as follows: This image shows the platform you are currently on in one view, with regions or objects marked with numbers for you to place objects accurately. The platform is object\_dining\_table\_mella\_platform\_0.}

\textcolor{purple}
{If the platform has no objects, a 3x3 grid will be marked on the platform to help you place objects, with each grid marked in numbers from 1 to 9. If there are objects on the platform, guiding lines will indicate the object numbers. Note that  if you see multiple images, it means the platform is too long to be captured in one image, and it's guaranteed that the same object will marked in THE SAME COLOR GUILDLINES AND THE SAME NUMBER. If you think you can't determine the number of an item in one picture, you can go through other pictures and see if this item is marked more clearly.}

\hdashrule{\linewidth}{0.4pt}{2mm}

\textcolor{blue}
{VLM\_response: place\_s\_[(3,8)]}

\hdashrule{\linewidth}{0.4pt}{2mm}

\textbf{
The object has been placed successfully. You can now call end if you think you've finished the task correctly, or can also pick up the object again if you think the placement is not correct.}

You are currently at object\_dining\_table\_mella\_platform\_0.

And you are holding nothing.

Your available action space:

- ['go\_to\_object\_hemnes\_day\_bed\_platform\_5', 'go\_to\_object\_bin\_6\_platform\_0', 'go\_to\_object\_Side\_Table\_317\_1\_platform\_1', 'go\_to\_object\_side\_table\_strind\_platform\_0', 'go\_to\_object\_side\_table\_strind\_platform\_1', 'go\_to\_object\_Side\_Table\_203\_1\_platform\_0', 'go\_to\_object\_Side\_Table\_203\_1\_platform\_1', 'go\_to\_object\_chair\_antnas\_platform\_0', 'go\_to\_object\_dresser\_birkeland\_platform\_0', 'go\_to\_object\_dresser\_birkeland\_platform\_1', 'go\_to\_object\_dresser\_birkeland\_platform\_2', 'go\_to\_object\_dresser\_birkeland\_platform\_3', 'go\_to\_object\_dresser\_birkeland\_platform\_4', 'go\_to\_object\_dresser\_birkeland\_platform\_5', 'go\_to\_object\_dresser\_birkeland\_platform\_6', 'go\_to\_object\_Basketball\_1\_platform\_0', 'go\_to\_object\_dining\_table\_mella\_platform\_0', 'go\_to\_object\_Floor\_Lamp\_13\_platform\_0', 'go\_to\_object\_armchair\_emmabo\_platform\_0', 'go\_to\_object\_BaseballBat\_2\_platform\_0']
- ['pick\_object\_1\_of\_current\_platform', 'pick\_object\_2\_of\_current\_platform', 'pick\_object\_3\_of\_current\_platform', 'pick\_object\_4\_of\_current\_platform', 'pick\_object\_5\_of\_current\_platform', 'pick\_object\_6\_of\_current\_platform']
- ['show\_receptacles\_of\_object\_1\_of\_current\_platform', 'show\_receptacles\_of\_object\_2\_of\_current\_platform', 'show\_receptacles\_of\_object\_3\_of\_current\_platform', 'show\_receptacles\_of\_object\_4\_of\_current\_platform', 'show\_receptacles\_of\_object\_5\_of\_current\_platform', 'show\_receptacles\_of\_object\_6\_of\_current\_platform']

- `CALL\_END`: Call end (terminate the task).
- `rotate\_observation\_view\_of\_current\_platform`: Change the viewing angle to observe the scene from another perspective.

Choose one of the available actions.

All your current available actions are listed as follows (all are without quotes and square brackets). Please don't mix up it with the action space of previous steps.

Current task:

Move Remote to Bowl's rear-left receptacles

Steps used: 5/20. You can only take at most 20 steps, so hurry if you've almost used all of them!

Here are some useful suggestions for you to better finish the task: 

1. Your extra actions do not affect the success of the task, you only need to keep your steps within the total step limit.

2. When you want to place an object but cannot find enough space, you can try changing your standing position to find the correct orientation, try combining multiple receptacles, or remove objects that are blocking your way first(although you generally do not need to do this). Specifically, for the task asking you put object to empty platforms, try combining adjacent receptacles may be very useful.

3. For tasks that involve placing objects near other objects, please remember to get a focused image of the target object and examine it carefully. This is important because the regions may not be intuitive. 

4. Important: Don't output anything other than the action you want to take. You will only receive the same hint informing you your invalid action. If you make mistakes, just start output correct actions from the next prompt without any apologize.

[* See Fig.~\ref{fig:failed_step5}]

\textcolor{purple}
{Here are 2 images, named ['Idle\_5\_1\_out\_of\_2.png', 'Idle\_5\_2\_out\_of\_2.png']  respectively.}

\textcolor{purple}
{Their description is as follows: This image shows the platform you are currently on in one view, with regions or objects marked with numbers for you to place objects accurately. The platform is object\_dining\_table\_mella\_platform\_0.}

\textcolor{purple}
{If the platform has no objects, a 3x3 grid will be marked on the platform to help you place objects, with each grid marked in numbers from 1 to 9. If there are objects on the platform, guiding lines will indicate the object numbers. Note that  if you see multiple images, it means the platform is too long to be captured in one image, and it's guaranteed that the same object will marked in THE SAME COLOR GUILDLINES AND THE SAME NUMBER. If you think you can't determine the number of an item in one picture, you can go through other pictures and see if this item is marked more clearly.}

\hdashrule{\linewidth}{0.4pt}{2mm}

\textcolor{blue}
{VLM Response: CALL\_END}

\hdashrule{\linewidth}{0.4pt}{2mm}

(call\_ended)

Task\_result:

Target object placed in wrong direction, expected: rear-left, found: rear-right. 

Task\_status: 

Failed

\end{tcolorbox}

\end{document}